\ifdefined\pdftexversion\pdfoutput=1\fi
\documentclass{article}
\usepackage{url}
\usepackage{amsmath}
\usepackage{amsfonts}
\usepackage{amssymb}
\usepackage{natbib}
\usepackage{graphicx}
\usepackage{arxiv_new}
\usepackage{nicefrac}
\usepackage{microtype}
\usepackage{hyperref}
\usepackage{mathtools}
\usepackage{subcaption}
\usepackage{xcolor}

\newcommand{\LI}{\mathrm{Gate}}
\newcommand{\diag}{\operatorname{diag}}
\newcommand{\one}{\mathbf{1}}
\allowdisplaybreaks
\let\cite\citep
\begin{document}

\title{ConwayNet: Interpretability, Activation Coherence, and Noise Resilience via a Game-of-Life-Inspired Lateral Inhibition in Artificial Neural Networks}

\author{Yair Cohen~\href{https://www.linkedin.com/in/yair-cohen-a5a0a11a1/}{\small\textcolor{blue}{(LinkedIn)}}\\ {\small\href{mailto:yair@lateralintuition.com}{\textcolor{blue}{yair@lateralintuition.com}}}\\[6pt] Scott Gilmore~\href{https://www.linkedin.com/in/scott-gilmore-md/}{\small\textcolor{blue}{(LinkedIn)}}}

\date{\today}

\maketitle

\vspace{0.3in}

\begin{abstract}
Deep neural networks achieve universal function approximation by stacking interconnected nonlinearities, but as a consequence they are sensitive to noise and resist mechanistic interpretation. In contrast, the local update rules of cellular automata, such as Conway's Game of Life, give rise to not merely interpretable but {\bf{programmable}} dynamics. We explore a bridge between these worlds: a lateral inhibition in neural networks based on continuous Game-of-Life rules that derives inhibitory and excitatory responses from the layer input and blends them into the activation map in a state-conditioned way. The mechanism is differentiable and architecture-agnostic, and its regulation is imposed rather than learned: the rule's parameters are fixed and are not optimised, while the rule's values and derivatives shape training throughout the network. Wired into fully connected, convolutional, wide-residual and graph convolutional networks, it gives rise to several properties. Activation maps become coherent across space, depth and seeds, with coarse structure persisting through depth while fine detail is renewed at every layer. The co-activation graph of the hidden units departs from an Erd\H{o}s--R\'enyi random graph: it is laid out on the image and its giant component grows in steps, as in a spatial graph. Networks started from nearly identical initialisations, under label noise, disagree on thirty times fewer test images. Saliency maps rest on the object rather than the background, and resilience to pixel noise not seen in training is substantially improved.
\end{abstract}

\section{Introduction}

Artificial neural networks are the main paradigm of artificial intelligence, behind many of the technological breakthroughs of the past decade. Cellular automata (CA) are a foundational modelling framework of artificial life, dating back to von Neumann~\cite{vonNeumann1966}. The resemblance between the two has long been noted: both consist of many simple units that apply a nonlinear rule in parallel, and a CA rule can be learned by a neural network~\cite{WulffHertz1992} or written exactly as a recurrent convolutional network~\cite{Gilpin2019}. In a multilayer perceptron the recursion unfolds across depth; in a CA, across time. The differences are as instructive. A CA rule is \emph{local}: a cell sees only its neighbours. A network layer is \emph{non-local}: an affine map or convolution lets each unit read many units below~\cite{Cybenko1989,Hornik1991}, deep convolutional units integrate large regions of the input~\cite{Luo2016}, and attention mixes all positions at once~\cite{Vaswani2017,Dosovitskiy2021}. A CA applies the same fixed rule at every step, whereas each layer of a network applies its own learned map. And in CA such as the Game of Life the update depends on the cell's own state, which a network layer does not distinguish.

A cellular automaton consists of a grid of cells that update synchronously by one local rule, mapping a cell's state and its neighbours' states to its next state~\cite{vonNeumann1966,Wolfram1984,Wolfram2002}. In Conway's Game of Life~\cite{Gardner1970} the state is binary and the rule \emph{outer-totalistic}: a dead cell becomes alive if its number of live Moore neighbours is in a \emph{Birth} set $\mathcal B$, a live cell survives if it is in a \emph{Survival} set $\mathcal S$, and all other cells die; the Game of Life is B3/S23 ($\mathcal B=\{3\}$, $\mathcal S=\{2,3\}$). The update is thus \emph{state-dependent}, and many complex (``class~IV'') Life-like rules need $\mathcal B\ne\mathcal S$: no rule with $\mathcal B=\mathcal S$ supports a compact glider from a $3\times3$ seed~\citep{Cohen2026}.

Seen from neural networks, the Game of Life is impressive less for being computationally universal---recurrent networks are too~\cite{SiegelmannSontag1995}---than for how it computes: through structures that can be programmed by hand. Persistent localised patterns act as memory and mobile ``gliders'' as signals, and practitioners arrange them in the initial grid into logic gates, counters, self-replicators and Turing machines~\cite{Gardner1970,Rendell2011,Cook2004}, even into a Game of Life emulated by the Game of Life itself. The initial condition is the ``software'', the structures the rule sustains are the ``hardware'', and both can be read off the grid: a \emph{native mechanistic interpretability}~\cite{Wolfram2002,Mitchell2009,Sipper1997}. Neural networks offer nothing comparable. The dense, non-local interconnection behind their expressivity makes their hidden layers hard to read, and their interpretation remains largely post hoc~\cite{Olah2017,Lipton2016,Guidotti2018,ZeilerFergus2014}.

This contrast has motivated fusions of the two, most prominently \emph{neural cellular automata} (NCA): a small shared neural update rule, applied recurrently to every cell from local percepts and a vector of visible and hidden channels, is rolled out for hundreds of steps and trained by gradient descent towards system-level objectives such as growing or regenerating a target image~\cite{Mordvintsev2020GrowingCA}. Extensions generate textures~\cite{Niklasson2021Textures} and 3D artefacts~\cite{Sudhakaran2021}, address anisotropy~\cite{Mordvintsev2022IsoNCA,Randazzo2023SteerableNCA}, learn spatio-temporal dynamics~\cite{Richardson2024}, segment images~\cite{Deutges2024MedNCA}, examine Turing completeness, which needs components immune to backpropagation~\cite{bena2025unca}, and solve multi-step visual reasoning tasks such as mazes, Sudoku and ARC-AGI-1~\cite{Etcheverry2026ReasoningNCA}. NCA \emph{learn} their rule. Their update reads the cell's own state, so it can come to depend on it, but nothing requires it to: its state dependence is whatever training favours (the growing NCA impose only one fixed state-dependent step, an alive mask~\cite{Mordvintsev2020GrowingCA}), and the learned dynamics can resemble Lindenmayer systems~\cite{Lindenmayer1968a,Lindenmayer1968b,Prusinkiewicz1990ABOP} more than the Game of Life, whose balance of growth and decay sits near an \emph{edge of chaos}~\cite{Langton1990EdgeOfChaos}.

We propose a complementary fusion of NN and CA: keep the network, and add to each layer a fixed, Game-of-Life-like rule that regulates it. We call the mechanism \emph{lateral inhibition}, in the broad sense of the regulation of a unit by its neighbours (through its Birth gate it also excites). A continuous Birth and Survival rule is computed from the \emph{layer input}, the occupancy of each unit's neighbourhood before the affine map (or any non-local operation), and applied to the \emph{layer output}, continuously blending Birth and Survival by the amplitude of the unit's own output state. The network keeps the non-local learned maps that carry the information flow learning needs, while a local, state-dependent rule without learnable parameters regulates each layer: the optimiser cannot repurpose the rule, only learn to work within it. The mechanism fits on top of essentially any layer or block; we wire it into fully connected, convolutional and wide residual networks, and the otherwise identical network without it, which we call \emph{plain}, is the baseline throughout.

\paragraph{Scope.} The rule is one instance of a large family of parameter-free rules by which a layer's input can regulate its output, and we compare it with several others; we study it mechanistically, reporting benefits and costs, rather than benchmark it against other methods. We compare rules in the multilayer perceptron (MLP), whose every hidden layer is a full-resolution grid, so that its stack of layers runs a CA-like simulation. We compare ten of the $2^{18}=262{,}144$ outer-totalistic Life-like rules and make no claim that the B5/S3 rule chosen here is the only or the best rule with the properties we report.

\section{Method}

\subsection{Rationale}

Neural networks learn by combining non-local linear maps with local nonlinearities; stacked over many layers, these operations produce vast degrees of freedom that give networks their learning power but make their interpretation an intractable bookkeeping task. Other many-element systems are interpretable because their scales separate: in fluid dynamics, thermodynamic laws emerge from the separation of scales between air parcels and their molecules, and describe the motion of each parcel without tracking every molecule within it. Interaction rules that separate scales in the activation maps might likewise make networks interpretable at large scale. The natural separation in a layer is between its input, before the non-local affine map, and its output, after it: to be local, an interaction rule must be computed from the input; to allow learning, it must act on the output. Class~IV CA, which produce complex dynamics in a single-variable system, are natural first candidates for such a rule, though not the only ones. Investigating one such rule, we find, depending on the backbone, a separation of scales in the activation maps across depth, activation maps and input attributions that lie on the object rather than the background, a co-activation graph laid out in the image plane, and higher accuracy under pixel noise, at some cost in clean accuracy. This work stops short of producing the analogue of thermodynamic laws: rules that would describe the network at the large scale without tracking its individual units.

\subsection{Formulation}
\label{sec:formulation-main}

\begin{figure}[t]
\centering
\includegraphics[width=\textwidth]{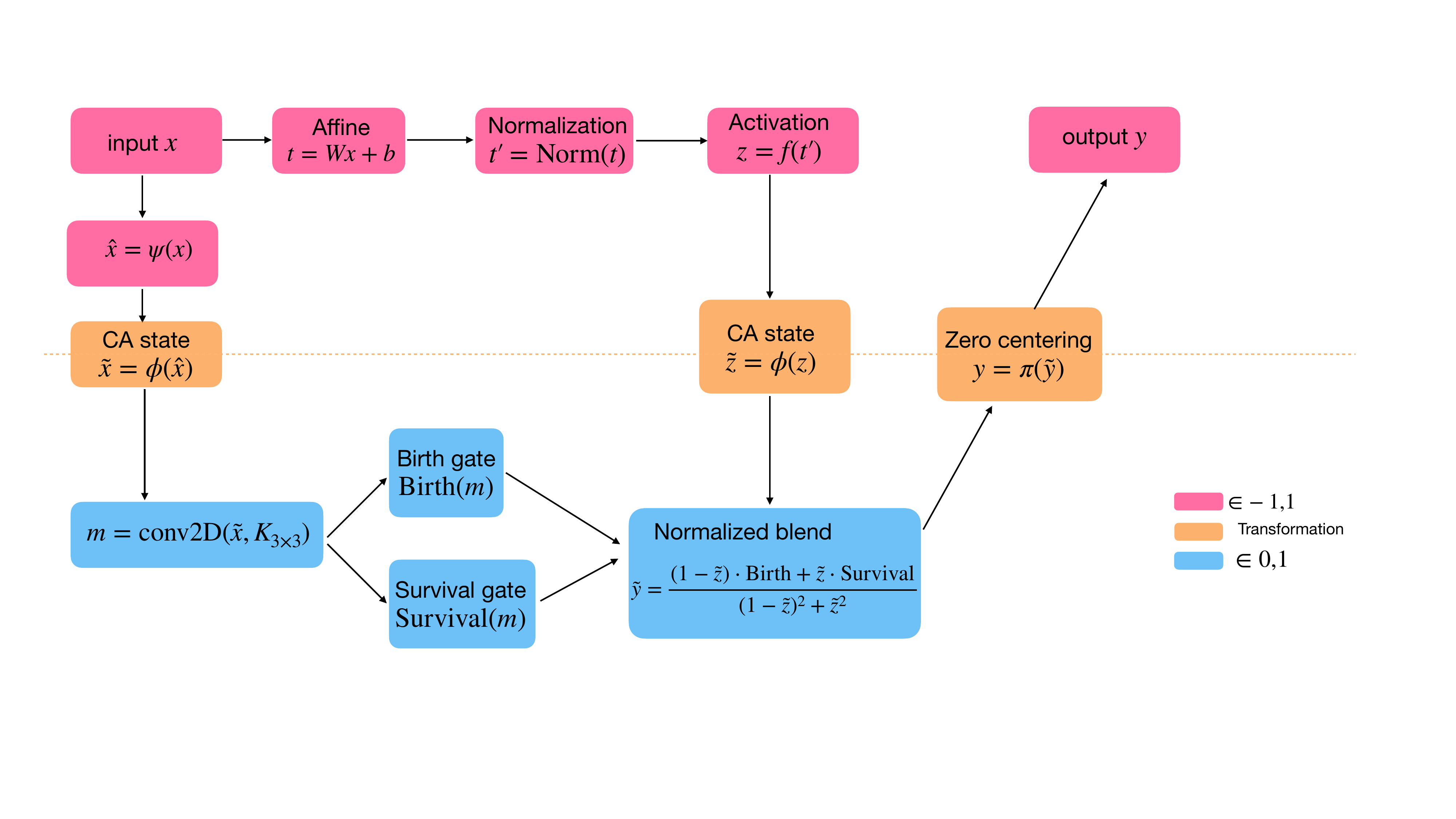}
\caption{\textbf{Overview of the mechanism.} The layer input $x$ feeds both the standard path (affine transformation, normalisation, and bounded activation producing $z$) and the Game-of-Life-inspired lateral inhibition that regulates $z$ into the output $y$ via input-conditioned Birth and Survival gates. Colours mark the value range of each quantity: pink boxes are network states bounded in $[-1,1]$ (input, activation, output), blue boxes are CA states bounded in $[0,1]$ (neighbourhood mean, gates, and their normalised blend), and orange boxes are the transformations that map between the two ranges. The symbols are defined in Section~\ref{sec:formulation-main} and Appendix~\ref{sec:formulation}.}
\label{fig:block_diagram}
\end{figure}

Throughout, a \emph{layer} is any transformation that a network applies to an input $x$ to produce a bounded output $z$: an affine map or convolution followed by its activation, a block of several such operations, or an attention module. Lateral inhibition regulates a layer's output by its own input (Figure~\ref{fig:block_diagram}). The input $x$ drives the standard path---an affine map or convolution $t=Wx+b$ (weights $W$, bias $b$), a normalisation $t'=\mathrm{Norm}(t)$ and the activation $z=f(t')$, $f=\tanh$, so $z\in[-1,1]$---and, in parallel, a fixed CA rule that regulates $z$ into the layer output $y$. Both $x$ and $z$ are mapped to occupancies in $[0,1]$ by $\phi(u)=\tfrac12(u+1)$, giving the CA states $\tilde{x}=\phi(x)$ and $\tilde{z}=\phi(z)$ (a tilde marks a CA state), the states on which CA rules are defined, so a bounded activation is essential. For every unit the rule reads $m=\mathrm{conv2D}(\tilde{x},K_{3\times3})$, the mean input occupancy over its eight Moore neighbours (the kernel $K_{3\times3}$ is $\tfrac18$ off the centre, $0$ at it), and evaluates two smooth gates: $\mathrm{Birth}(m)$, high when $m$ is near a birth count, and $\mathrm{Survival}(m)$, high near a survival count (for B5/S3, $m\approx5/8$ and $3/8$); each gate is a bump of logistic steps of slope $\gamma=15$, and the bumps of several counts combine by De~Morgan's soft-OR. When $x$ and $z$ differ in spatial size, the gate maps, computed on the grid of $x$, are resampled to the grid of $z$ by bilinear interpolation $\psi$ (the identity when the sizes agree). The new CA state $\tilde{y}$ blends the two gates by the unit's own state $\tilde{z}$---Birth for a unit that is off, Survival for one that is on, a mix in between---with a normalisation, the denominator $(1-\tilde{z})^2+\tilde{z}^2$, that keeps the signal from shrinking over many layers; a zero-centring $\pi$ returns it to the activation range as $y=\pi(\tilde{y})$ ($\pi(u)=2u-1$ in the MLP). The gates have no learnable parameters, so training cannot reshape the rule, although every layer after the first can steer the occupancy its gates read through the weights below (Section~\ref{sec:gradients-main}). Appendix~\ref{sec:formulation} gives the equations, the variants used in each network and the gates (Figure~\ref{fig:gate_blend}).

\subsection{Gradients}
\label{sec:gradients-main}

The logistic gates and their blend are smooth; the network as a whole is piecewise differentiable. Three operations are not smooth: the clip that bounds the blend (inactive for \texttt{B5/S3}; Appendix~\ref{app:range}), the CNN's max pooling, and the per-channel image normalisation of the CNN's first layer (Appendix~\ref{sec:channels}), whose statistics are held fixed in the backward pass, so that the input gradient there is a fixed-statistic surrogate (Appendix~\ref{app:perchgrad}). The error that a layer passes to the layer below is the sum of two routes,
\begin{equation}
    \frac{\partial L}{\partial x}\;=\;\underbrace{\Bigl(\frac{\partial L}{\partial x}\Bigr)_{\text{trunk}}}_{x\,\to\,z\,\to\,y}\;+\;\underbrace{\Bigl(\frac{\partial L}{\partial x}\Bigr)_{\text{gate}}}_{x\,\to\,m\,\to\,y}.
    \label{eq:two-routes}
\end{equation}
The trunk route is the usual backward pass through the weights, scaled at every unit by how strongly the output of lateral inhibition responds to the unit's own state ($S-B$ for the plain blend; Appendix~\ref{app:li-backward}). It is the only route to the layer's weights: the rule's fixed parameters never receive a gradient, but its values scale every weight gradient of the layer. The gate route runs through $m$ and contains no weights and no activation derivative; it carries error to the earlier layers through the slopes of the rule, which thereby enter the weight gradients of all layers below. In the trained deep networks it carries nearly all of the weight gradient, so the network learns mainly by steering where the rule acts (Appendix~\ref{app:steering}).

\subsection{Wiring to fully connected, convolutional, and wide-residual networks}
\label{sec:wiring}

Every network uses the same wiring, $y=\LI(x,z)$, with $\tanh$ activations. A channel mismatch between $x$ and $z$ is resolved by a parameter-free channel reduction (Appendix~\ref{sec:channels}), a spatial mismatch by bilinear resampling of the gate maps, computed on the grid of $x$, to the grid of $z$. Maps with a spatial dimension below five get no lateral inhibition.

\paragraph{Multilayer perceptron (MLP).}
An MLP on flattened $28\times28$ grayscale MNIST-family images, with $10$ hidden layers ($2$ in the matched-accuracy experiment, below), $\mathrm{Linear}\to\mathrm{LayerNorm}\to\tanh\to\LI$ per layer, and a linear classifier head ($6.18$M parameters on MNIST and $6.21$M on EMNIST with $10$ layers). Each hidden state is reshaped to the $28\times28$ grid for the neighbourhood, and the first layer's gates read the image. Without downsampling, it isolates the effect of the mechanism on the representation.

\paragraph{Baseline CNN.}
Two convolutional stages (32 and 64 channels, $3\times3$ kernels, no padding, no normalisation) and a linear classifier ($111$k parameters on MNIST, $452$k on EMNIST; Appendix~\ref{app:arch}). Stage~1 regulates $\tanh(\mathrm{conv}_1(\text{image}))$ by the image, through the per-channel image normalisation of Eq.~\eqref{eq:perchannel-gate} for grayscale and colour alike; stage~2 regulates $\mathrm{maxpool}_{2\times2}(\tanh(\mathrm{conv}_2(y_1)))$ by $y_1$, with the channel mean and the gates resampled to the pooled grid. The same network is used for every MNIST-family dataset.

\paragraph{Wide residual network (WRN).}
On CIFAR-10 we use a wide residual network~\citep{Zagoruyko2016} (WRN-40-2, GroupNorm; $2.33$M parameters, $2.54$M on EMNIST; Appendix~\ref{app:arch}). Its stem convolution lifts the image to 16 channels before any lateral inhibition, so every lateral-inhibition layer reads learned features, with the channel mean. Each basic block applies lateral inhibition to its post-activation map, with a parameter-free z-scored pre-state as input; the shortcut stays linear. The classifier reads the flattened final map rather than its spatial average: lateral inhibition drives every channel toward a common spatial mean, so global average pooling would discard the structure it induces (the CNN uses a flatten head for the same reason).

\paragraph{Graph convolutional network (GCN).}
A 2-layer graph convolutional network~\citep{KipfWelling2017} on the pixel graph of MNIST: one node per pixel, joined to its eight Moore neighbours, with $16$ channels per node; each layer computes $\tanh(\hat{A}hW+b)$ with $\hat{A}=D^{-1/2}(A+I)D^{-1/2}$, and a linear classifier reads the flattened final layer ($126$k parameters). Its non-local operation is message passing over the graph rather than a dense or convolutional map, and lateral inhibition is wired as in the other networks: the gates of each layer read the layer input $h$ (the image, in the first layer) and blend the layer output $z=\tanh(\hat{A}hW+b)$ according to its own state, $y=\LI(h,z)$. The neighbourhood mean $m$ is taken over a node's graph neighbours, excluding the node itself: on interior pixels it is the Moore mean of the grid, on border pixels the mean over the neighbours that exist. The graph convolution sums a node and its neighbours under one weight matrix; the gate treats them separately, as a cellular automaton does, reading the neighbours through $m$ and the node through its own state.

\paragraph{Extensibility.}
The same wiring applies to residual and attention backbones, with tokens reshaped to the patch grid and the class token held aside; we do not evaluate them here, as deep residual and attention stacks with this interaction need additional optimisation care.

\subsection{Training recipe, hyperparameters, and initialisation}
\label{sec:hyperparams}

All networks are trained with softmax cross-entropy, without label smoothing, with gradients clipped at norm $5$. The MLP uses Adam~\cite{KingmaBa2015} (learning rate $10^{-3}$, cosine schedule, batch $512$, weight decay $5\times10^{-4}$, $300$ epochs); the CNN Muon~\cite{jordan2024muon} (momentum $0.9$, with AdamW at a tenth of the learning rate on the remaining parameters; constant learning rate $10^{-3}$, weight decay $10^{-3}$, batch $128$, $200$ epochs); the WRN SGD (learning rate $0.1$, momentum $0.9$, weight decay $5\times10^{-4}$, cosine schedule, batch $128$, $400$ epochs on CIFAR-10 and $200$ on EMNIST); the GCN Adam (constant learning rate $10^{-3}$, no weight decay, batch $256$, $200$ epochs). Hidden weights use a uniform initialisation (the GCN: Xavier with the $\tanh$ gain) and the classifier a small-scale one, which gives the most coherent weight maps. Input scaling matters because lateral inhibition reads the image as an occupancy: MNIST-family inputs are normalised with mean and standard deviation $0.5$, so a black background sits at occupancy $0$, below the bands; a colour image is rescaled per channel against that channel's background by the layer that reads it (Appendix~\ref{sec:channels}); CIFAR-10 uses the standard normalisation, harmless because no lateral-inhibition layer of the WRN reads the image. The two networks of a comparison use the same augmentation, which never overlaps the tested corruptions: none for the MNIST-family models (except the matched-accuracy MLP, below), and random crops, flips and AugMix~\cite{Hendrycks2020} for the WRN, whose whole-image operations add no pixel noise (its brightness and contrast operations match two CIFAR-10-C corruptions that we do not evaluate).

\paragraph{Matched-accuracy MLP.} The 2-layer MLP of Figure~\ref{fig:noise_resilience} (first row; $1.24$M parameters on MNIST) is trained so that the networks with and without lateral inhibition are compared at matched clean accuracy: each pair is retrained from the same initialisation until both reach a common target validation accuracy (Appendix~\ref{app:matched}).

\paragraph{Compute.} Most of the method was developed on a laptop (MacBook Pro, 16-inch, 2023, Apple M2 Max). Larger ablations ran on on-demand cloud GPUs, and the WRN and the larger CNN ablations on a single NVIDIA GeForce RTX 5090 (32\,GB).

\paragraph{Corrupted test data.}
Robustness is measured on held-out test images corrupted by Gaussian (sensor and amplifier), shot (photon counting) and impulse (dead pixels, bit errors) noise~\cite{Hendrycks2019}: each image is scaled to $[0,1]$, corrupted, clipped, quantised to 8 bits and normalised as for training, on a ladder of nine severities whose first five are those of the common-corruptions benchmark. We generate the corruptions ourselves because no release covers the range from clean to pure noise on every dataset; on MNIST-C~\cite{MuGilmer2019} and CIFAR-10-C our protocol gives the same results (Appendix~\ref{app:noise}). We report accuracy at every severity, averaged over the three noise types, and the gain with lateral inhibition over severities $1$--$5$, $\Delta_{1\text{--}5}$, and $1$--$9$, $\Delta_{1\text{--}9}$ (\emph{noise-mCA} is the mean over severities $1$--$9$); single noise accuracies use severities $1$--$5$. All networks of a comparison see the same corrupted images.

\section{Rule selection and training behaviour on MNIST}
\label{sec:rulesweep}

\paragraph{Rule search.}
Class~IV Life-like rules are associated with persistent, spatially structured dynamics rather than with collapse or mixing. The network does not run a discrete automaton: the continuous gates reproduce the Game of Life as a standalone automaton only at steep slopes, not at the $\gamma=15$ used here, where each layer applies the rule once (Appendix~\ref{app:gol}). We therefore asked which $(\mathcal{B},\mathcal{S})$ masks give \emph{neural} dynamics that are trainable and spatially coherent. We first chose \texttt{B5/S3} in a 10-epoch screen of Life-like rules, where it was, with Day~\&~Night, among the most coherent rules that still trained well. Part of the choice was qualitative: as CA rules have been classified largely by looking at their patterns~\cite{Wolfram1984}, we favoured rules whose activation maps looked like class~IV dynamics, structure that persists from layer to layer without dying out or flooding the field. Day~\&~Night does not look so: in the trained networks it grows the digit into a saturated region that fills most of the field, whereas \texttt{B5/S3} keeps the shape of the digit through all ten layers. Yet on the summary measures below Day~\&~Night is close to \texttt{B5/S3}, as smooth ($0.95$ against $0.94$) and nearly as simple ($0.37$ against $0.33$), because a field flooded to a uniform value is smooth and compressible without carrying the shape of the input; looking at the maps separated the two. To test the choice at full length we trained the MNIST MLP with ten rules, five seeds each (paper recipe, Section~\ref{sec:hyperparams}): \texttt{B5/S3}; its neighbour \texttt{B5/S23}; the symmetric \texttt{B3/S3}; the Game of Life \texttt{B3/S23}; HighLife \texttt{B36/S23}; Day~\&~Night \texttt{B3678/S34678}; Seeds \texttt{B2/S}; the Replicator \texttt{B1357/S1357}; 3-4~Life \texttt{B34/S34}; and Walled~Cities \texttt{B45678/S2345}.

\begin{figure}[t]
\centering
\includegraphics[width=\textwidth]{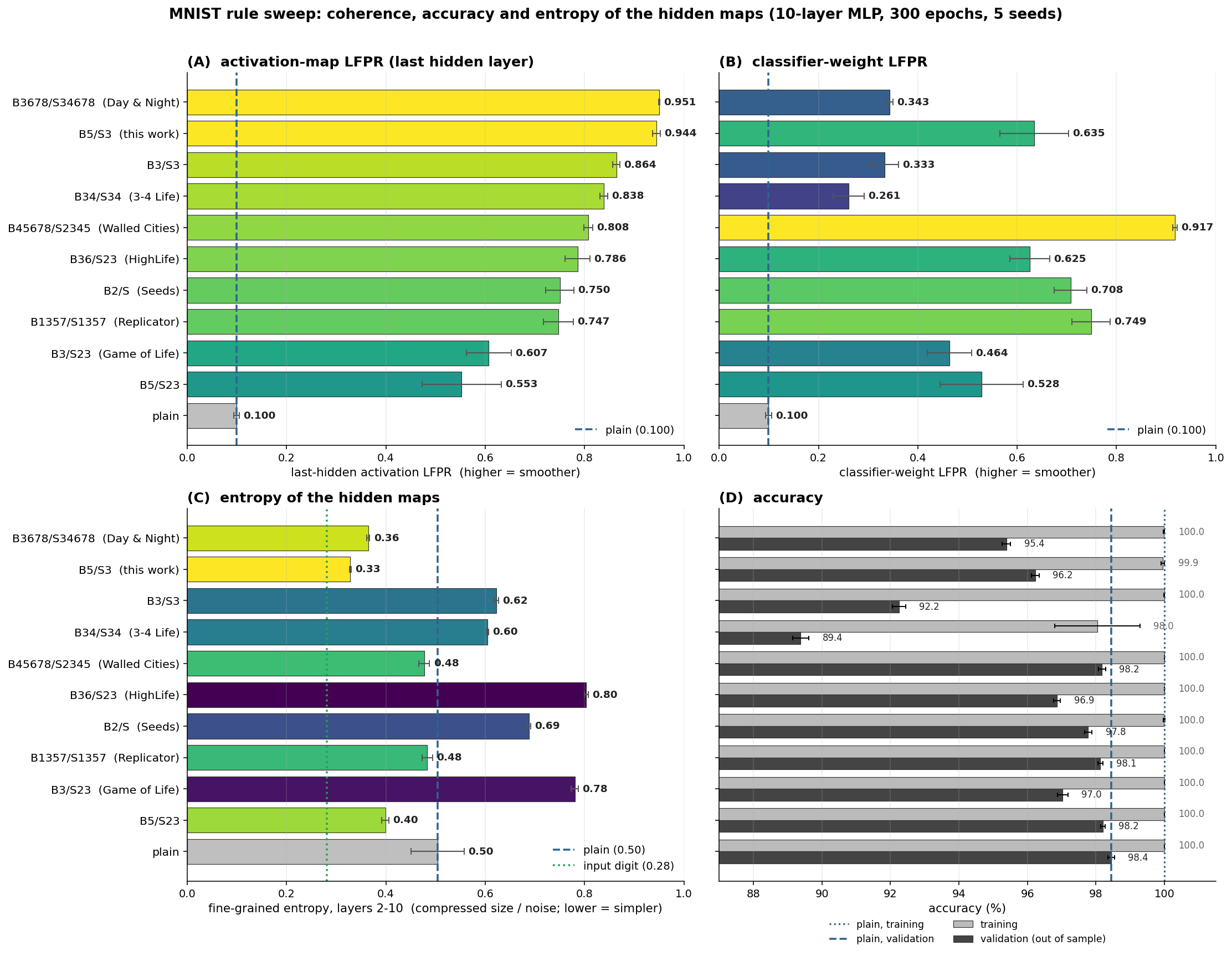}
\caption{\textbf{MNIST rule sweep.} Class-mean LFPR (Eq.~\eqref{eq:lfpr}) of (A)~the class-averaged final hidden maps and (B)~the per-class classifier weights; (C)~fine-grained entropy of the hidden maps (layers $2$--$10$, $500$ test images; compressed size at $8$ levels relative to a random map; dotted: the input digit); (D)~training accuracy (first $10{,}000$ training images) and validation accuracy. Ten rules and the plain network (grey row, dashed lines); 10-layer MLP, paper recipe, mean $\pm$ s.d.\ over five seeds; rows sorted by panel~A.}
\label{fig:mnist_rule_sweep_metrics}
\end{figure}

We measure how smooth---rather than salt-and-pepper---a map is by its \emph{low-frequency power ratio} (LFPR). For a 2-D map $X$ on the $28\times28$ grid we remove its mean, $\tilde{X}=X-\tfrac{1}{HW}\sum_{i,j}X_{ij}$, take the shifted power spectrum $P=\bigl|\mathrm{fftshift}(\mathrm{FFT2}(\tilde{X}))\bigr|^2$, let $r(i,j)\in[0,1]$ be the radial frequency normalised by its maximum, and, for a cutoff $\rho$,
\begin{equation}
    \mathrm{LFPR}(X;\rho)=\frac{\sum_{i,j:\,r(i,j)\le\rho} P_{ij}}{\sum_{i,j} P_{ij}}\;\in[0,1],
    \label{eq:lfpr}
\end{equation}
so $\mathrm{LFPR}\to1$ for smooth maps and $\to0$ for salt-and-pepper ones. We use $\rho=0.25$ and average over classes, for the per-class classifier weights and for the class-averaged final hidden maps (Figure~\ref{fig:mnist_rule_sweep_metrics}). After full training every rule makes the network far more coherent than the plain network (Figure~\ref{fig:mnist_rule_sweep_metrics}A,B), and every rule fits the training set, most at a validation accuracy close to the plain network's (panel~D). Spatial coherence alone does not single out \texttt{B5/S3}: Day~\&~Night's activation maps are as smooth, and several rules have smoother classifier weights. The entropy of the hidden maps does (panel~C): measured as in \citet{Aaronson2014}, by their compressed size relative to noise, \texttt{B5/S3} keeps them nearly as simple as the input digit, the lowest of all rules, while most other rules leave them as busy as the plain network's or busier. Early coherence does not predict final coherence: in the 10-epoch screen the Replicator sat at the plain network's floor. What distinguishes \texttt{B5/S3} is coherence in depth and across independently trained networks (Section~\ref{sec:scalesep}), which is why we keep it as the default.

\section{Results}

The Game of Life can be read as performing three functions: (i)~\emph{removing noise}---a live cell whose Survival rule is unmet dies; (ii)~\emph{isolating important features}---a live cell whose Survival rule is met persists; and (iii)~\emph{joining fractured features}---a dead cell whose Birth rule is met comes alive. We first examine the organisation that lateral inhibition induces (functions ii and iii), then its effect on input noise (function i). The MLP is a transparent probe of the rule inside a learned network, not a competitive model: with every hidden unit on the input grid, the induced structure is directly visible. In the convolutional and residual networks some of this structure fades into the architecture, while noise resilience emerges as the substantive benefit.

\paragraph{Coherence in space, time and across the ensemble.}
For the fully connected network we read the stack of hidden layers as a field, as in the analysis of fluid flows: each hidden map is space, depth is time, and independently trained networks (seeds) are an ensemble of realisations; a coherent structure persists along an axis rather than decorrelating at once. Lateral inhibition produces coherence along all three axes: smooth maps in space (Section~\ref{sec:coherence}), maps that stay consistent from layer to layer and carry a slow, large-scale component through depth (Section~\ref{sec:scalesep}), and networks that converge on the same maps, unit for unit (Figure~\ref{fig:seed_agreement}). The plain network shows none of the three.

\subsection{Activation coherence}
\label{sec:coherence}

\begin{figure}[t]
\centering
\includegraphics[width=\textwidth]{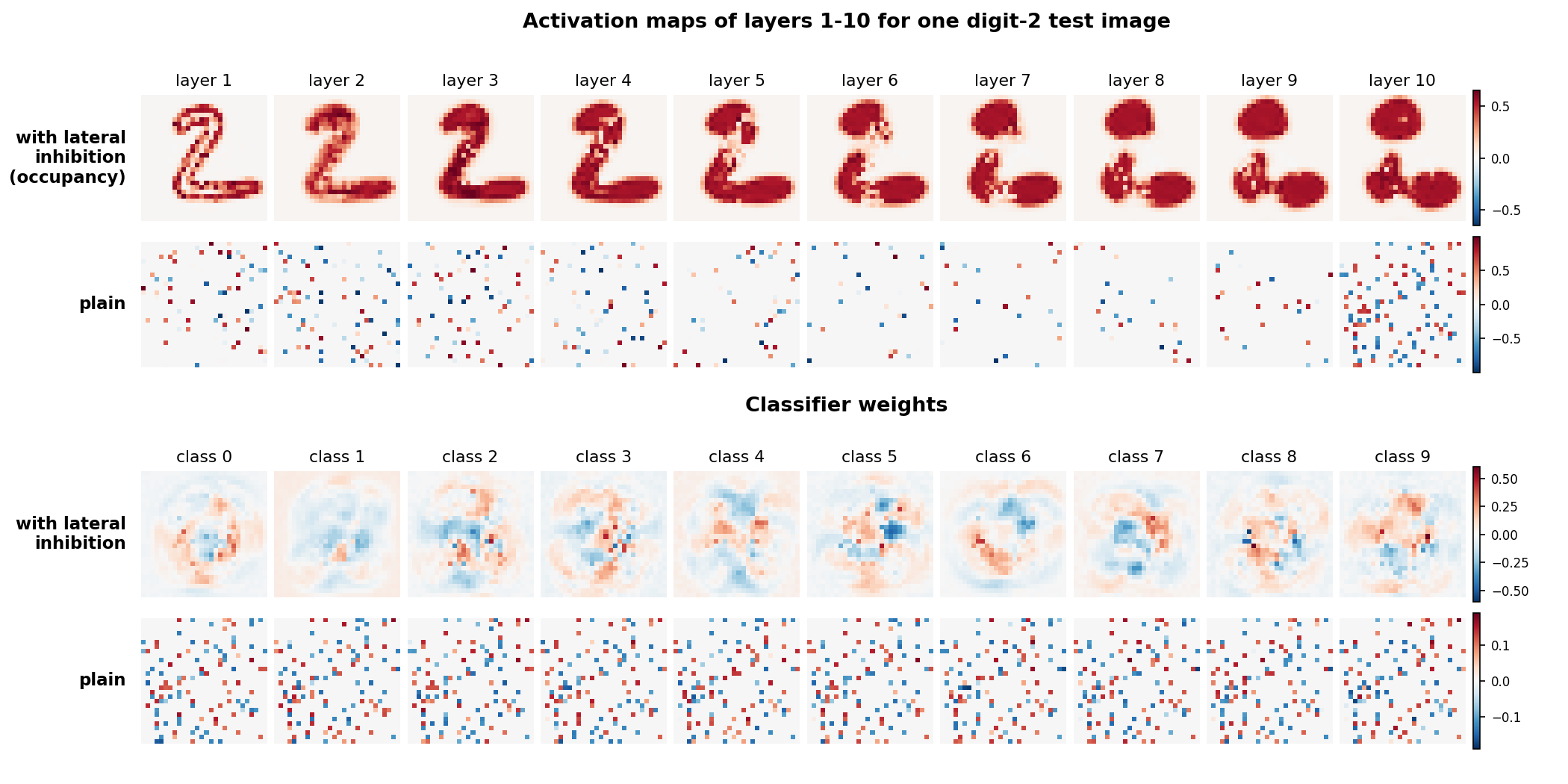}
\caption{\textbf{Activation maps across depth and classifier weights (MNIST MLP).} Top: the activation maps of hidden layers $1$--$10$ for one digit-2 test image, with lateral inhibition (shown as the occupancy $(y+1)/2$, so the background at $y=-1$ is white) and plain. Bottom: the classifier weight map of each class, with lateral inhibition and plain. One signed colour scale per row (white is zero). Lateral inhibition yields smooth, digit-shaped structure; the plain network is high-frequency and diffuse.}
\label{fig:act_maps}
\end{figure}

Figure~\ref{fig:act_maps} shows the activation maps of the ten hidden layers of the MNIST MLP for one test digit, and the classifier weight map of each class. With lateral inhibition the maps are smooth and contiguous at every layer, digit-shaped in the first layers and coarsening into a few blobs with depth, with only a minority of the grid active; in the plain network they are a speckle over the whole grid with no trace of the digit (a video for all ten digits is at \href{https://doi.org/10.6084/m9.figshare.34750569}{doi.org/10.6084/m9.figshare.34750569}). Over five networks the activation-map LFPR rises from $0.09\pm0.01$ (plain) to $0.96\pm0.02$ with lateral inhibition, and the classifier-weight LFPR from $0.10\pm0.01$ to $0.70\pm0.10$. The entropy and apparent complexity that \citet{Aaronson2014} defined for a CA of coffee mixing with cream tell the same story: treating depth as time, the plain network ``mixes'' in its first layer, whereas with lateral inhibition both stay close to those of the input digit through all ten layers. Controls separating the role of the input-aligned wiring from that of the rule are collected in Appendix~\ref{app:controls}: rule-free blurs of the layer input and output (Appendix~\ref{app:frame}), the structure of the co-activation graph (Appendices~\ref{app:gridrule}, \ref{app:collapse} and~\ref{app:percolation}), and the coherence already present at initialisation (Appendix~\ref{app:init}).

\subsection{The co-activation graph}
\label{sec:netsci}

\begin{figure}[t]
\centering
\includegraphics[width=\textwidth]{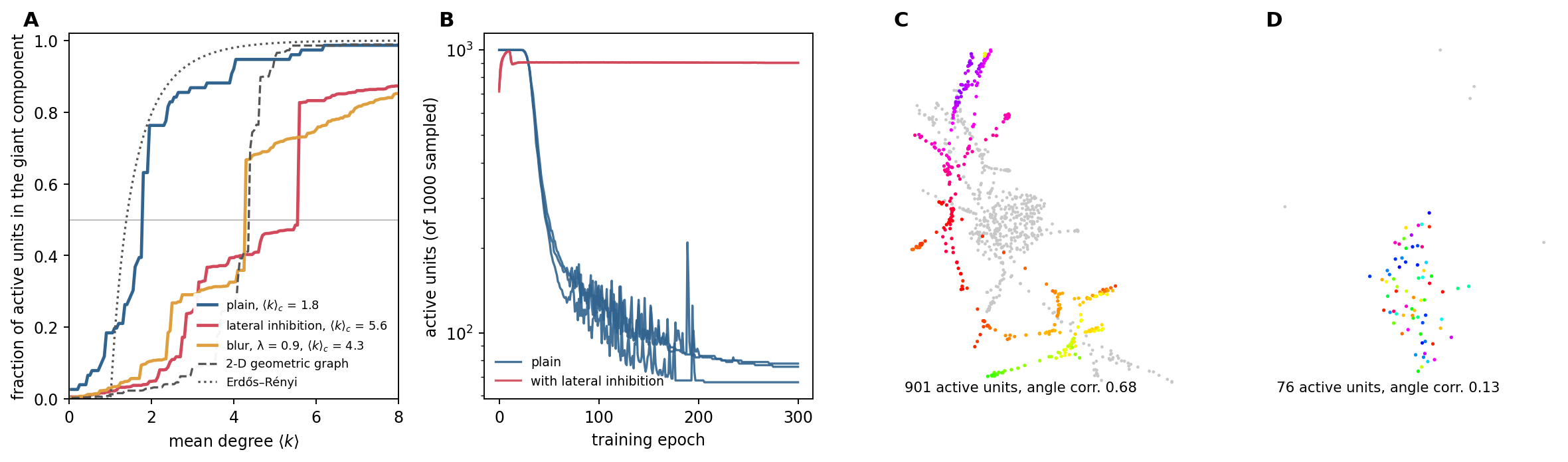}
\caption{\textbf{The co-activation graph of the MNIST MLP.} (A)~Fraction of the active hidden units in the giant component as links are added strongest first, against the mean degree $\langle k\rangle$, at the end of training: plain (blue), with lateral inhibition (red) and with the input blur of Appendix~\ref{app:frame} at $\lambda=0.9$ (orange), next to a two-dimensional random geometric graph (dashed) and an Erd\H{o}s--R\'enyi graph (dotted) of the same size; the grey line marks half of the units, which defines $\langle k\rangle_c$. (B)~Active hidden units, of $1000$ sampled, over training, three networks each. (C,~D)~Force-directed layout of the graph at $\langle k\rangle=4$, with lateral inhibition and plain; units in the giant component are coloured by the angle of their pixel about the image centre, the others are grey. With lateral inhibition the layout, computed from the links alone, recovers the arrangement of the image. One network ($\langle k\rangle_c$ and layouts) or three (B); positive correlations over $1000$ test images.}
\label{fig:netsci}
\end{figure}

Viewed from the perspective of network science, lateral inhibition turns the hidden units of the MLP into a spatial graph laid out on the input grid (Figure~\ref{fig:netsci}; Appendix~\ref{app:percolation}, with videos of the graph forming). We join pairs of hidden units by the positive correlation of their activations over $1000$ test images, counting only the units that respond to the input, and add the links strongest first, as in studies of percolation in living neural networks~\citep{Breskin2006}, following the size of the giant component, the largest set of units joined by chains of links. We plot it against the mean degree $\langle k\rangle$, the number of links per unit, so that networks with different levels of correlation are compared at the same number of links. The reference is the Erd\H{o}s--R\'enyi random graph~\citep{ErdosRenyi1960}, in which links join units at random: almost every link then reaches a unit not yet connected, a giant component appears at $\langle k\rangle=1$ and covers half of the units at $\langle k\rangle\approx1.4$. Structured graphs spend links inside groups that are already connected and percolate later. With lateral inhibition the giant component forms late, covering half of the units at $\langle k\rangle_c=4.2$--$6.0$ (three networks), against $1.7$--$2.1$ in the plain MLP, and it grows in steps, single links joining large groups of units (largest jump $0.29$--$0.36$ of the units, against $0.16$--$0.19$). This is not a weaker connectivity: with lateral inhibition each unit has about $28$ partners correlated above $0.5$, against about $3$ in the plain MLP. It reflects where the links are: $68$--$70\%$ of these strongly correlated pairs are units within two pixels of each other, against $2\%$ in the plain MLP, so the giant component can only grow by joining one region of the image to the next. A two-dimensional random geometric graph~\citep{Dall2002}, whose links join nearby points of a plane, percolates in the same way ($\langle k\rangle_c=4.4$, largest jump $0.29$; Figure~\ref{fig:netsci}A), and a force-directed layout of the graph, which sees only the links, recovers the layout of the image (Figure~\ref{fig:netsci}C; angle correlation $0.59$--$0.78$, chance $0.05$). The plain MLP's curve lies close to that of a random graph: its strongest correlations join units across the whole network, with no image geometry (Figure~\ref{fig:netsci}D), although its graph is not random (its clustering is about six times that of a random graph). By the end of training only $7$--$10\%$ of its hidden units still respond to the input; the others stop, in every network, from about epoch $25$ on (Figure~\ref{fig:netsci}B). Neither graph is scale-free, and the graph with lateral inhibition is a lattice rather than a small world: its clustering is $90$--$160$ times that of a random graph, but its paths are $2.6$--$2.9$ times longer. In the CNN the shift is smaller ($\langle k\rangle_c=3.5$--$4.1$ against $2.8$--$2.9$) and almost all units stay active in both networks. The input blur of Appendix~\ref{app:frame}, the simplest rule by which the input regulates the output, reproduces the late threshold, the steps and the geometry, so these properties follow from the input regulating the output through a local rule, not from the B5/S3 rule.

\subsection{Scale separation and consistency across depth}
\label{sec:scalesep}

\begin{figure}[t]
\centering
\includegraphics[width=\textwidth]{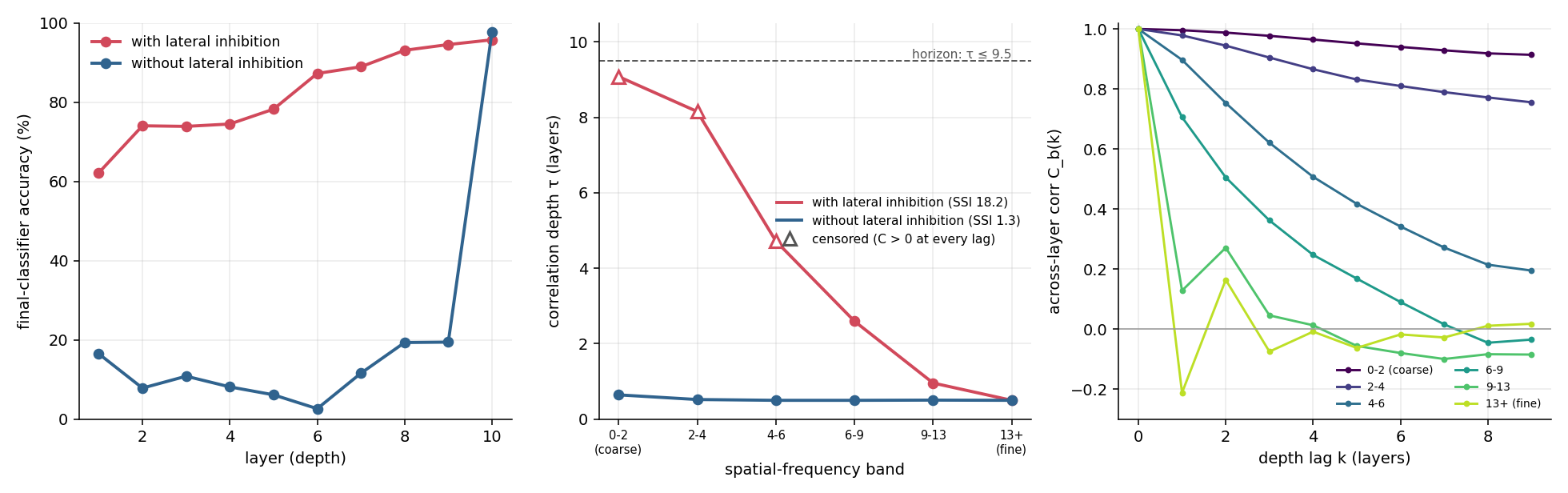}
\caption{\textbf{Depth diagnostics (MNIST MLP, one network).} \emph{Left:} accuracy of the trained final classifier applied to each hidden layer. \emph{Middle:} correlation depth $\tau_b$ of each spatial-frequency band, the number of layers over which its maps stay correlated (Appendix~\ref{app:ssi}); SSI $18.2$ with lateral inhibition and $1.3$ plain for this network ($18.2\pm0.1$ and $1.1\pm0.1$ over five networks); open triangles: censored (still correlated at the largest lag, $9$, so $\tau_b$ is a lower bound); dashed: the largest value ten layers allow. \emph{Right:} correlation $C_b(k)$ between the band-$b$ components of two activation maps $k$ layers apart, averaged over all such pairs of layers, with lateral inhibition. First $1000$ test images.}
\label{fig:depth_diag}
\end{figure}

Two diagnostics (Figure~\ref{fig:depth_diag}) show that with lateral inhibition the representation is laid down coarsely and early and refined with depth. First, we apply the trained final classifier to every hidden layer (a ``logit lens''). This tests consistency in time, not the information a layer holds: it succeeds only if the layer already lies in the coordinates the classifier reads, whereas a classifier fitted to each layer could find class information in coordinates that change from layer to layer. With lateral inhibition the final classifier reads the first hidden layer at $57.5\pm4.0\%$, rising monotonically to $96.5\pm0.3\%$ at the last (five networks); the layers of the plain network, each in coordinates of its own, are read near chance until the final layers. Second, treating depth as time, we band-pass each layer's map by spatial frequency and measure the correlation $C_b(k)$ between maps $k$ layers apart and the \emph{correlation depth} $\tau_b$ it implies, the number of layers over which band $b$ stays correlated, the analogue of a correlation time with depth in place of time (Appendix~\ref{app:ssi}). With lateral inhibition the coarsest band is still correlated at $0.92\pm0.01$ after nine layers, the largest lag ten layers allow, while the finest decorrelates within one layer ($C=-0.17\pm0.03$ at lag~1); in the plain network every band decorrelates at once ($|C|<0.08$ at every lag). The ratio of the two correlation depths, the scale-separation index (SSI), is $18.2\pm0.1$ with lateral inhibition and $1.1\pm0.1$ without it (plain MLP). Ten layers cap it at $19$ and censor the coarse band's correlation depth, so the SSI records that the coarse maps persist through the whole stack while the fine maps change at every layer, not an intrinsic ratio of timescales: a representational counterpart of the separation of scales that makes other many-element systems interpretable.

\begin{figure}[t]
\centering
\includegraphics[width=\textwidth]{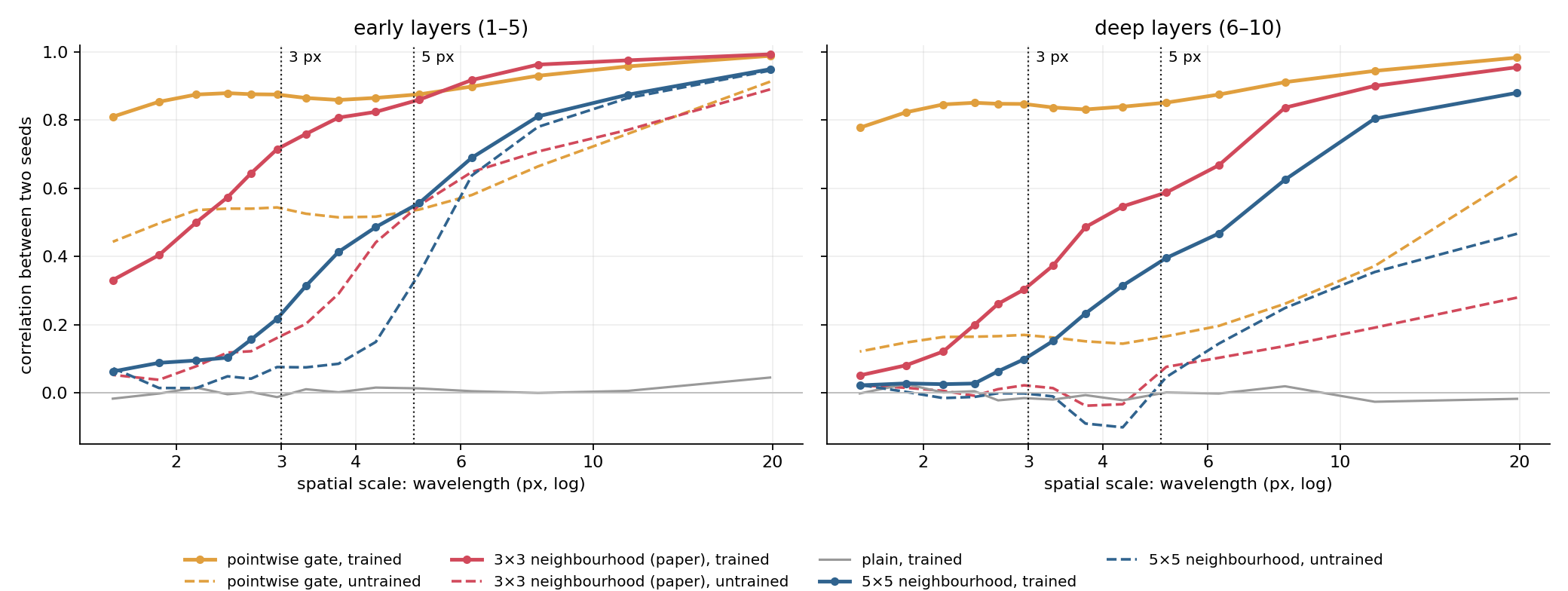}
\caption{\textbf{What part of the agreement between independently trained networks is learned, and at which scales (MNIST MLP).} Unit-wise correlation between the band-passed hidden maps of two networks trained from different seeds, against wavelength; mean over three seed pairs and over the early (layers~1--5, left) or deep (6--10, right) hidden layers. Solid: trained; dashed: the same seeds untrained; colour: the neighbourhood (pointwise, $3\times3$ as in the paper, $5\times5$); grey: plain.}
\label{fig:seed_agreement}
\end{figure}

\paragraph{Coherence across the ensemble: what is learned and what is inherited.}
Independently trained networks with lateral inhibition converge to nearly the same hidden maps, unit for unit: across five seeds the MNIST MLP's maps correlate at $0.96$ in the first hidden layer and $0.82$ in the last ($1000$ test images). The plain network shows no unit-wise agreement ($\approx0$), although its representations are similar as a whole (linear CKA $0.89$--$0.97$): each seed spreads the same information over its units in its own basis. To separate what the architecture provides from what training adds, we compare, per spatial-frequency band, the agreement between two trained networks with that between the same seeds' untrained initialisations (Figure~\ref{fig:seed_agreement}). In the early layers much of the coarse agreement (wavelengths above ${\approx}6$~px) is inherited: lateral inhibition imposes the input's layout before any learning. This inheritance fades with depth, and in the deep layers almost all of the agreement is learned, from the scale of the digit down to a few pixels. The finest scale of agreement is set by the neighbourhood: with the paper's $3\times3$ neighbourhood, deep-layer agreement falls from $0.49$ at a wavelength of $3.7$~px to $0.05$ at $1.6$~px; with $5\times5$ the fall-off moves to coarser scales ($0.23$ and $0.02$); with a pointwise neighbourhood agreement extends to single pixels ($0.78$--$0.98$). As along depth, the large scales are coherent across the ensemble and the small scales are not. In the language of Reynolds averaging~\citep{Pope2000}, the network with lateral inhibition admits a shared mean field: the depth average of one network's deep layers matches the ensemble average as closely as two halves of the ensemble match each other ($r=0.91$ vs $0.90$), with fluctuations below $10\%$ of the variance; the plain network does not. Independently trained networks can therefore be averaged. Averaging the weights of five plain MLPs gives a network at chance ($10.1\%$; $50.3\%$ for three), because their units do not correspond; with lateral inhibition, whose units are anchored to the same pixels in every seed, the five-seed average still classifies $88.8\%$ of the test set (individual networks $96.1\%$).

\begin{figure}[t]
\centering
\includegraphics[width=\textwidth]{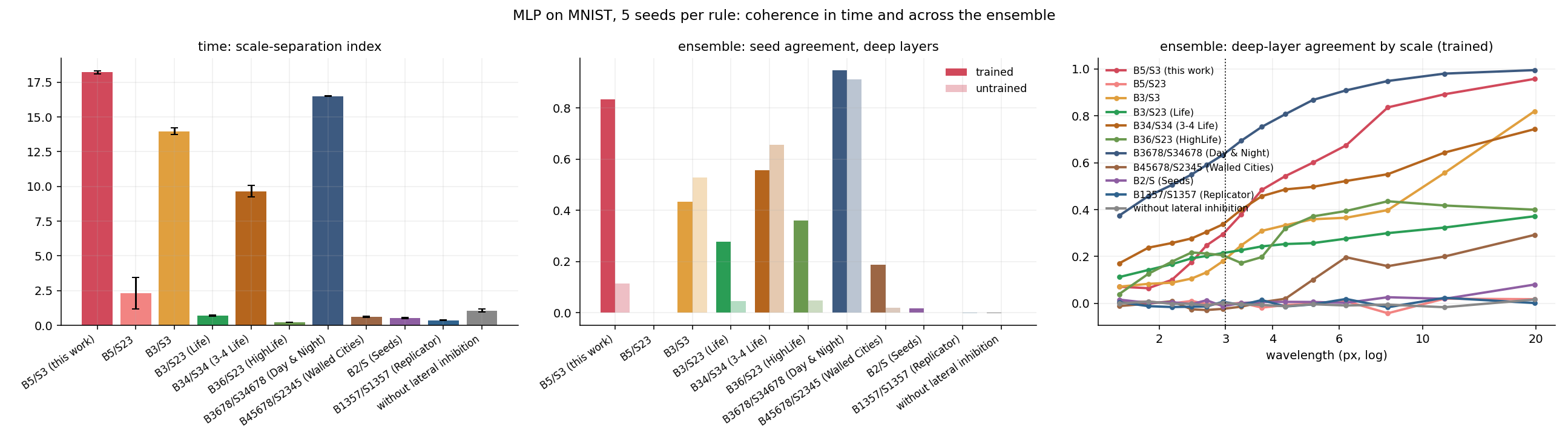}
\caption{\textbf{Coherence in time and across the ensemble for ten rules (MNIST MLP, five seeds per rule, paper recipe).} From left: coherence in depth (scale-separation index); agreement between independently trained networks in the deep layers (layers~6--10), trained and untrained; deep-layer agreement of trained networks by spatial scale. Coherence in space, the LFPR of the class-averaged final hidden maps, is panel~A of Figure~\ref{fig:mnist_rule_sweep_metrics}. Most rules give spatially smooth maps, but only \texttt{B5/S3} is coherent along all three axes. Error bars: s.d.\ over seeds.}
\label{fig:rule_coherence}
\end{figure}

\paragraph{Coherence across rules.}
Is coherence along the three axes a property of input regulation in general, or of the rule? For each of the ten rules of the sweep (five seeds each) we measured coherence in space (Figure~\ref{fig:mnist_rule_sweep_metrics}A), in time (the SSI and the final classifier's accuracy at the first hidden layer) and across the ensemble (deep-layer agreement between seeds and the accuracy of the five-seed weight average; Figure~\ref{fig:rule_coherence}). Spatial coherence is largely generic (activation LFPR $0.54$--$0.95$ for most rules, $0.09$ plain); coherence in time and across the ensemble is not. Only with \texttt{B5/S3} does the final classifier read the first hidden layer ($57.5\%$, against $10$--$17\%$ for every other rule and plain), and it has the highest SSI ($18.2$), followed by Day~\&~Night ($16.5$), \texttt{B3/S3} ($14.0$) and 3-4~Life ($9.7$), all others below $2.5$. \texttt{B5/S3} seeds agree in the deep layers at $0.83$ (untrained $0.11$), and their weight average classifies $88.8\%$ of the test set. Day~\&~Night seeds agree more ($0.95$; weight average $75.1\%$), but already at $0.91$ untrained: for it and the rules that come closest (\texttt{B3/S3}, 3-4~Life) the agreement is imposed by the rule at initialisation rather than learned ($0.53$--$0.91$ untrained, $0.43$--$0.95$ trained). The remaining rules agree at $0.00$--$0.36$, and their weight averages fail ($7.5$--$47.9\%$). The neighbouring rule \texttt{B5/S23} is the most accurate rule ($98.2\%$ validation) but incoherent in time (SSI $2.3$) and across the ensemble (agreement $0.00$). Smooth maps follow from input regulation with almost any rule; coherence along all three axes is specific to the rule.

\paragraph{Reproducibility from nearly identical initialisations.}
A stricter test of reproducibility than agreement between seeds asks how far two networks drift apart when they start from almost the same weights. \citet{LyGong2026} showed that the basins of attraction of training can be riddled: two initialisations that differ by $\varepsilon$ in every parameter end in networks that disagree on a fraction of the test images, the \emph{churn}, that stops shrinking as $\varepsilon\to0$, so that no precision of the initialisation makes the trained function reproducible. We applied their test to the MNIST CNN, with pairs of networks perturbed from one reference initialisation by $\varepsilon=10^{-1}$ to $10^{-8}$ and trained deterministically on the same batches (Appendix~\ref{app:riddling}). With half of the training labels replaced by a wrong class, the setting in which they observe riddling, the plain CNN's churn levels off at small $\varepsilon$ at the level they report for riddled training, whereas with lateral inhibition it levels off about thirty times lower, at a higher test accuracy ($94.7\%$ against $91.7\%$; Figure~\ref{fig:riddling}A). With clean labels neither network approaches riddling, and lateral inhibition still lowers the churn several-fold (Figure~\ref{fig:riddling}B).

\subsection{Interpretability: attribution localises on the object}
\label{sec:saliency}

\begin{figure}[t]
\centering
\includegraphics[width=\textwidth]{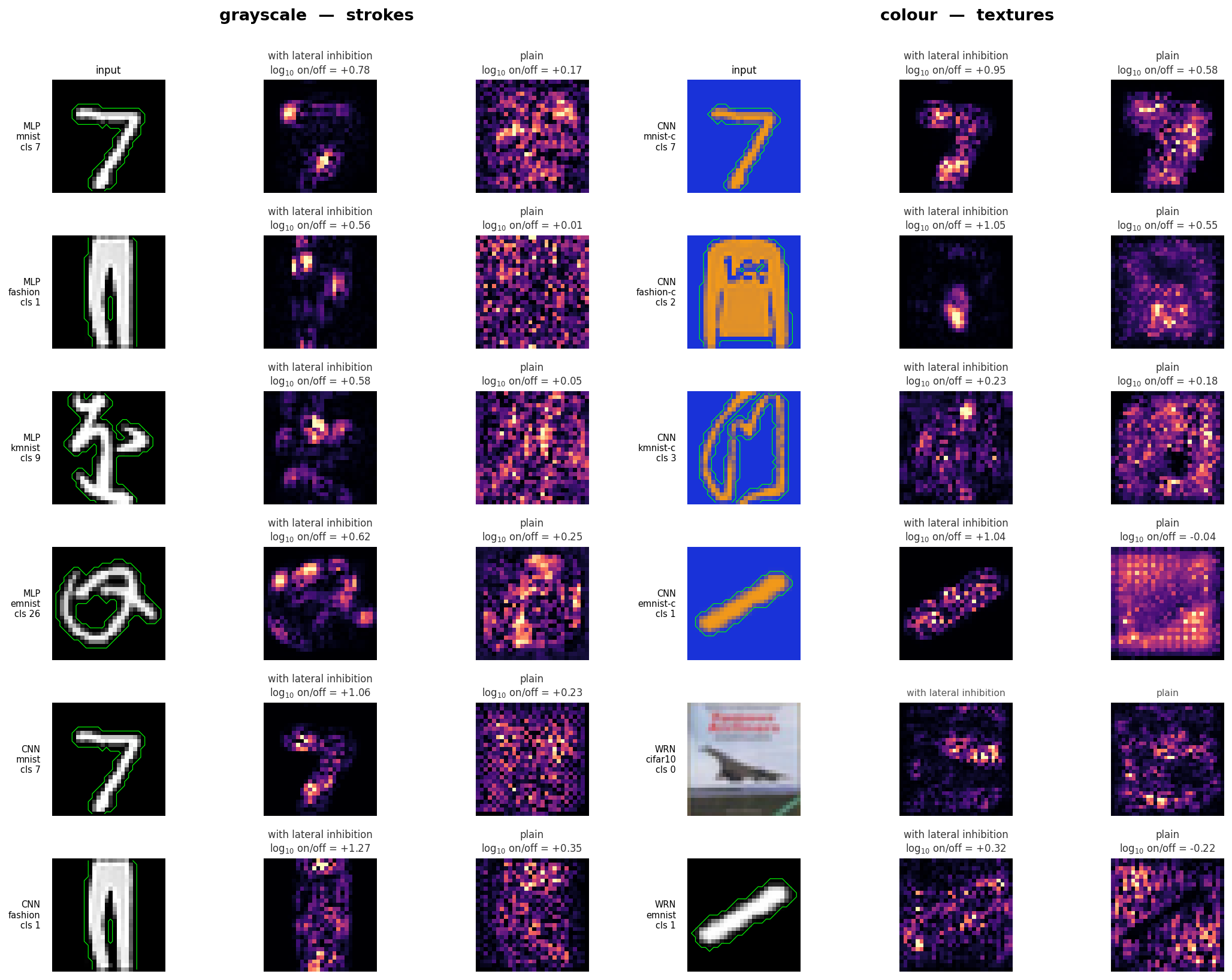}
\caption{\textbf{Gradient saliency on and off the object.} Saliency $|\partial\,\mathrm{logit}_y/\partial x|$ of the true class for representative samples: grayscale (left: MLP and CNN on the MNIST family) and colour (right: the CNN on the colour MNIST family; the WRN on CIFAR-10 and grayscale EMNIST). Each block shows the input with its dilated object contour and the saliency with and without lateral inhibition, annotated with the sample's log-ratio $\log_{10}(\overline{s}_\text{on}/\overline{s}_\text{off})$ ($>0$: object; omitted for CIFAR-10, which has no object mask). For the MLP the localised gradient comes from the gate route rather than from reliance on the object.}
\label{fig:saliency}
\end{figure}

The organisation that lateral inhibition imposes also makes the network's decisions legible. We ask where the gradient saliency of the true class, $s=|\partial\,\mathrm{logit}_y/\partial x|$, falls relative to the object, and score each image by the log-ratio $\log_{10}(\overline{s}_\text{on}/\overline{s}_\text{off})$ of saliency per pixel on and off an object mask, positive when the object dominates (protocol and further scores in Appendix~\ref{app:saliency}). Every trained network with lateral inhibition favours the object (Figures~\ref{fig:saliency} and~\ref{fig:saliency_dist}): on the CNN, grayscale or colour, saliency per pixel is $4$--$14$ times higher on the object than off it, in most images. The plain CNN localises much less, and on EMNIST it favours the background. The MLP's maps also favour the object, for a different reason (below). Correctly classified and misclassified images, and the saliency of the predicted class, give the same picture (Figure~\ref{fig:saliency_dist}). The colour result depends on the per-channel image normalisation (Appendix~\ref{sec:channels}): with the channel mean, the colour CNN localises worse than plain. On the WRN the effect is weak (EMNIST) or cannot be scored (CIFAR-10 has no object mask), consistent with its stem convolution: inserting such a stem in front of the CNN removes its localisation too.

\paragraph{Is the localisation learned reliance?}
Gradient saliency measures sensitivity, not reliance, and part of it comes from lateral inhibition itself: its parameter-free route from output back to input is, at the layer that reads the image, largest where the occupancy sits near a band edge, that is, near the strokes (Appendix~\ref{app:li-backward}). For the CNN the saliency is moreover a surrogate, its image statistics being held fixed in the backward pass; differentiating through them changes the magnitudes of the maps (median relative difference $0.48$--$0.75$) but hardly their ranking (median per-image rank correlation $0.95$--$0.97$), and the localisation holds (log-ratio $+0.54$ to $+0.95$, $91$--$100\%$ of images; Appendix~\ref{app:perchgrad}). The principal causal evidence is therefore a gradient-free test, with the maps as corroboration (MNIST, five seeds, $1000$ test images; Appendix~\ref{app:saliency}, Table~\ref{tab:salchecks}). Replacing as many object pixels as the digit has with uniform noise lowers the accuracy of the CNN with lateral inhibition to $60.5\%$, whereas noising as many background pixels leaves it at $96.3\%$; the plain CNN is the reverse ($74.4\%$ and $67.0\%$). With every gate route detached, so that only the learned trunk contributes, the CNN's log-ratio rises from $+1.13$ to $+1.28$: the localisation is carried by the trunk. Untrained networks are already partly localised ($+0.89$ with lateral inhibition, $+1.09$ without); training sharpens this with lateral inhibition and weakens it without ($+0.47$). The CNN with lateral inhibition places $1\%$ of its saliency on the background at least three pixels from the digit (plain: $13\%$) and is unaffected by noise there ($98.2\%$, plain $86.2\%$). For the CNN, lateral inhibition thus yields attribution maps that can be read as ``the network looked here''.

The MLP does not pass these checks. With every gate route detached its input gradient vanishes ($4\times10^{-13}$ per pixel against $4\times10^{-2}$), so its saliency is entirely the gate route, and it does not rely mainly on the object: noising the background costs more accuracy than noising as many object pixels ($79.8\%$ against $85.3\%$). Its saliency maps reflect the sensitivity built into lateral inhibition, not what the network relies on.

\begin{figure}[t]
\centering
\includegraphics[width=\textwidth]{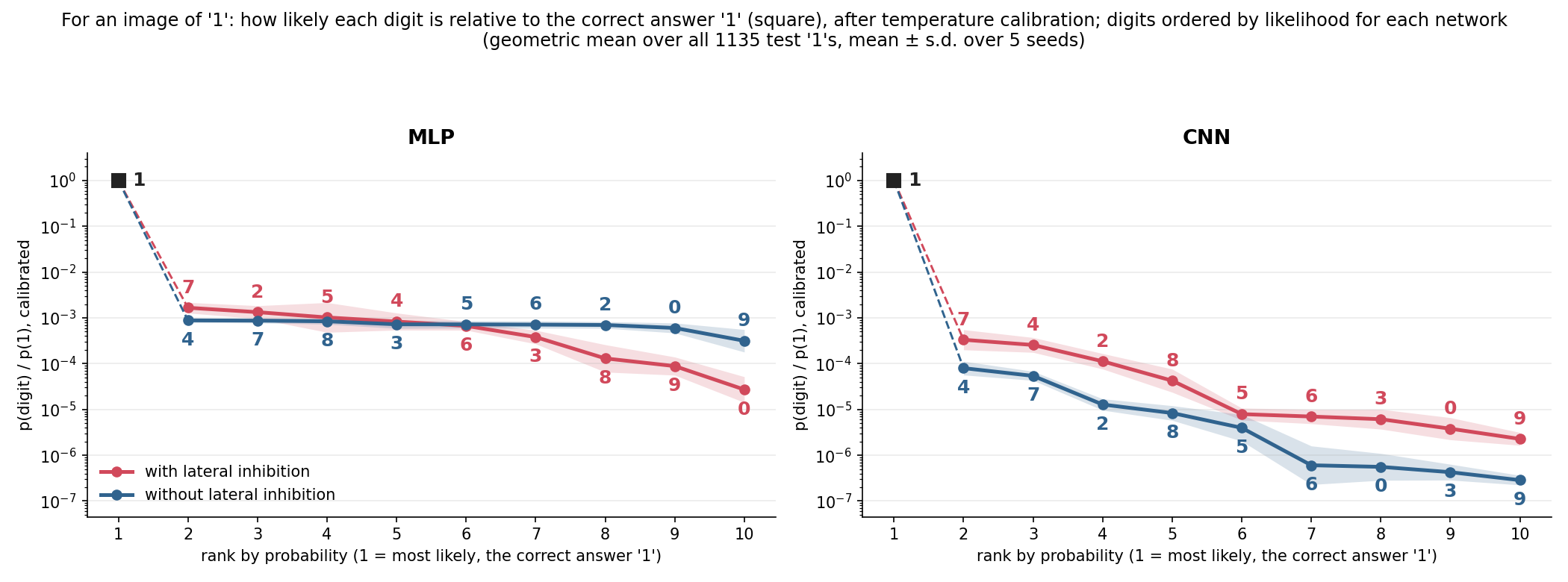}
\caption{\textbf{Alternatives to a ``1''.} MLP (left) and CNN (right), with (red) and without (blue) lateral inhibition: the probability of every other digit relative to ``1'' (square), as the geometric mean of $p(d\,|\,x)/p(1\,|\,x)$ over the $1135$ MNIST test images of ``1'', after temperature scaling; digits ranked by probability and labelled beside their curve; band: s.d.\ over five seeds. The plain MLP's alternatives are nearly flat (Appendix~\ref{app:ranking}).}
\label{fig:perdigit}
\end{figure}

We also asked how the networks rank the alternatives to the correct digit. For an image $x$ of digit $s$, $\log[p(d\,|\,x)/p(s\,|\,x)]=z_d-z_s$, so the ranking of alternatives is a ranking of logit gaps. Their size depends on the scale of the logits, so we compare the networks after temperature scaling~\citep{Guo2017}, with $T$ fitted on the validation split, which rescales the ratios without reordering them. Figure~\ref{fig:perdigit} shows, for $s=1$, the geometric mean of the calibrated $p(d\,|\,x)/p(1\,|\,x)$ over the $1135$ test images of ``1''; Appendix~\ref{app:ranking} covers every source digit, and the numbers below are means over source digits. For a ``1'', lateral inhibition puts ``7'', the digit most similar to a ``1'', ahead of ``4'' in both networks, as one expects: consistently in the MLP (in $84\%$ of the images, against $46\%$ plain, whose order changes from seed to seed), and on average in the CNN ($59\%$ against $42\%$; the plain CNN puts ``4'' first). Otherwise the CNN's alternatives are much the same with and without lateral inhibition: the top alternative lies at $10^{-4.7}$ plain and $10^{-4.3}$ with lateral inhibition, the alternatives span $3.5$ and $4.1$ orders of magnitude, and the ranking is as consistent between seeds (Spearman $0.96$ against $0.97$) and as close to a model-free visual confusability, the class shares among each image's $50$ nearest training images ($0.61$ against $0.58$). On the MLP the difference goes further: plain, the alternatives are nearly flat ($0.4$ orders of magnitude, against $2.4$), and their order agrees less between seeds ($0.73$ against $0.90$) and with visual confusability ($0.54$ against $0.65$).

Where on the birth--survival rule each architecture operates is analysed in Appendix~\ref{app:occupancy}: the MLP reads a sparse field and acts mainly through survival, the CNN sits between the two set-points, and the WRN reads a dense field near the birth set-point.

\subsection{Noise resilience}

\begin{figure}[t]
\centering
\includegraphics[width=\textwidth]{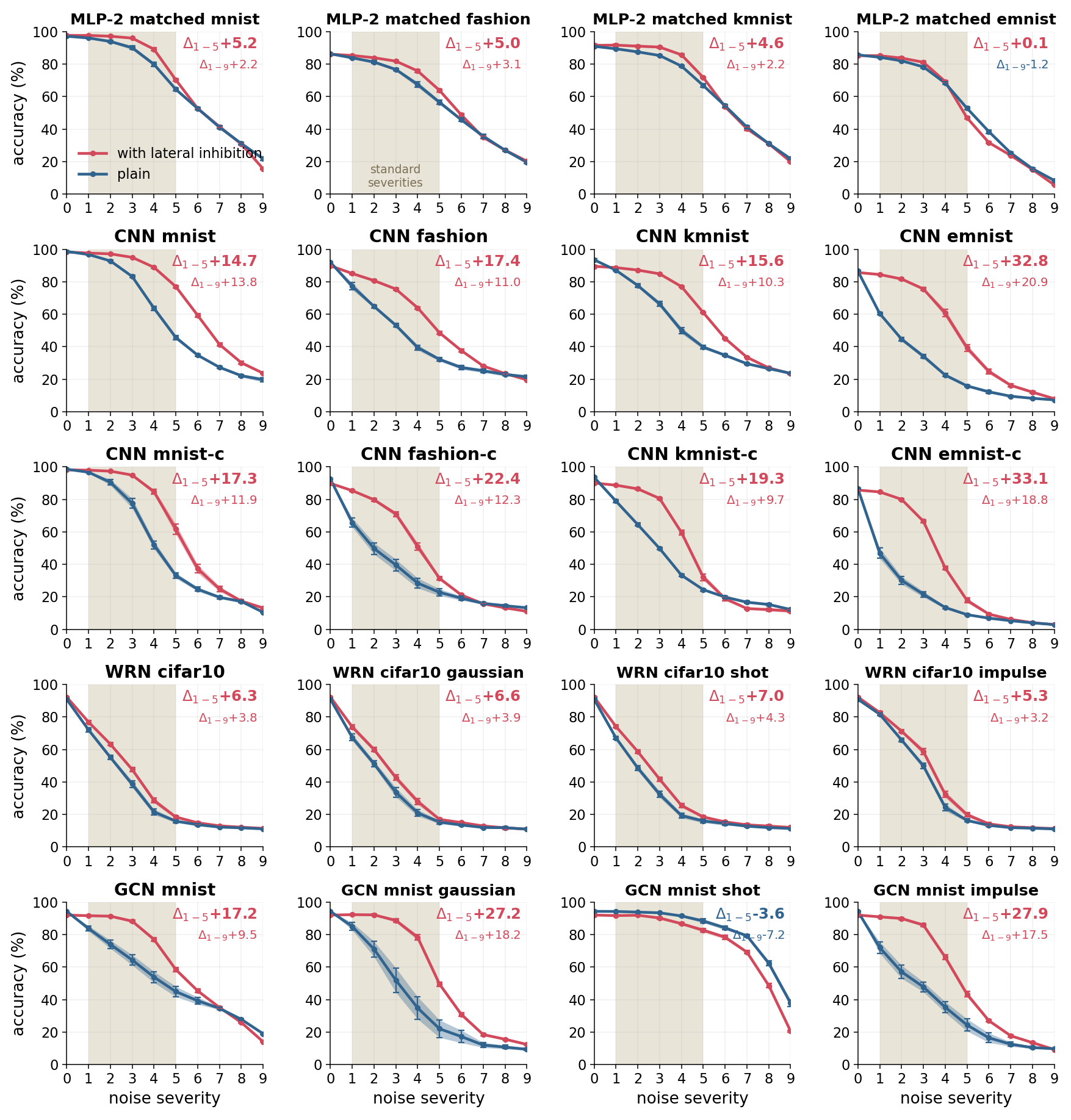}
\caption{\textbf{Noise resilience across networks and datasets.} Accuracy versus noise severity, with lateral inhibition (red) and plain (blue). Rows: the 2-layer MLP at matched clean accuracy; the CNN on the grayscale MNIST family; the CNN on the colour variants; the WRN on CIFAR-10; the GCN on MNIST. Mean over Gaussian, shot and impulse noise, except the last three panels of the bottom two rows, which show the WRN and the GCN under each noise type. Mean over five networks, $\pm1$ s.d.\ as a band and error bars (at most $3.6$ points, and $7.5$ on the GCN per-type panels). Severity $0$ is clean; the shaded band marks the standard severities $1$--$5$, and $6$--$9$ extend them toward pure noise. $\Delta_{1\text{--}5}$ (bold) and $\Delta_{1\text{--}9}$: mean gain with lateral inhibition.}
\label{fig:noise_resilience}
\end{figure}

For each backbone we train the network with lateral inhibition and its plain counterpart with the same recipe and evaluate both on held-out data under increasing noise. Figure~\ref{fig:noise_resilience} shows the mean accuracy over Gaussian, shot and impulse noise for the 2-layer MLP at matched clean accuracy, the CNN on the grayscale and colour MNIST family, the WRN on CIFAR-10, and the GCN on MNIST; Appendix~\ref{app:noise} breaks this down by noise type and adds a blend with uniform noise (Figures~\ref{fig:nb_fcn2}--\ref{fig:nb_cnncolour}).

On the \emph{shallow convolutional} network lateral inhibition raises accuracy under noise by double digits on every dataset, grayscale and colour, most on EMNIST (Figure~\ref{fig:noise_resilience}; Figure~\ref{fig:noise_gap} plots the gap at each severity). Every seed pair gains on every dataset, and the $95\%$ intervals that resample seed pairs and test images lie far from zero (Appendix~\ref{app:uncertainty}); on the grayscale datasets lateral inhibition recovers roughly half to two-thirds of the accuracy that noise costs the plain network. The clean-accuracy cost is about a point on MNIST, EMNIST and their colour versions and a few points on FashionMNIST and KMNIST. On the \emph{wide residual network} it raises it by single digits, for each noise type, several times the clean difference. On the \emph{graph convolutional network} it raises accuracy by double digits under Gaussian and impulse noise, at a clean cost of about two points, and lowers it under shot noise (Figure~\ref{fig:noise_resilience}, bottom row). Trained to matched clean accuracy, the 2-layer MLP with lateral inhibition is the more robust on MNIST, FashionMNIST and KMNIST, mostly from impulse noise, and indistinguishable from plain on EMNIST; the plain MLP keeps gaining robustness as it trains while the one with lateral inhibition does not, so the comparison depends on the training budget (Appendix~\ref{app:snapshots}). The noise benefit is thus largest on the shallow networks whose first lateral-inhibition layer reads the image through a local operation, the CNN and the GCN, and smaller on the WRN, whose lateral-inhibition layers read only learned features. The official benchmarks agree (Table~\ref{tab:officialc}): on CIFAR-10-C the WRN gains, and on MNIST-C impulse noise the CNN gains.

\begin{figure}[t]
\centering
\includegraphics[width=0.85\textwidth]{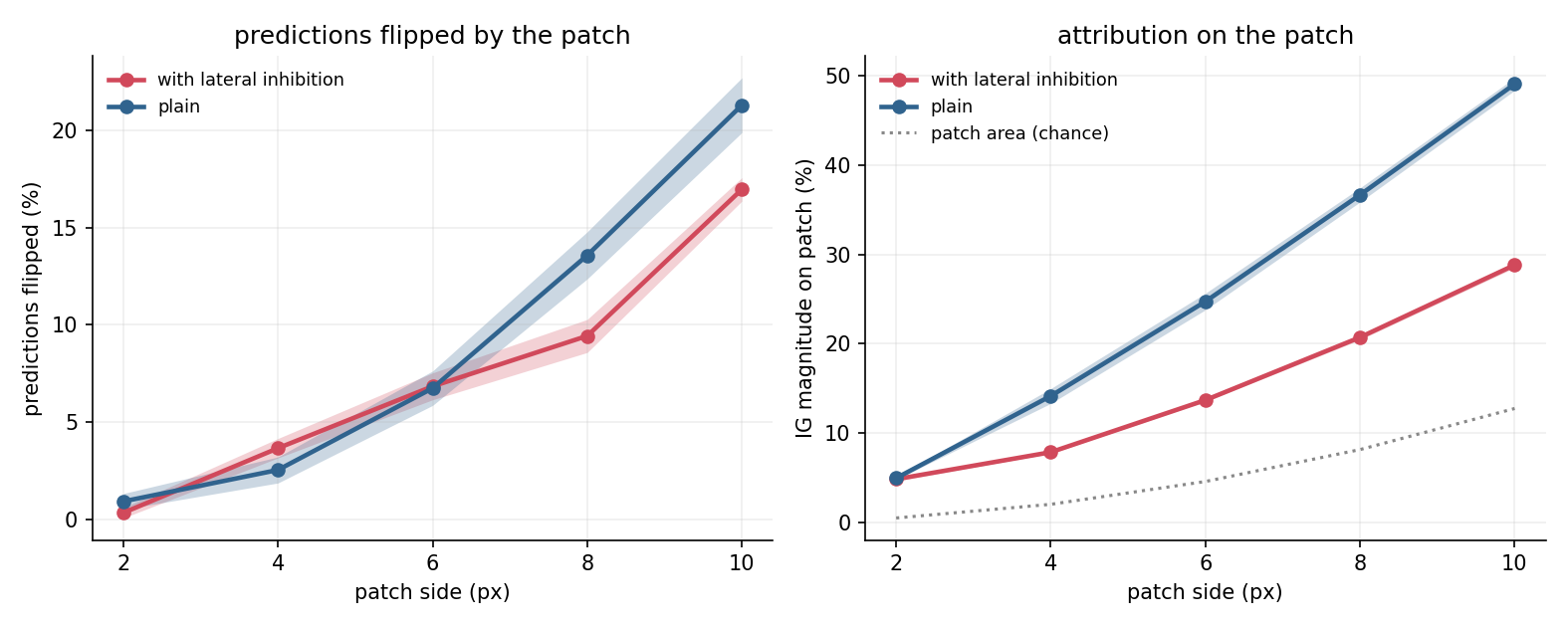}
\caption{\textbf{Robustness to an inference-time decoy patch} (clean-trained CNN, MNIST; mean over five networks, shaded $\pm1$ s.d.). \emph{Left:} fraction of correct predictions flipped vs.\ patch side. \emph{Right:} share of integrated-gradient magnitude falling on the patch, with the patch-area chance line. Patches larger than $10\times10$ cover most of the digit and act as occlusions; they are not shown.}
\label{fig:decoy}
\end{figure}

\paragraph{Robustness to a localised decoy.}
We also pasted a bright square patch at a random location into held-out MNIST images at inference, for the clean-trained CNN (five networks per arm), up to $10\times10$ ($13\%$ of the image; larger patches cover most of the digit and act as occlusions). Patches up to $6\times6$ flip few correct predictions in either network (below $7\%$); larger ones flip fewer with lateral inhibition ($9.4\%$ against $13.6\%$ at $8\times8$, $17.0\%$ against $21.3\%$ at $10\times10$; Figure~\ref{fig:decoy}). From $4\times4$ on, the network with lateral inhibition also draws far less integrated-gradient attribution onto the patch ($28.8\%$ against $49.0\%$ at $10\times10$; chance $12.8\%$), for positive and negative attribution alike.

\section{Discussion}

The main contribution of this work is to regulate the output of a network layer, or block, with a fixed, parameter-free function of its own input. Had the optimiser access to the regulation---a learned map from the layer input into the gates, say---it could dial the gates to whatever lowers the loss, and lateral inhibition would become one more learned layer. A related principle, a structure fixed by construction that holds at every inference step and complements rather than replaces regularisation, was recently applied to latent world models~\cite{yalovetzky2026lehami}; there the constraint forbids contraction, here it removes fine-scale structure, but in neither case can the optimiser trade it away. This matters because our goals pull in two directions: training optimises clean accuracy, while the organised representation and the robustness we observe are never part of the loss, and a learnable regulation would be shaped for the first whether or not that serves the second. The rule's parameters receive no gradient (Appendix~\ref{app:li-backward}), but its values and slopes shape training throughout the network: the gate values scale every weight gradient of their layer, and through the error a layer passes down, its slopes reach the weight gradients of the layers below. The rule is fixed, but its input is not: in the trained deep networks this gate route carries nearly all of the weight gradient of every layer but the last, so the network learns largely by placing its activity where the fixed rule acts (Appendix~\ref{app:steering}). Cutting it leaves the behaviour under noise and the localised saliency unchanged, so these come from the rule acting forward, but it costs the MLP its coherent classifier weights and half of its scale separation and leaves the WRN unable to fit its training set. Conversely, when we gave the optimiser a learnable path around the rule---a learned colour map in front of the layer that reads the image---it took it, routing colour content around the rule and losing the object-localised saliency (Appendix~\ref{sec:channels}). The design assumed that a network has enough degrees of freedom to reach the same clean accuracy with and without the regulation. This held only in part: the networks with lateral inhibition give up clean accuracy for properties the loss never asked for---nothing on the wide residual network, which gains $0.8$ points, but up to five points on the MLP. When $\mathcal{B}=\mathcal{S}$ the gates coincide and the blend of Eq.~\eqref{eq:blend} reduces to a fixed nonlinear filter of the input, $\tilde{y}=B(m)/N$; what distinguishes the rule from a fixed local filter is the asymmetric, state-dependent blend, the same asymmetry that favours complex dynamics in Life-like automata~\citep{Cohen2026}.

The Game-of-Life rule is only one choice among the $2^{18}$ Life-like rules and beyond, taken first because two-dimensional CA turn local updates into complex large-scale structure. That space could be searched by evolutionary algorithms or by automated searches with a foundation model as judge~\citep{Kumar2024ASAL}, selecting the rule for a second objective while gradient descent trains the network for the first; we leave this to future work.

The simplest alternative rule we tested blends a low-pass-filtered copy of the input into the output, $y=(1-\lambda)z+\lambda\,\pi(m)$ (the \emph{input blur}). It is not an external baseline but a second instance of the same approach, with a linear rule in place of birth and survival, so comparing the two asks what the CA rule adds (Appendices~\ref{app:frame} and~\ref{app:saliency}). A strong enough input blur reproduces the noise resilience of the CNN ($87.2$--$90.9\%$, against $91.7\%$ with lateral inhibition and $76.4\%$ plain; five seeds) and, in the MLP, the smooth maps and classifier weights, the consistency across depth and the persistence of the coarse maps, at higher accuracy; a blur of the output, which never reads $x$, smooths the maps as much but reproduces none of the rest, so these properties come from reading the input. Three properties are distinctive to the CA rule among the regulations we tested. First, the CNN's object-localised saliency: log-ratio $+1.13$, against $+0.37$ to $+0.48$ for the input blurs, $+0.47$ plain and $+0.04$ for a published surround-modulation filter~\citep{Hasani2019}, which also recovers most of the noise gain ($86.7\%$). Second, learned agreement between independently trained MLPs: $0.11$ at initialisation and $0.83$ after training with \texttt{B5/S3}, whereas a blur strong enough to agree after training ($0.97$) agrees already at initialisation ($0.74$), and weaker blurs do not agree at all. Third, an input-dependent gain: the CA rule scales each unit by $S(m)-B(m)$, $8.5$ times larger on the strokes than off them, whereas the blur scales every unit by $1-\lambda$. We therefore attribute noise resilience, smoothness and consistency across depth to input regulation, and localisation and learned agreement to the specific CA rule; other rules may share them. Without the normaliser, lateral inhibition scales and shifts each unit of the output, $y=\gamma(m)\odot z+\beta(m)$ with $\gamma=S-B$ and $\beta=S+B-1$, the algebraic form of FiLM~\citep{Perez2018} and SPADE~\citep{Park2019}. It differs in being \emph{self-conditioned} (computed from the layer's own input, not an external signal), \emph{parameter-free} (a fixed closed-form function of the neighbourhood density, with no learned generator) and a \emph{spatially-varying field} (FiLM applies one scalar per channel; SPADE varies spatially but is learned and externally conditioned).

Not all of the MLP's coherence carries over to the convolutional backbones. Independently trained CNNs agree unit for unit in the first layer, whose gates read the image, but not beyond it: lateral-inhibition layers that read features reduce over channels whose order differs between trained networks, so averaging the weights of independently trained networks fails, as it does for plain networks. The spatial effects within each feature map, object-localised saliency and noise resilience, persist. On the graph convolutional network, whose message passing already sums each node with its neighbours, the gate adds the separate treatment of a node's own state and its neighbours' that defines a cellular automaton; it raises accuracy under Gaussian and impulse noise as on the CNN, but not under shot noise.

In artificial neural networks, lateral inhibition usually enters in a simple form, as normalisation: local response and local contrast normalisation divide each unit's activity by that of its neighbours~\citep{Krizhevsky2012,Jarrett2009}, after divisive normalisation in the brain~\citep{CarandiniHeeger2012}, and batch normalisation and its variants standardise each unit against a population of others~\citep{IoffeSzegedy2015}. Spiking networks model it more elaborately, as an actual interaction between units that is fast and non-local, making them compete as in winner-take-all circuits~\citep{Kaplanoglou2024,diehl2015stdp}. Since the non-binary activations of standard networks resemble rate coding~\citep{Bennett2023}, the lateral inhibition proposed here acts on the rate-coded value itself: a slow, spatially local modulation rather than a fast competition between spikes, closer to the modulatory influence of glial cells~\citep{Araque1999}. In CA terms, the fixed local rule plays the role of hardware and the learnable non-local weights that of software.

To make this organisation directly visible, an interactive demonstration at \url{https://lateralintuition.com} lets the reader draw a digit and follow it, layer by layer, through a 10-layer MLP with and without lateral inhibition, computed entirely in the browser.

\section{Caveats}

The method in its current form has several caveats that we have not resolved.

\begin{itemize}
    \item \emph{Training is harder.} Networks with lateral inhibition learn more slowly, often need longer schedules, and pay up to a few points of clean accuracy. Not every rule trains, and deep stacks need the normaliser and a bounded activation.
    \item \emph{Wiring takes work for each architecture.} Which input each lateral-inhibition layer reads, how channels are reduced, whether a stem precedes it and how residual paths pass around it all change the outcome; we have no recipe that guarantees a working design for a new architecture.
    \item \emph{Lateral inhibition is sensitive to the input normalisation.} The rule reads an occupancy derived from the normalised input (Eq.~\eqref{eq:occupancy-normalisation}), so the normalisation moves the data relative to the bands; other choices, colour backgrounds and inverted contrast can switch its effect off or reverse it.
    \item \emph{The benefits depend on backbone and data.} With lateral inhibition, accuracy under noise rises by double digits on the shallow CNN and, except under shot noise, on the GCN, and by single digits on the WRN, and some coherence properties of the MLP do not carry over to convolutional backbones.
\end{itemize}

\section*{Code availability}
The code that trains the networks of this paper and reproduces its figures, tables and numbers is available at \url{https://github.com/yairchn/Conway-Net}. More on this work is at \url{https://lateralintuition.com}.

\section*{Competing interests and patent disclosure}
The method described in this paper is the subject of a pending International (PCT) Patent Application, No.~PCT/IL2026/050267, filed 24~March 2026, which claims priority to U.S.\ Provisional Patent Application No.~63/860,671, filed 8~August 2025. Yair Cohen is the sole inventor and owner of both applications. The code accompanying this paper is released under the PolyForm Noncommercial License~1.0.0, which permits noncommercial use, including academic research and teaching. Commercial use of the code or of the method requires a separate licence from Yair Cohen; licensing inquiries: \href{mailto:yair@lateralintuition.com}{\textcolor{blue}{yair@lateralintuition.com}}.

\clearpage
\bibliographystyle{abbrvnat}
\bibliography{ref}

\begin{thebibliography}{65}
\providecommand{\natexlab}[1]{#1}
\providecommand{\url}[1]{\texttt{#1}}
\expandafter\ifx\csname urlstyle\endcsname\relax
  \providecommand{\doi}[1]{doi: #1}\else
  \providecommand{\doi}{doi: \begingroup \urlstyle{rm}\Url}\fi

\bibitem[Aaronson et~al.(2014)Aaronson, Carroll, and Ouellette]{Aaronson2014}
S.~Aaronson, S.~M. Carroll, and L.~Ouellette.
\newblock Quantifying the rise and fall of complexity in closed systems: The
  coffee automaton, 2014.
\newblock URL \url{https://arxiv.org/abs/1405.6903}.

\bibitem[Araque et~al.(1999)Araque, Parpura, Sanzgiri, and Haydon]{Araque1999}
A.~Araque, V.~Parpura, R.~P. Sanzgiri, and P.~G. Haydon.
\newblock Tripartite synapses: glia, the unacknowledged partner.
\newblock \emph{Trends in Neurosciences}, 22\penalty0 (5):\penalty0 208--215,
  1999.
\newblock \doi{10.1016/S0166-2236(98)01349-6}.

\bibitem[B{\'e}na et~al.(2025)B{\'e}na, Faldor, Goodman, and
  Cully]{bena2025unca}
G.~B{\'e}na, M.~Faldor, D.~F.~M. Goodman, and A.~Cully.
\newblock A path to universal neural cellular automata.
\newblock In \emph{Proceedings of the Genetic and Evolutionary Computation
  Conference Companion (GECCO '25 Companion)}, pages 2099--2107, 2025.
\newblock \doi{10.1145/3712255.3734310}.

\bibitem[Bennett(2023)]{Bennett2023}
M.~Bennett.
\newblock \emph{A Brief History of Intelligence: Evolution, {AI}, and the Five
  Breakthroughs That Made Our Brains}.
\newblock Mariner Books, New York, 2023.

\bibitem[Breskin et~al.(2006)Breskin, Soriano, Moses, and Tlusty]{Breskin2006}
I.~Breskin, J.~Soriano, E.~Moses, and T.~Tlusty.
\newblock Percolation in living neural networks.
\newblock \emph{Physical Review Letters}, 97\penalty0 (18):\penalty0 188102,
  2006.
\newblock \doi{10.1103/PhysRevLett.97.188102}.

\bibitem[Carandini and Heeger(2012)]{CarandiniHeeger2012}
M.~Carandini and D.~J. Heeger.
\newblock Normalization as a canonical neural computation.
\newblock \emph{Nature Reviews Neuroscience}, 13\penalty0 (1):\penalty0 51--62,
  2012.
\newblock \doi{10.1038/nrn3136}.
\newblock URL
  \url{https://www.cns.nyu.edu/heegerlab/content/publications/Carandini-NRN2012.pdf}.

\bibitem[Chan(2019)]{Chan2019}
B.~W.-C. Chan.
\newblock Lenia: Biology of artificial life.
\newblock \emph{Complex Systems}, 28\penalty0 (3):\penalty0 251--286, 2019.

\bibitem[Clauset et~al.(2009)Clauset, Shalizi, and Newman]{Clauset2009}
A.~Clauset, C.~R. Shalizi, and M.~E.~J. Newman.
\newblock Power-law distributions in empirical data.
\newblock \emph{SIAM Review}, 51\penalty0 (4):\penalty0 661--703, 2009.
\newblock \doi{10.1137/070710111}.

\bibitem[Cohen(2026)]{Cohen2026}
Y.~Cohen.
\newblock Rule-table structure of life-like cellular automata: activity,
  asymmetry, sensitivity, and the propagation--storage factorisation, 2026.
\newblock Submitted to \emph{Complex Systems}.

\bibitem[Cook(2004)]{Cook2004}
M.~Cook.
\newblock Universality in elementary cellular automaton rule 110.
\newblock \emph{Complex Systems}, 15\penalty0 (1):\penalty0 1--40, 2004.

\bibitem[Cybenko(1989)]{Cybenko1989}
G.~Cybenko.
\newblock Approximation by superpositions of a sigmoidal function.
\newblock \emph{Mathematics of Control, Signals and Systems}, 2\penalty0
  (4):\penalty0 303--314, 1989.
\newblock \doi{10.1007/BF02551274}.

\bibitem[Dall and Christensen(2002)]{Dall2002}
J.~Dall and M.~Christensen.
\newblock Random geometric graphs.
\newblock \emph{Physical Review E}, 66\penalty0 (1):\penalty0 016121, 2002.

\bibitem[Deutges et~al.(2024)Deutges, Bensch, Martin,
  et~al.]{Deutges2024MedNCA}
M.~Deutges, M.~Bensch, K.~Martin, et~al.
\newblock Neural cellular automata for lightweight, robust and input-invariant
  medical image segmentation.
\newblock In \emph{Medical Image Computing and Computer-Assisted Intervention
  (MICCAI) Workshops}, 2024.
\newblock URL \url{https://papers.miccai.org/miccai-2024/paper/3422_paper.pdf}.

\bibitem[Diehl and Cook(2015)]{diehl2015stdp}
P.~U. Diehl and M.~Cook.
\newblock Unsupervised learning of digit recognition using
  spike-timing-dependent plasticity.
\newblock \emph{Frontiers in Computational Neuroscience}, 9:\penalty0 99, 2015.
\newblock \doi{10.3389/fncom.2015.00099}.
\newblock URL
  \url{https://www.frontiersin.org/articles/10.3389/fncom.2015.00099/full}.

\bibitem[Dosovitskiy et~al.(2021)Dosovitskiy, Beyer, Kolesnikov, Weissenborn,
  Zhai, Unterthiner, Dehghani, Minderer, Heigold, Gelly, Uszkoreit, and
  Houlsby]{Dosovitskiy2021}
A.~Dosovitskiy, L.~Beyer, A.~Kolesnikov, D.~Weissenborn, X.~Zhai,
  T.~Unterthiner, M.~Dehghani, M.~Minderer, G.~Heigold, S.~Gelly, J.~Uszkoreit,
  and N.~Houlsby.
\newblock An image is worth 16x16 words: Transformers for image recognition at
  scale.
\newblock In \emph{International Conference on Learning Representations
  (ICLR)}, 2021.
\newblock URL \url{https://arxiv.org/abs/2010.11929}.

\bibitem[Erd{\H{o}}s and R{\'e}nyi(1960)]{ErdosRenyi1960}
P.~Erd{\H{o}}s and A.~R{\'e}nyi.
\newblock On the evolution of random graphs.
\newblock \emph{Publications of the Mathematical Institute of the Hungarian
  Academy of Sciences}, 5:\penalty0 17--61, 1960.

\bibitem[Etcheverry et~al.(2026)Etcheverry, Miotti, Sirbu, Sch{\"u}rholt,
  Drozdova, Ghosh, Ag{\"u}era~y Arcas, Manyika, Richards, and
  Niklasson]{Etcheverry2026ReasoningNCA}
M.~Etcheverry, P.~Miotti, A.~Sirbu, K.~Sch{\"u}rholt, M.~Drozdova, A.~Ghosh,
  B.~Ag{\"u}era~y Arcas, J.~Manyika, B.~Richards, and E.~Niklasson.
\newblock Reasoning with neural cellular automata, 2026.
\newblock arXiv:2609.36126.

\bibitem[Gardner(1970)]{Gardner1970}
M.~Gardner.
\newblock Mathematical games: The fantastic combinations of {John Conway's} new
  solitaire game ``{Life}''.
\newblock \emph{Scientific American}, 223\penalty0 (4):\penalty0 120--123,
  1970.

\bibitem[Gilpin(2019)]{Gilpin2019}
W.~Gilpin.
\newblock Cellular automata as convolutional neural networks.
\newblock \emph{Physical Review E}, 100\penalty0 (3):\penalty0 032402, 2019.

\bibitem[Guidotti et~al.(2018)Guidotti, Monreale, Ruggieri, Turini, Pedreschi,
  and Giannotti]{Guidotti2018}
R.~Guidotti, A.~Monreale, S.~Ruggieri, F.~Turini, D.~Pedreschi, and
  F.~Giannotti.
\newblock A survey of methods for explaining black box models.
\newblock \emph{ACM Computing Surveys}, 51\penalty0 (5):\penalty0 93:1--93:42,
  2018.
\newblock \doi{10.1145/3236009}.

\bibitem[Guo et~al.(2017)Guo, Pleiss, Sun, and Weinberger]{Guo2017}
C.~Guo, G.~Pleiss, Y.~Sun, and K.~Q. Weinberger.
\newblock On calibration of modern neural networks.
\newblock In \emph{Proceedings of the 34th International Conference on Machine
  Learning}, volume~70 of \emph{Proceedings of Machine Learning Research},
  pages 1321--1330, 2017.

\bibitem[Hasani et~al.(2019)Hasani, Soleymani~Baghshah, and
  Aghajan]{Hasani2019}
H.~Hasani, M.~Soleymani~Baghshah, and H.~Aghajan.
\newblock Surround modulation: A bio-inspired connectivity structure for
  convolutional neural networks.
\newblock In \emph{Advances in Neural Information Processing Systems
  (NeurIPS)}, 2019.

\bibitem[Hendrycks and Dietterich(2019)]{Hendrycks2019}
D.~Hendrycks and T.~Dietterich.
\newblock Benchmarking neural network robustness to common corruptions and
  perturbations.
\newblock In \emph{International Conference on Learning Representations
  (ICLR)}, 2019.

\bibitem[Hendrycks et~al.(2020)Hendrycks, Mu, Cubuk, Zoph, Gilmer, and
  Lakshminarayanan]{Hendrycks2020}
D.~Hendrycks, N.~Mu, E.~D. Cubuk, B.~Zoph, J.~Gilmer, and B.~Lakshminarayanan.
\newblock {AugMix}: A simple data processing method to improve robustness and
  uncertainty.
\newblock In \emph{International Conference on Learning Representations
  (ICLR)}, 2020.

\bibitem[Hornik(1991)]{Hornik1991}
K.~Hornik.
\newblock Approximation capabilities of multilayer feedforward networks.
\newblock \emph{Neural Networks}, 4\penalty0 (2):\penalty0 251--257, 1991.
\newblock \doi{10.1016/0893-6080(91)90009-T}.

\bibitem[Ioffe and Szegedy(2015)]{IoffeSzegedy2015}
S.~Ioffe and C.~Szegedy.
\newblock Batch normalization: Accelerating deep network training by reducing
  internal covariate shift.
\newblock In \emph{Proceedings of the 32nd International Conference on Machine
  Learning (ICML)}, pages 448--456, 2015.
\newblock URL \url{https://proceedings.mlr.press/v37/ioffe15.html}.

\bibitem[Jarrett et~al.(2009)Jarrett, Kavukcuoglu, Ranzato, and
  LeCun]{Jarrett2009}
K.~Jarrett, K.~Kavukcuoglu, M.~Ranzato, and Y.~LeCun.
\newblock What is the best multi-stage architecture for object recognition?
\newblock In \emph{IEEE International Conference on Computer Vision (ICCV)},
  pages 2146--2153, 2009.
\newblock \doi{10.1109/ICCV.2009.5459469}.
\newblock URL
  \url{https://www.cs.toronto.edu/~ranzato/publications/architecture-iccv09.pdf}.

\bibitem[Jordan et~al.(2024)Jordan, Jin, Boza, You, Cesista, Newhouse, and
  Bernstein]{jordan2024muon}
K.~Jordan, Y.~Jin, V.~Boza, J.~You, F.~Cesista, L.~Newhouse, and J.~Bernstein.
\newblock Muon: An optimizer for hidden layers in neural networks.
\newblock \url{https://kellerjordan.github.io/posts/muon/}, 2024.

\bibitem[Kaplanoglou and Diamantaras(2024)]{Kaplanoglou2024}
P.~I. Kaplanoglou and K.~Diamantaras.
\newblock Learning local discrete features in explainable-by-design
  convolutional neural networks, 2024.
\newblock URL \url{https://arxiv.org/abs/2411.00139}.

\bibitem[Kingma and Ba(2015)]{KingmaBa2015}
D.~P. Kingma and J.~Ba.
\newblock Adam: A method for stochastic optimization.
\newblock In \emph{International Conference on Learning Representations
  (ICLR)}, 2015.

\bibitem[Kipf and Welling(2017)]{KipfWelling2017}
T.~N. Kipf and M.~Welling.
\newblock Semi-supervised classification with graph convolutional networks.
\newblock In \emph{International Conference on Learning Representations
  (ICLR)}, 2017.

\bibitem[Krizhevsky et~al.(2012)Krizhevsky, Sutskever, and
  Hinton]{Krizhevsky2012}
A.~Krizhevsky, I.~Sutskever, and G.~E. Hinton.
\newblock {ImageNet} classification with deep convolutional neural networks.
\newblock In \emph{Advances in Neural Information Processing Systems
  (NeurIPS)}, volume~25, 2012.
\newblock URL
  \url{https://proceedings.neurips.cc/paper/4824-imagenet-classification-with-deep-convolutional-neural-networks.pdf}.

\bibitem[Kumar et~al.(2024)Kumar, Lu, Kirsch, Tang, Stanley, Isola, and
  Ha]{Kumar2024ASAL}
A.~Kumar, C.~Lu, L.~Kirsch, Y.~Tang, K.~O. Stanley, P.~Isola, and D.~Ha.
\newblock Automating the search for artificial life with foundation models,
  2024.

\bibitem[Langton(1990)]{Langton1990EdgeOfChaos}
C.~G. Langton.
\newblock Computation at the edge of chaos: Phase transitions and emergent
  computation.
\newblock \emph{Physica D: Nonlinear Phenomena}, 42\penalty0 (1-3):\penalty0
  12--37, 1990.
\newblock \doi{10.1016/0167-2789(90)90064-V}.

\bibitem[Lindenmayer(1968{\natexlab{a}})]{Lindenmayer1968a}
A.~Lindenmayer.
\newblock Mathematical models for cellular interactions in development. {I}.
  filaments with one-sided inputs.
\newblock \emph{Journal of Theoretical Biology}, 18\penalty0 (3):\penalty0
  280--299, 1968{\natexlab{a}}.
\newblock \doi{10.1016/0022-5193(68)90079-9}.

\bibitem[Lindenmayer(1968{\natexlab{b}})]{Lindenmayer1968b}
A.~Lindenmayer.
\newblock Mathematical models for cellular interactions in development. {II}.
  simple and branching filaments with two-sided inputs.
\newblock \emph{Journal of Theoretical Biology}, 18\penalty0 (3):\penalty0
  300--315, 1968{\natexlab{b}}.
\newblock \doi{10.1016/0022-5193(68)90080-5}.

\bibitem[Lipton(2018)]{Lipton2016}
Z.~C. Lipton.
\newblock The mythos of model interpretability.
\newblock \emph{Communications of the ACM}, 61\penalty0 (10):\penalty0 36--43,
  2018.
\newblock \doi{10.1145/3233231}.

\bibitem[Luo et~al.(2016)Luo, Li, Urtasun, and Zemel]{Luo2016}
W.~Luo, Y.~Li, R.~Urtasun, and R.~Zemel.
\newblock Understanding the effective receptive field in deep convolutional
  neural networks.
\newblock In \emph{Advances in Neural Information Processing Systems
  (NeurIPS)}, 2016.
\newblock URL \url{https://arxiv.org/abs/1701.04128}.

\bibitem[Ly and Gong(2026)]{LyGong2026}
A.~Ly and P.~Gong.
\newblock Fractal basin geometry and the limits to predictability and
  reproducibility in deep learning.
\newblock \emph{Nature Machine Intelligence}, 2026.
\newblock \doi{10.1038/s42256-026-01336-3}.

\bibitem[Mitchell(2009)]{Mitchell2009}
M.~Mitchell.
\newblock Complexity: A guided tour.
\newblock \emph{Oxford University Press}, 2009.

\bibitem[Mordvintsev et~al.(2020)Mordvintsev, Randazzo, Niklasson, and
  Levin]{Mordvintsev2020GrowingCA}
A.~Mordvintsev, E.~Randazzo, E.~Niklasson, and M.~Levin.
\newblock Growing neural cellular automata.
\newblock \emph{Distill}, 5\penalty0 (2), 2020.
\newblock \doi{10.23915/distill.00023}.
\newblock URL \url{https://distill.pub/2020/growing-ca}.

\bibitem[Mordvintsev et~al.(2022)Mordvintsev, Randazzo, and
  Fouts]{Mordvintsev2022IsoNCA}
A.~Mordvintsev, E.~Randazzo, and C.~Fouts.
\newblock Growing isotropic neural cellular automata.
\newblock In \emph{Artificial Life Conference Proceedings}, pages 65--73, 2022.
\newblock URL
  \url{https://direct.mit.edu/isal/proceedings-pdf/isal2022/34/65/2035305/isal_a_00552.pdf}.

\bibitem[Mu and Gilmer(2019)]{MuGilmer2019}
N.~Mu and J.~Gilmer.
\newblock {MNIST-C}: A robustness benchmark for computer vision.
\newblock \emph{arXiv preprint arXiv:1906.02337}, 2019.

\bibitem[Niklasson et~al.(2021)Niklasson, Mordvintsev, Randazzo, and
  Levin]{Niklasson2021Textures}
E.~Niklasson, A.~Mordvintsev, E.~Randazzo, and M.~Levin.
\newblock Self-organising textures.
\newblock \emph{Distill}, 6\penalty0 (2), 2021.
\newblock \doi{10.23915/distill.00027.003}.
\newblock URL \url{https://distill.pub/selforg/2021/textures}.

\bibitem[Olah et~al.(2017)Olah, Mordvintsev, and Schubert]{Olah2017}
C.~Olah, A.~Mordvintsev, and L.~Schubert.
\newblock Feature visualization.
\newblock \emph{Distill}, 2017.
\newblock \doi{10.23915/distill.00007}.
\newblock URL \url{https://distill.pub/2017/feature-visualization/}.

\bibitem[Park et~al.(2019)Park, Liu, Wang, and Zhu]{Park2019}
T.~Park, M.-Y. Liu, T.-C. Wang, and J.-Y. Zhu.
\newblock Semantic image synthesis with spatially-adaptive normalization.
\newblock In \emph{IEEE/CVF Conference on Computer Vision and Pattern
  Recognition (CVPR)}, 2019.

\bibitem[Perez et~al.(2018)Perez, Strub, de~Vries, Dumoulin, and
  Courville]{Perez2018}
E.~Perez, F.~Strub, H.~de~Vries, V.~Dumoulin, and A.~Courville.
\newblock {FiLM}: Visual reasoning with a general conditioning layer.
\newblock In \emph{AAAI Conference on Artificial Intelligence}, 2018.

\bibitem[Pesce et~al.(2026)Pesce, He, and Caldarelli]{Pesce2026}
D.~Pesce, Y.-H. He, and G.~Caldarelli.
\newblock Phase transitions in neural networks pruning, 2026.

\bibitem[Pope(2000)]{Pope2000}
S.~B. Pope.
\newblock \emph{Turbulent Flows}.
\newblock Cambridge University Press, Cambridge, 2000.

\bibitem[Prusinkiewicz and Lindenmayer(1990)]{Prusinkiewicz1990ABOP}
P.~Prusinkiewicz and A.~Lindenmayer.
\newblock \emph{The Algorithmic Beauty of Plants}.
\newblock Springer, New York, 1990.
\newblock \doi{10.1007/978-1-4613-8476-2}.

\bibitem[Rafler(2011)]{Rafler2011}
S.~Rafler.
\newblock Generalization of {Conway's} ``{Game of Life}'' to a continuous
  domain -- {SmoothLife}.
\newblock arXiv preprint arXiv:1111.1567, 2011.

\bibitem[Randazzo et~al.(2023)Randazzo, Mordvintsev, and
  Fouts]{Randazzo2023SteerableNCA}
E.~Randazzo, A.~Mordvintsev, and C.~Fouts.
\newblock Growing steerable neural cellular automata.
\newblock In \emph{Artificial Life Conference Proceedings}, pages 2--11, 2023.
\newblock \doi{10.1162/isal_a_00564}.
\newblock URL \url{https://arxiv.org/abs/2302.10197}.

\bibitem[Rendell(2011)]{Rendell2011}
P.~Rendell.
\newblock A turing machine in conway's game of life.
\newblock \emph{arXiv}, 2011.
\newblock URL \url{https://arxiv.org/abs/1108.1232}.

\bibitem[Richardson et~al.(2024)Richardson, Brackston, Girolami, and
  Jones]{Richardson2024}
A.~D. Richardson, D.~R. Brackston, M.~Girolami, and E.~J. Jones.
\newblock Learning spatio-temporal patterns with neural cellular automata.
\newblock \emph{PLOS Computational Biology}, 20\penalty0 (10):\penalty0
  e1011589, 2024.
\newblock \doi{10.1371/journal.pcbi.1011589}.
\newblock URL
  \url{https://journals.plos.org/ploscompbiol/article?id=10.1371/journal.pcbi.1011589}.

\bibitem[Siegelmann and Sontag(1995)]{SiegelmannSontag1995}
H.~T. Siegelmann and E.~D. Sontag.
\newblock On the computational power of neural nets.
\newblock \emph{Journal of Computer and System Sciences}, 50\penalty0
  (1):\penalty0 132--150, 1995.

\bibitem[Sipper(1997)]{Sipper1997}
M.~Sipper.
\newblock \emph{Evolution of Parallel Cellular Machines: The Cellular
  Programming Approach}.
\newblock Springer, Berlin, Heidelberg, 1997.
\newblock \doi{10.1007/978-3-662-03400-5}.

\bibitem[Sudhakaran et~al.(2021)Sudhakaran, Grbic, Li, Katona, Najarro,
  Glanois, and Risi]{Sudhakaran2021}
S.~Sudhakaran, D.~Grbic, S.~Li, A.~Katona, E.~Najarro, C.~Glanois, and S.~Risi.
\newblock Growing 3d artefacts and functional machines with neural cellular
  automata.
\newblock In \emph{Artificial Life Conference Proceedings}, pages 421--430,
  2021.
\newblock URL \url{https://pure.itu.dk/files/86421939/isal_a_00451.pdf}.

\bibitem[Vaswani et~al.(2017)Vaswani, Shazeer, Parmar, Uszkoreit, Jones, Gomez,
  Kaiser, and Polosukhin]{Vaswani2017}
A.~Vaswani, N.~Shazeer, N.~Parmar, J.~Uszkoreit, L.~Jones, A.~N. Gomez,
  {\L}.~Kaiser, and I.~Polosukhin.
\newblock Attention is all you need.
\newblock In \emph{Advances in Neural Information Processing Systems
  (NeurIPS)}, 2017.
\newblock URL \url{https://arxiv.org/abs/1706.03762}.

\bibitem[von Neumann and Burks(1966)]{vonNeumann1966}
J.~von Neumann and A.~W. Burks.
\newblock \emph{Theory of Self-Reproducing Automata}.
\newblock University of Illinois Press, Urbana, IL, 1966.

\bibitem[Wolfram(1984)]{Wolfram1984}
S.~Wolfram.
\newblock Universality and complexity in cellular automata.
\newblock \emph{Physica D: Nonlinear Phenomena}, 10\penalty0 (1--2):\penalty0
  1--35, 1984.
\newblock \doi{10.1016/0167-2789(84)90245-8}.

\bibitem[Wolfram(2002)]{Wolfram2002}
S.~Wolfram.
\newblock \emph{A New Kind of Science}.
\newblock Wolfram Media, Champaign, IL, 2002.

\bibitem[Wulff and Hertz(1992)]{WulffHertz1992}
N.~Wulff and J.~A. Hertz.
\newblock Learning cellular automaton dynamics with neural networks.
\newblock In \emph{Advances in Neural Information Processing Systems 5}, pages
  631--638, 1992.

\bibitem[Yalovetzky et~al.(2026)Yalovetzky, Balestriero, and
  LeCun]{yalovetzky2026lehami}
R.~Yalovetzky, R.~Balestriero, and Y.~LeCun.
\newblock {LeHami}: Symplectic latent dynamics for world models.
\newblock \emph{arXiv preprint arXiv:2610.15267}, 2026.
\newblock URL \url{https://arxiv.org/abs/2610.15267}.

\bibitem[Zagoruyko and Komodakis(2016)]{Zagoruyko2016}
S.~Zagoruyko and N.~Komodakis.
\newblock Wide residual networks.
\newblock In \emph{Proceedings of the British Machine Vision Conference
  (BMVC)}, 2016.

\bibitem[Zeiler and Fergus(2014)]{ZeilerFergus2014}
M.~D. Zeiler and R.~Fergus.
\newblock Visualizing and understanding convolutional networks.
\newblock In \emph{European Conference on Computer Vision (ECCV)}, volume 8689
  of \emph{Lecture Notes in Computer Science}, pages 818--833, 2014.
\newblock \doi{10.1007/978-3-319-10590-1_53}.

\end{thebibliography}

\clearpage
\appendix
\begin{center}
{\LARGE\bfseries Appendix}
\end{center}
\vspace{1em}

\section{Mathematical formulation of lateral inhibition}
\label{sec:formulation}

We describe the wiring first for a fully connected network and later extend it to convolutional and wide-residual architectures (Section~\ref{sec:wiring}). Let $x$ be the input to a layer (from data or a previous layer). Following the rationale above, $x$ is used in two ways (Figure~\ref{fig:block_diagram}). First, in the standard path, it drives the non-local linear map and a pointwise nonlinearity:
\begin{equation}
    t = W\ast x + b,\qquad t' = \mathrm{Norm}(t),\qquad z = f(t'),
\end{equation}
where $W,b$ are the learnable weights and bias, $\ast$ denotes convolution (or a linear map in MLPs), $\mathrm{Norm}$ is a normalisation that keeps the pre-activation in a useful range for the bounded nonlinearity, and $f=\tanh$ throughout this study, so $z\in[-1,1]$. The choice of a \emph{bounded} activation is essential rather than stylistic, and is a deliberate departure from the now-standard unbounded rectifiers (ReLU, GELU): the cellular-automaton rule is defined on a bounded $[0,1]$ occupancy, so the maps $\tilde{x}=\phi(x)$ and $\tilde{z}=\phi(z)$ below are only well-posed when the activation is itself bounded---an unbounded pre-image has no meaningful occupancy, and the neighbour mean and birth/survival gates would be undefined. We use tanh (rather than, say, a sigmoid) because it is additionally \emph{zero-centred}, which matters for gradient flow as described next.

\begin{figure}[t]
\centering
\includegraphics[height=0.5\textwidth]{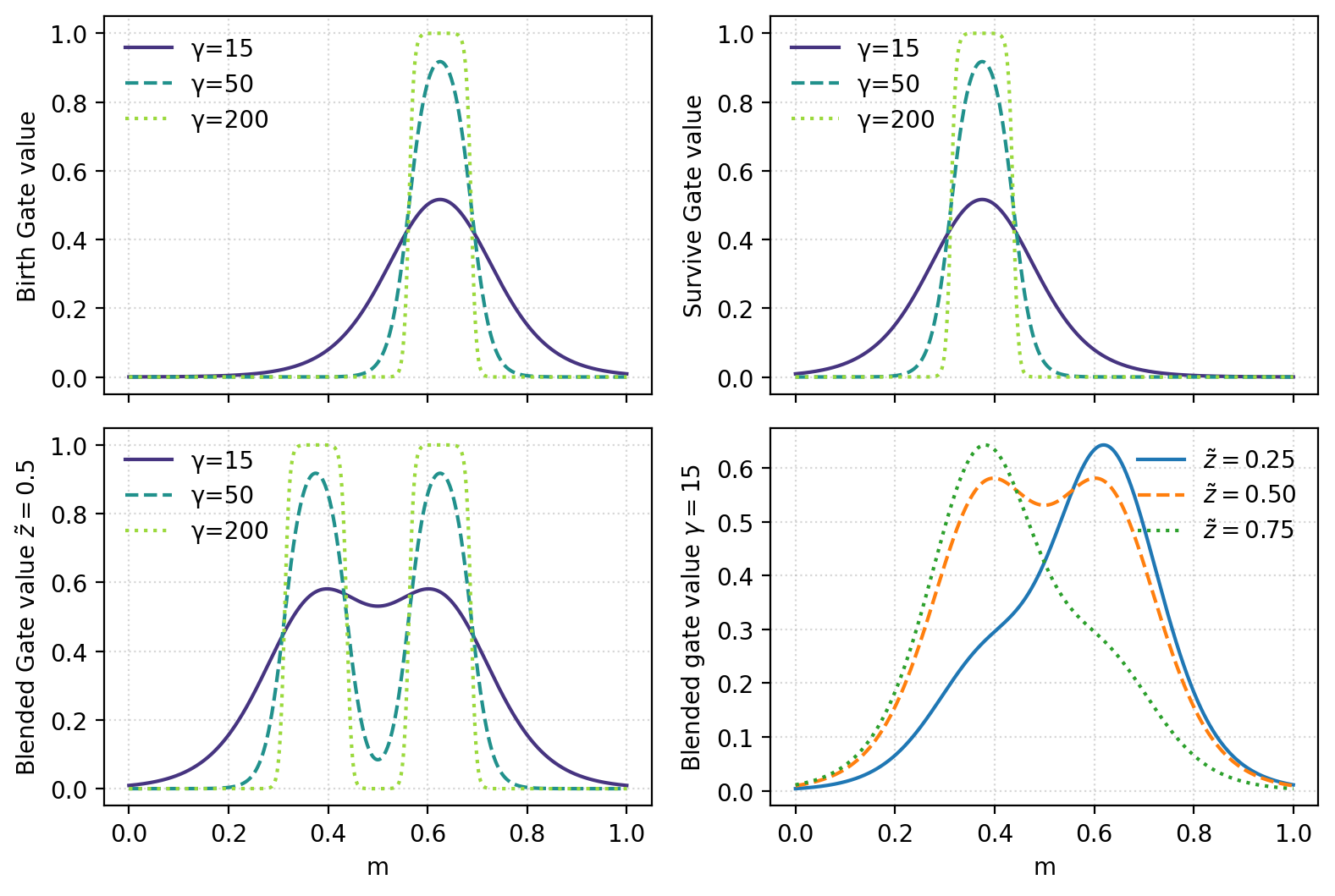}
\caption{\textbf{Gates and normalised blend for B5/S3.} Logistic Birth gate, logistic Survival gate, the normalised blend at $\tilde{z}=0.5$ for several slopes $\gamma$, and the blend at $\gamma=15$ for several values of $\tilde{z}$.}
\label{fig:gate_blend}
\end{figure}

In the second path, $x$ drives the CA Birth and Survival gates applied to $z$. The gates are computed on the grid of $x$; when $x$ and $z$ differ in spatial size, the gate maps are then resampled to the grid of $z$ by a fixed map $\psi$ (bilinear interpolation; the identity when the sizes agree), and a mismatch in channel count is resolved inside the lateral-inhibition layer by a parameter-free channel reduction (Appendix~\ref{sec:channels}). Because the gates are nonlinear, resampling the gates is not the same as computing them from a resampled input; we resample the gates. The mechanism uses three fixed transformations (Figure~\ref{fig:block_diagram}): the map $\phi$ from the activation range $[-1,1]$ to a CA state in $[0,1]$, marked by a tilde; the resampling $\psi$ of the gates to the grid of $z$; and the zero-centring $\pi$ that returns the output of lateral inhibition to the activation range. Since $x,z\in[-1,1]$ while CA rules are defined on $[0,1]$, both are mapped to their CA states (occupancies),
\begin{equation}
    \tilde{x} = \phi(x),\qquad \tilde{z} = \phi(z),\qquad \phi(u)=\tfrac{1}{2}(u+1)\in[0,1].
\end{equation}
The outer-totalistic neighbour mean $m$ is computed from $\tilde{x}$ over the Moore neighbourhood, excluding the centre, as in the Game of Life:
\begin{equation}
    m_{i,j} = \frac{1}{8}\!\left(\sum_{a=-1}^{1}\sum_{b=-1}^{1}\tilde{x}_{\,i+a,\,j+b} \;-\; \tilde{x}_{\,i,j}\right)\in[0,1],
\end{equation}
that is, $m=\mathrm{conv2D}(\tilde{x},K_{3\times3})$ with the Moore kernel $K_{3\times3}$ ($\tfrac18$ off the centre, $0$ at the centre).
To enable backpropagation we use a simplified continuous CA rule that replaces the discrete if/else logic with smooth gates. Given the Birth and Survival sets $\mathcal{B},\mathcal{S}$ (e.g.\ $\mathcal{B}=\{3\}$, $\mathcal{S}=\{2,3\}$ for the Game of Life), a differentiable bump is placed at each set-point $c$,
\begin{equation}
    \beta_{\gamma,\delta}(m;c) = \theta_{\gamma}\!\left(m-\tfrac{c-\delta}{8}\right)\,\theta_{\gamma}\!\left(\tfrac{c+\delta}{8}-m\right),
\end{equation}
where $\theta_{\gamma}(u)=1/(1+e^{-\gamma u})$ is the logistic step of slope $\gamma$ (used throughout; a ReLU ramp is an alternative that is not smooth), and $\delta$ sets the half-width of the band, $\delta/8$ in occupancy. At the operating point, $\gamma=15$ and $\delta=0.5$, the band is narrow compared with the step, so each bump has no plateau: it peaks at $0.52$ and its slope is non-zero at every occupancy. The bumps of each set are combined using De~Morgan's soft-OR, $a\lor b=1-(1-a)(1-b)$, which is differentiable:
\begin{equation}
    \mathrm{Birth}(m) = 1 - \prod_{c\in\mathcal{B}}\bigl(1-\beta_{\gamma,\delta}(m;c)\bigr),\qquad
    \mathrm{Survival}(m) = 1 - \prod_{c\in\mathcal{S}}\bigl(1-\beta_{\gamma,\delta}(m;c)\bigr).
\end{equation}
These continuous rules are deliberately simpler than SmoothLife or Lenia~\cite{Rafler2011,Chan2019}; their simplicity is an advantage inside a neural layer. The occupancy state is updated by a blend of the two gates, selected by the current (continuous) state $\tilde{z}$: a convex combination of $B=\psi(\mathrm{Birth}(m))$ and $S=\psi(\mathrm{Survival}(m))$, divided by a normaliser and bounded to $[0,1]$,
\begin{equation}
    \tilde{y} = \min\!\Bigl(1,\;\frac{(1-\tilde{z})\,B \;+\; \tilde{z}\,S}{(1-\tilde{z})^2 + \tilde{z}^2}\Bigr),
    \label{eq:blend}
\end{equation}
where the denominator (the blend's self-energy) $N=(1-\tilde{z})^2+\tilde{z}^2$ ranges over $[\tfrac12,1]$: it equals $1$ at the dead and live rails ($\tilde{z}\in\{0,1\}$), where $\tilde{y}$ is just the convex numerator ($\le 1$), and falls to its minimum $\tfrac12$ at $\tilde{z}=\tfrac12$, where the numerator is divided by $\tfrac12$ and hence amplified by up to a factor of two. The normalisation therefore leaves fully dead or alive states unchanged while \emph{lifting} intermediate states, counteracting the shrinkage of dynamic range across many stacked layers. In principle $\tilde{y}\in[0,2]$ and so $y\in[-1,3]$; in practice this headroom is barely used, because the amplification requires the numerator near $1$ \emph{and} $\tilde{z}\approx\tfrac12$ simultaneously. With the $B5/S3$ rule, $\gamma=15$ and $\delta=0.5$ the birth and survival bumps peak at $m\approx5/8$ and $3/8$ and overlap only through their tails ($\mathrm{Birth}\cdot\mathrm{Survival}\le0.07$), so the numerator stays well below unity and the blend is at most $0.66$; measured on the trained deep MLP (the three networks are described in Section~\ref{sec:wiring}) the per-layer output of lateral inhibition stays well inside the activation range (maximum $\approx0.3$, with no values outside $[-1,1]$ at any of the ten layers), so the clip of $\tilde y$ to $[0,1]$ never acts for this rule; it acts only for rules whose birth and survival sets share a count, most of all Day~\&~Night (Appendix~\ref{app:range}). The normalised blend is thus not itself a convex combination, which is why a bound is needed for those rules. We found this self-energy normalisation preserves the signal through depth better than the two alternatives we tried---a $z$-score output transform, and a detached per-map peak renormalisation whose stop-gradient estimator introduces gradient pathologies in deep stacks---and we use it in the deep MLP, the wide residual network (WRN) and the graph convolutional network (GCN); the shallow convolutional network (CNN), with only two lateral-inhibition sites, does not need it and uses the plain blend ($N\!\equiv\!1$). In the deep MLP the occupancy is finally returned to the activation range by the zero-centring map $\pi$,
\begin{equation}
    y = \pi(\tilde{y}) = 2\tilde{y}-1,
\end{equation}
before being passed to the next layer. This zero-centring is important for training the deep, ten-layer MLP, where the output of one lateral-inhibition layer feeds the next directly: a bounded activation saturates at its rails ($\pm1$), where its derivative vanishes, and the recursive blend otherwise tends to push the signal onto a rail, stalling the backward pass through depth. Returning the output to a zero-centred range keeps activations away from the rails; we find it improves stability and removes the loss spikes otherwise seen with depth. The CNN and WRN pass the output of lateral inhibition on as an occupancy, $y=\pi(\tilde{y})=\tilde{y}\in[0,1]$ ($\pi$ the identity), without this remap: the CNN has only two lateral-inhibition layers, and in the WRN each of their outputs enters a convolution and is renormalised by the next block's group normalisation before any further lateral inhibition. Because $m$ is a function of $x$ only---never of $z$ or of any parameter downstream of it---the regulation is \emph{parameter-free}: the optimiser cannot change the rule, and the network must learn to operate \emph{within} the regulation rather than around it. It can still steer the occupancy a gate reads, through the weights of the layers below (Appendix~\ref{app:steering}). Examples of the Birth and Survival gates and their normalised blend are shown in Figure~\ref{fig:gate_blend}.

\section{Output range of lateral inhibition}
\label{app:range}

The range of the normalised blend depends on the gate functions and their parameters, so we state it for the actual continuous functions. The normalised blend $P/N$ is not a convex combination: $N\in[\tfrac12,1]$ can lift it above $1$, up to $2$ when both gates are $1$. If at every $m$ only one gate were non-zero, the bound would be $\max_{\tilde z}(1-\tilde z)/((1-\tilde z)^2+\tilde z^2)=(1+\sqrt2)/2\approx1.207$ times the peak of that gate. The logistic gates are never exactly zero, so this is an idealisation: even for disjoint birth and survival sets, the tails of the bumps overlap, and bumps of neighbouring counts overlap more. We therefore computed the maximum of $P/N$ over $m$ and $\tilde z$ directly, on a dense grid. At the operating point of \texttt{B5/S3} ($\gamma=15$, $\delta=0.5$) each bump peaks at $0.516$, the two gates overlap only through their tails ($\mathrm{Birth}\cdot\mathrm{Survival}\le0.07$), and the blend is at most $0.656$, so the clip never acts; the measured per-layer maximum of the output on the trained MLP is $\approx0.31$, with no values outside $[-1,1]$ at any of the ten layers. The bound is specific to these parameters: steeper slopes raise each bump's peak towards $1$, and $\tilde y$ can exceed $1$ for $\gamma\gtrsim40$.

\paragraph{Clipping in the rule sweep.} The implementation clips $\tilde y$ to $[0,1]$, so that every layer's output stays in the range the rules are written for. For \texttt{B5/S3} the clip never acts: at $\gamma=15$ the normalised blend is at most $0.66$. It acts only for rules whose birth and survival sets share a count. At a neighbour mean $m$ near such a count both gates are high at once, the numerator $(1-\tilde z)B+\tilde zS$ approaches the common gate value for every state, and dividing by the self-energy $N\ge\tfrac12$ lifts intermediate states above $1$. When the two sets are disjoint, the gates overlap only through their tails, and at $\gamma=15$, $\delta=0.5$ a single bump peaks well below $1$ ($0.52$ for an isolated count). Of the ten rules of the sweep (Section~\ref{sec:rulesweep}), the three with disjoint sets---\texttt{B5/S3}, \texttt{B5/S23} and Seeds (\texttt{B2/S})---have a maximal blend, computed as above, of $0.66$, $0.83$ and $0.62$, and the clip never acts in their trained networks; these values hold for these parameters, not for disjoint sets in general. The other seven share at least one count and reach $1.03$ (\texttt{B3/S3}) to $1.48$ (Day~\&~Night). Day~\&~Night shares four counts ($3,6,7,8$), and its runs of neighbouring counts merge into broad bumps that peak at $0.74$, so it overflows most: in the trained MLP the clip acts on $9\%$ of the units at the first layer, rising to $72\%$ at the tenth. Replacing the clip with a smooth bound, $\tilde y\mapsto\tilde y-\log(1+e^{\beta(\tilde y-1)})/\beta$ with $\beta=10$, which leaves \texttt{B5/S3} unchanged to within $0.003$, changed neither the accuracy nor the coherence measures of any rule beyond the variation between seeds (one seed per rule).

\section{Multi-channel inputs}
\label{sec:channels}

In the notation of Appendix~\ref{sec:formulation}, when the aligned input $\hat{x}$ and the regulated output $z$ share the same channel count, lateral inhibition is applied per channel. When they differ, every lateral-inhibition layer that reads a learned feature map uses the channel mean of the occupancies, $\bar{m}=\tfrac{1}{C_\mathrm{in}}\sum_c m_c$. The lateral-inhibition layer that reads the raw image needs more care. With input normalisation $x=(p-\mu)/\sigma$ of a pixel value $p\in[0,1]$, the occupancy the rule reads is
\begin{equation}
    \tilde{x}=\phi(x)=\frac{p-\mu}{2\sigma}+\frac12 ,
    \label{eq:occupancy-normalisation}
\end{equation}
which for the MNIST family ($\mu=\sigma=\tfrac12$) is the raw intensity: the black background sits at occupancy $0$, below every band, and only the object's contour reaches the survival ($3/8$) and birth ($5/8$) bands. In a colour image each channel is an affine copy of the object with its own background, contrast and sign, and the channel mean lets opposite-polarity channels cancel: for a gold digit on a blue background the mean occupancy is $0.38$ on the background, at the survival set-point, so the survival gate fires over the whole background and the contour signal is lost. At the layer that reads the image we therefore normalise each channel to its own background and contrast, apply the rule per channel and combine afterwards:
\begin{equation}
\begin{gathered}
    o_c=\frac{\lvert m_c-\mathrm{med}_c\rvert}{\kappa_c},\qquad \kappa_c=\max\lvert m_c-\mathrm{med}_c\rvert,\\
    B=\sum_c w_c\,\mathrm{Birth}(o_c),\qquad S=\sum_c w_c\,\mathrm{Survival}(o_c),\qquad w_c=\frac{\kappa_c}{\sum_{c'}\kappa_{c'}} ,
\end{gathered}
    \label{eq:perchannel-gate}
\end{equation}
with the median and maximum taken over the positions of one image; $\kappa_c$ is bounded below by $10^{-6}$. The median stands for the background and maps to $0$, the strongest deviation of either sign to $1$, and weighting by contrast keeps a structureless channel from contributing amplified noise. In the backward pass we hold $\mathrm{med}_c$, $\kappa_c$ and $w_c$ fixed. Because this layer reads the image, the statistics depend on no weight, so the choice leaves training unchanged; it affects only gradients with respect to the input, which are then a fixed-statistic surrogate rather than the derivative of the forward map. The two differ: the forward map is invariant to scaling an image's contrast about its median, which the surrogate gradient does not reflect (Appendix~\ref{app:perchgrad}). The normalisation also makes the layer blind to the absolute contrast of the image: an image with no contrast ($\kappa_c<10^{-6}$ in every channel) switches every gate off, while residual variation just above that bound is stretched to full contrast. For a two-colour image every channel becomes the grayscale object; on black-background grayscale the transform is close to the identity. Three lessons emerged: combine channels after the rule, not before (birth and survival responses are non-negative and cannot cancel); confine colour handling to the layer that reads the image (applying it at the layers that read features cost one to three points of clean accuracy); and keep learnable layers out of the path of lateral inhibition (a learned colour stem let the network route colour around the rule, and a learned convolutional stem in front of the first lateral-inhibition layer removed the object localisation of the saliency).

The per-image median assumes the background is the majority of the image. When the object fills much of the frame (FashionMNIST) the network with lateral inhibition pays three to four points of clean accuracy; cluttered natural images have no single background, so we use the per-channel image normalisation only for the MNIST family, while in the WRN the stem convolution precedes every lateral-inhibition layer and none reads the image. The colour datasets (``MNIST-colour'', ``FashionMNIST-colour'', ``KMNIST-colour'', ``EMNIST-colour'') render each grayscale image with a fixed blue background and gold ink.

\section{Control experiments}
\label{app:controls}

All experiments use the 10-layer MLP on MNIST with the paper
recipe unless noted. Notation: \textbf{LFPR} is the low-frequency power ratio of the class-averaged
final-hidden activation maps (random/permuted null $\approx 0.087$; class-mean-input ceiling $0.971$);
\textbf{lens@$\ell$} is logit-lens accuracy, the trained final classifier applied at hidden layer
$\ell$; \textbf{SSI} is the scale-separation index; \textbf{iso} is the fraction of co-activation-graph
nodes that are isolated; \textbf{grid-local} is the fraction of graph edges within a $3\times3$ Moore
neighbourhood. Single seed unless a seed count is given. The controls of \S\ref{app:frame} and \S\ref{app:gridrule} are trained with the paper recipe and training loop at seeds 0, 1 and 2 and saved; their with/without-lateral-inhibition rows are seed replicates of the paper pair. Values are means over the three seeds.

\subsection{The input gradient of the image normalisation is a surrogate}
\label{app:perchgrad}

The CNN's first lateral-inhibition layer normalises each image channel by its median and contrast (Eq.~\eqref{eq:perchannel-gate}), and these statistics are held fixed in the backward pass. This changes no weight gradient, but the gradient with respect to the image is then a fixed-statistic surrogate rather than the derivative of the forward map, and it is this gradient that the saliency maps of Section~\ref{sec:saliency} show. We compared it with the exact derivative, obtained by keeping the statistics in the graph (the median and maximum pass their gradient to the selected position), and with finite differences. None of this changes the trained networks; the forward map is the same in all cases.

\paragraph{The gate alone.} Scaling the contrast of an occupancy map about its median leaves the normalised occupancy, and so the gates, unchanged. Along that direction the finite-difference derivative of $\sum(B+S)$ is $0$; the exact gradient gives $0$ as well, the surrogate $-9.0$. Along a random direction the finite difference is $-2.86$, the exact gradient $-2.86$ and the surrogate $-1.49$.

\paragraph{Trained networks.} Table~\ref{tab:perchgrad} repeats the saliency analysis of Section~\ref{sec:saliency} with the exact derivative, for the CNN with lateral inhibition on all eight datasets (five seeds, first $1000$ test images, true class). The exact maps are weaker on the object than the surrogate ones (log-ratio lower by $0.04$--$0.18$) and differ from them in magnitude by about half to three quarters of their norm, but they rank the pixels almost identically, and every conclusion of Section~\ref{sec:saliency} holds: every network favours the object, in $91$--$100\%$ of images (more than with the surrogate). Along the direction of contrast scaling the exact gradient agrees with finite differences to $10^{-7}$, whereas the surrogate is off by $3.6$ to $6.6$ times the true value. Along a random direction neither agrees well. The forward map is not differentiable at these images: the median falls on a tie, since many background positions share the same occupancy ($70\%$ of positions on MNIST, $17\%$ on Fashion-MNIST), so the derivative differs between the two sides of the image and no gradient reproduces the central finite difference.

\begin{table}[h]\centering\small
\setlength{\tabcolsep}{4pt}
\begin{tabular}{lcccccccccc}
\hline
 & \multicolumn{2}{c}{log-ratio} & \multicolumn{2}{c}{favours object (\%)} & rank & rel.\ & \multicolumn{2}{c}{FD error, contrast dir.} & \multicolumn{2}{c}{FD error, random dir.\ (\%)} \\
dataset & surr. & exact & surr. & exact & corr. & diff. & surr.\ (\%) & exact & surr. & exact \\
\hline
MNIST & $+1.13$ & $+0.95$ & 100 & 100 & 0.97 & 0.75 & 657 & $10^{-7}$ & 25 & 36 \\
Fashion-MNIST & $+0.64$ & $+0.59$ & 83 & 91 & 0.97 & 0.70 & 623 & $10^{-7}$ & 57 & 17 \\
KMNIST & $+0.74$ & $+0.66$ & 95 & 97 & 0.97 & 0.57 & 570 & $10^{-7}$ & 35 & 17 \\
EMNIST & $+0.91$ & $+0.81$ & 100 & 100 & 0.97 & 0.61 & 473 & $10^{-8}$ & 41 & 11 \\
MNIST (colour) & $+0.98$ & $+0.87$ & 100 & 100 & 0.97 & 0.65 & 361 & $10^{-7}$ & 39 & 28 \\
Fashion-MNIST (colour) & $+0.58$ & $+0.54$ & 86 & 91 & 0.95 & 0.60 & 471 & $10^{-7}$ & 37 & 7 \\
KMNIST (colour) & $+0.64$ & $+0.59$ & 95 & 97 & 0.97 & 0.48 & 408 & $10^{-7}$ & 46 & 21 \\
EMNIST (colour) & $+0.77$ & $+0.71$ & 100 & 100 & 0.96 & 0.57 & 432 & $10^{-8}$ & 44 & 15 \\
\hline
\end{tabular}
\caption{Saliency of the CNN with lateral inhibition with the surrogate gradient (surr., statistics of the image normalisation held fixed, as in Section~\ref{sec:saliency}) and with the exact derivative through the statistics (five seeds, first $1000$ test images, true class). Log-ratio and favours object as in Section~\ref{sec:saliency}. Rank corr.: median over images of the Spearman correlation between the two maps; rel.\ diff.: median of $\lVert s_\text{surr}-s_\text{exact}\rVert/\lVert s_\text{surr}\rVert$. FD error: median relative error of the directional derivative of the true-class logit from autograd against a central finite difference (float64, step $10^{-5}$, $50$ images of the seed-$0$ network), along the scaling of the image's contrast about its per-channel median and along a random unit direction.}
\label{tab:perchgrad}
\end{table}

\paragraph{Constant and near-constant images.} The contrast $\kappa_c$ is bounded below by $10^{-6}$, and below that bound the gates read no structure. We fed the MNIST and colour MNIST networks (seed $0$, $200$ images) a uniform grey image with added uniform noise of amplitude $a$, and test digits with their contrast about grey scaled by $a$. For a constant image ($a=0$, and $a=10^{-8}$, which float32 does not resolve) the gates are zero, every unit of the first layer sits at $-1$, and every image is assigned to the same class with probability $1.00$, a confident output for an image without content. Once the contrast exceeds the bound ($a\ge10^{-6}$) the normalisation stretches it to full range: the mean birth and survival gates are the same for noise of amplitude $10^{-5}$ as for $10^{-1}$ ($0.07$ and $0.19$), and a digit whose contrast is reduced to $10^{-5}$, invisible to the eye, is still classified correctly in $80\%$ of cases (MNIST; $92\%$ in colour) against $94\%$ and $99\%$ at $10^{-1}$. In this range the input gradient grows as $1/a$ in both the surrogate and the exact form, from ${\sim}30$ at $a=10^{-1}$ to ${\sim}10^{6}$ at $a=10^{-6}$. The layer that reads the image is thus blind to the absolute contrast of its input, which the noise experiments do not probe (none of their corruptions lowers the contrast), and its behaviour at vanishing contrast is set by the bound rather than by the image.

\subsection{The rule versus the layer input: blur rules within the same approach}
\label{app:frame}

Which properties come from the CA rule, and which from regulating the layer output by the layer input $x$? We compare lateral inhibition with two rule-free local averages (``blurs''). The \emph{input blur}, $y=(1-\lambda)z+\lambda\,\pi(m)$ with $m$ the Moore-neighbour mean of the input occupancy $\tilde x$, keeps the wiring of lateral inhibition and reads the same input. It is not a baseline from outside this work: it regulates the layer output by the layer input as lateral inhibition does, with the CA rule replaced by a linear blend, and so it is another rule within the same approach. The \emph{output blur}, $y=(1-\lambda)z+\lambda\,\pi(m_z)$ with $m_z$ the same neighbour mean computed from the output occupancy $\tilde z$, smooths the output just as much but never sees $x$. The pair separates the role of the rule (lateral inhibition against the input blur) from the role of reading the input (input blur against output blur). A random-neighbour regulation (rules kept, grid destroyed) completes the comparison.

\begin{table}[h]\centering\small
\begin{tabular}{lccccccc}
\hline
arm & frame & rules & val & LFPR & lens@1 & lens@10 & SSI \\
\hline
with lateral inhibition (B5/S3) & grid & yes & 0.962 & 0.941 & 57.5 & 96.5 & 18.2 \\
plain & none & -- & 0.984 & 0.090 & 9.6 & 98.4 & 1.1 \\
input blur ($\lambda=0.5$) & grid & no & 0.987 & 0.132 & 10.9 & 97.8 & 3.5 \\
output blur ($\lambda=0.5$) & grid & no & 0.984 & 0.722 & 10.4 & 97.6 & 1.0 \\
random-neighbour & random & yes & 0.301 & 0.101 & 10.1 & 22.8 & 1.1 \\
\hline
\end{tabular}
\caption{With lateral inhibition and plain: the paper's five networks; other arms: three networks each. Neither blur at $\lambda=0.5$ reproduces the coherence or the consistency across depth of lateral inhibition; the output blur smooths the final maps (LFPR $0.72$) without consistency across depth or scale separation; a random-neighbour regulation does not train (gain collapse, see \S\ref{app:collapse}).}
\end{table}

\begin{table}[h]\centering\small
\begin{tabular}{lcccc}
\hline
arm & val & LFPR & lens@1 & SSI \\
\hline
with lateral inhibition & 0.962 & 0.941 & 57.5 & 18.2 \\
plain & 0.984 & 0.090 & 9.6 & 1.1 \\
input blur $\lambda=0.25$ & 0.985 & 0.097 & 8.6 & 1.7 \\
input blur $\lambda=0.50$ & 0.987 & 0.132 & 10.9 & 3.5 \\
input blur $\lambda=0.75$ & 0.986 & 0.274 & 31.5 & 14.6 \\
input blur $\lambda=0.80$ & 0.985 & 0.350 & 49.0 & 15.9 \\
input blur $\lambda=0.85$ & 0.985 & 0.702 & 73.9 & 18.4 \\
input blur $\lambda=0.90$ & 0.981 & 0.967 & 75.9 & 18.8 \\
input blur $\lambda=0.95$ & 0.955 & 0.997 & 79.7 & 19.0 \\
input blur $\lambda=1.00$ & 0.893 & 1.000 & 80.0 & 19.0 \\
output blur $\lambda=0.50$ & 0.984 & 0.722 & 10.4 & 1.0 \\
output blur $\lambda=0.75^{\dagger}$ & 0.984 & 0.888 & 6.9 & 1.0 \\
output blur $\lambda=0.90^{\dagger}$ & 0.984 & 0.941 & 17.4 & 1.0 \\
output blur $\lambda=0.95^{\dagger}$ & 0.980 & 0.928 & 12.9 & 1.0 \\
output blur $\lambda=1.00^{\dagger}$ & 0.983 & 0.932 & 18.0 & 1.3 \\
shuffled-$m$ & \multicolumn{4}{c}{did not train (val $0.112$, chance)} \\
\hline
\end{tabular}
\caption{Blend-weight frontier of the two blurs (with lateral inhibition and plain: the paper's five networks; input-blur arms and the output blur at $\lambda=0.5$: three networks each; $^{\dagger}$one network). At $\lambda=0.9$ the input blur \emph{exceeds} lateral inhibition on accuracy
($0.981$ vs $0.962$), LFPR ($0.967$ vs $0.941$) and lens@1 ($75.9$ vs
$57.5$), and both reach the ceiling of the SSI ($18.8$ and $18.2$, against a maximum of $19$; Appendix~\ref{app:ssi}), where differences are not meaningful. The output blur reaches the same smooth final maps (LFPR up to $0.94$) at no cost in accuracy, but at no blend weight does it make the early layers readable by the final classifier (lens@1 $7$--$18$, as for the plain network) or separate scales (SSI ${\approx}1$). Consistency across depth and scale separation on the MLP therefore come from regulating the output by the layer input, which lateral inhibition and the input blur share, not from the CA rule, and not from smoothing alone. (shuffled-$m$
did not train, like random-neighbour, so is inconclusive.)}
\end{table}

\begin{figure}[h]
\centering
\includegraphics[width=\textwidth]{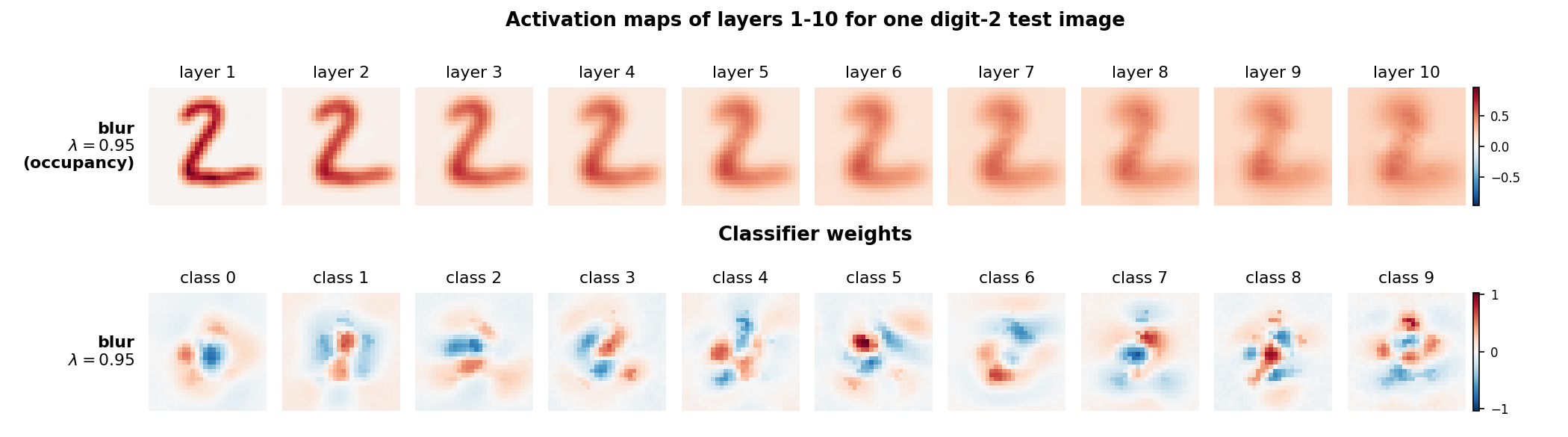}
\caption{\textbf{Activation maps and classifier weights of the input blur (MNIST MLP, $\lambda=0.95$).} The counterpart of Figure~\ref{fig:act_maps}. Top: the activation maps of hidden layers $1$--$10$ for the same digit-2 test image, shown as the occupancy $(y+1)/2$. Bottom: the classifier weight map of each class. One signed colour scale per row (white is zero).}
\label{fig:blur_maps}
\end{figure}

\paragraph{What the input blur's coherence looks like.} Figure~\ref{fig:blur_maps} is the counterpart of Figure~\ref{fig:act_maps} for the trained blur at $\lambda=0.95$ (seed $0$; $94.5\%$ on the first $4000$ test images), on the same digit-2 test image. Each hidden layer is a progressively softer copy of the digit, the input blurred once more at every layer, rather than the blob-like regions that lateral inhibition builds and reshapes with depth. Its maps and classifier weights are smooth by construction (activation LFPR $0.998$, classifier-weight LFPR $0.946$, against $0.96$ and $0.70$ for the network with lateral inhibition): the coherence is inherited from the input, as the agreement of its untrained seeds below confirms.

\paragraph{FiLM-gain identity.} Both regulations scale and shift each unit of the trunk output, $y=\gamma\,z+\beta$, with $\beta$ a
function of the input $m$. The blur has a \emph{constant} gain $\gamma=(1-\lambda)$; lateral inhibition (plain
blend) an \emph{input-adaptive} gain $\gamma=S(m)-B(m)$. Measured at layer~1 on $2000$ MNIST test images (model-free), $\gamma_{\rm
CA}$ has $\mathrm{mean}|\gamma|=0.069$ but is $8.5\times$ larger on stroke pixels (pixel value $>0.1$; $0.262$) than off them
($0.031$), and exceeds $0.01$ on $29\%$ of the field. To reach the same low average gain the blur must shrink
$\gamma$ uniformly ($\lambda\to1$, trunk killed everywhere); lateral inhibition concentrates it on informative
positions instead.

\paragraph{Coherence across the ensemble.} The same controls, measured as in Section~\ref{sec:scalesep}: unit-wise agreement between the deep hidden layers (6--10) of networks trained from different seeds, and between the same seeds' untrained initialisations, and the test accuracy of the network obtained by averaging the three seeds' weights.

\begin{table}[h]\centering\small
\begin{tabular}{lccc}
\hline
regulation & agreement, trained & agreement, untrained & weight average (\%) \\
\hline
with lateral inhibition (B5/S3) & 0.83 & 0.11 & 88.8 \\
input blur $\lambda=0.5$ & 0.01 & 0.01 & 27.7 \\
input blur $\lambda=0.75$ & 0.22 & 0.04 & 28.8 \\
input blur $\lambda=0.9$ & 0.97 & 0.74 & 89.1 \\
input blur $\lambda=0.95$ & 1.00 & 0.96 & 94.2 \\
plain & 0.00 & 0.00 & 10.1 \\
\hline
\end{tabular}
\caption{Deep-layer (6--10) unit-wise agreement between seeds, and accuracy of the network whose weights are the average of the seeds' weights. With lateral inhibition and plain: the paper's five networks; blur arms: three networks each. Strong blurs make independently trained networks agree, but they agree already at initialisation: each layer is largely a blurred copy of its input. The agreement under lateral inhibition is learned. Weaker blurs are not coherent across seeds.}
\end{table}

\subsection{Grid versus rule for the co-activation-graph structure}
\label{app:gridrule}

\begin{table}[h]\centering\small
\begin{tabular}{lcccc}
\hline
regulation & val & iso & $Q$ (whole-graph) & grid-local \\
\hline
with lateral inhibition (B5/S3) & 0.962 & 0.24 & 0.82 & 1.00 \\
plain & 0.984 & 0.69 & 0.75 & 0.01 \\
plain, active units only & -- & 0.24 & 0.50 & 0.01 \\
local-average (grid, no rules) & 0.984 & 0.25 & 0.88 & 0.35 \\
random-neighbour (rules, no grid) & 0.301 & 0.82 & 0.33 & 0.01 \\
\hline
\end{tabular}
\caption{With lateral inhibition and plain: the paper's five networks; other arms, and plain on active units only: three networks each. $Q$ is the modularity of greedy-modularity communities for every arm. A random neighbourhood fragments the co-activation graph (isolates $0.82$), while a rule-free (shallow) local average of $z$ is as connected as with lateral inhibition
(isolates $0.25$ vs $0.24$) and more modular ($0.88$). The plain network's high isolate fraction is largely an effect of its hidden units that no longer respond to the input ($44$--$47\%$ of the sampled units; Appendix~\ref{app:percolation}): counting active units only, its isolates fall to $0.24$, as with lateral inhibition, and its modularity to $0.50$. On active units the two differ in modularity ($0.82$ vs $0.50$), not in connectedness. The deep rule-free local average of
\S\ref{app:frame} reaches active-map coherence $0.75$ and data-alignment $0.43$ at $\lambda=0.5$ and exceeds lateral inhibition at
$\lambda=0.9$ ($0.96$ / $0.61$ vs $0.92$ / $0.50$) --- these are frame/wiring properties, not rule properties.}
\end{table}

\subsection{Representational collapse is conditional}
\label{app:collapse}

\begin{table}[h]\centering\small
\begin{tabular}{lccccc}
\hline
dataset & classes & eff.\ rank (with/plain) & train acc (with/plain) & iso (with/plain) \\
\hline
MNIST & 10 & 10.9 / 8.7 & 0.999 / 1.000 & 0.22 / 0.72 \\
KMNIST & 10 & 12.1 / 8.7 & 0.999 / 1.000 & 0.25 / 0.71 \\
FashionMNIST & 10 & 9.9 / 6.8 & 0.996 / 0.971 & 0.20 / 0.38 \\
EMNIST & 47 & 19.3 / 16.1 & 0.913 / 0.930 & 0.24 / 0.21 \\
\hline
\end{tabular}
\caption{``With'': with lateral inhibition; ``plain'': without. The plain network collapses (high isolates, low effective rank) only where training fits
easily (few classes, perfect train accuracy); on 47-class EMNIST it does not collapse and has fewer
isolates than with lateral inhibition, whose isolate fraction is dataset-invariant. ``Lateral inhibition prevents
collapse'' therefore holds only where training would otherwise produce it (consistent with neural
collapse). The plain network's isolate fractions on MNIST and KMNIST come largely from hidden units that no longer respond to the input; among active units they are $0.17$ and $0.29$ (Appendix~\ref{app:percolation}).}
\end{table}

\subsection{Coherence is present at initialisation and reproduced by lateral inhibition alone}
\label{app:init}

Active-map coherence and data-alignment of the MLP with lateral inhibition: untrained (random init) $0.95$ / $0.73$;
trained $0.92$ / $0.50$ (five networks); lateral inhibition applied to the raw data with \emph{no network} gives $1.00$ / $0.97$
(= the data-variance map). Coherence and data-alignment are thus properties of the input-aligned
wiring, present before training, not products of learning.

\subsection{Scale-separation index: definition, horizon and censoring}
\label{app:ssi}

Each hidden layer $\ell=1,\dots,L$ of the MLP is a map on the $28\times28$ grid. We take its two-dimensional Fourier transform, centre the zero frequency (bin $14$ after the shift), and band-pass it with six radial masks $r\in[0,2),[2,4),[4,6),[6,9),[9,13),[13,20)$, where $r$ is the distance from the zero-frequency bin in Fourier bins (cycles per $28$ pixels; wavelengths in Table~\ref{tab:ssi}). Every mask is conjugate symmetric, so each band-passed map is real. For band $b$ and image $i$ let $v_{b,\ell,i}$ be the band-passed map of layer $\ell$ with its mean removed, normalised to unit length. The across-layer correlation at depth lag $k$ is
\begin{equation}
C_b(k)=\frac{1}{(L-k)\,n}\sum_{\ell=1}^{L-k}\sum_{i=1}^{n}\langle v_{b,\ell,i},\,v_{b,\ell+k,i}\rangle ,
\end{equation}
so $C_b(0)=1$, and the correlation depth is $\tau_b=\tfrac12+\sum_{k=1}^{K}C_b(k)$, the sum stopped before the first $k$ with $C_b(k)\le0$. The scale-separation index is $\mathrm{SSI}=\tau_\text{coarse}/\tau_\text{fine}$, the ratio for the first and the last band. We use $n=1000$ test images and the same horizon $K=9$ at every depth: a ten-layer network has no larger lag, and for the $100$-layer network $C_b(k)$ is averaged over all $100-k$ layer pairs at each lag up to $9$. With this horizon $\tfrac12\le\tau_b\le9.5$ and $\mathrm{SSI}\le19$. A band that is still positively correlated at lag $9$ has a right-censored correlation depth: $\tau_b$ is then a lower bound set by the horizon, not an estimate of how long the band persists.

\begin{table}[h]\centering\small
\begin{tabular}{lccccccccc}
\hline
 & & \multicolumn{3}{c}{with lateral inhibition, depth 10} & \multicolumn{2}{c}{plain, depth 10} & \multicolumn{3}{c}{with lateral inhibition, depth 100} \\
band $r$ & wavelength (px) & $\tau_b$ & $C_b(1)$ & $C_b(9)$ & $\tau_b$ & $C_b(1)$ & $\tau_b$ & $C_b(1)$ & $C_b(9)$ \\
\hline
$[0,2)$ & $>14$ & 9.11$^{*}$ & $+1.00$ & $+0.92$ & 0.54 & $-0.01$ & 8.85$^{*}$ & $+0.99$ & $+0.87$ \\
$[2,4)$ & 7--14 & 8.22$^{*}$ & $+0.98$ & $+0.77$ & 0.51 & $-0.01$ & 8.17$^{*}$ & $+0.98$ & $+0.71$ \\
$[4,6)$ & 4.7--7 & 4.68$^{*}$ & $+0.90$ & $+0.18$ & 0.52 & $+0.01$ & 5.81$^{*}$ & $+0.93$ & $+0.27$ \\
$[6,9)$ & 3.1--4.7 & 2.61 & $+0.72$ & $-0.05$ & 0.51 & $+0.00$ & 3.83$^{*}$ & $+0.82$ & $+0.06$ \\
$[9,13)$ & 2.2--3.1 & 1.00 & $+0.17$ & $-0.08$ & 0.50 & $-0.00$ & 1.88 & $+0.46$ & $-0.05$ \\
$[13,20)$ & 1.4--2.2 & 0.50 & $-0.17$ & $+0.02$ & 0.50 & $-0.00$ & 1.03 & $+0.18$ & $-0.02$ \\
\hline
SSI & & \multicolumn{3}{c}{$18.2\pm0.1$} & \multicolumn{2}{c}{$1.1\pm0.1$} & \multicolumn{3}{c}{$8.6$} \\
\hline
\end{tabular}
\caption{Per-band correlation depths and correlations of the MNIST MLP (depth 10: five networks, means; depth 100: one network; the plain network was not trained at depth 100). $^{*}$Censored: $C_b(k)>0$ at every lag up to $9$, so $\tau_b$ is a lower bound. $1000$ test images.}
\label{tab:ssi}
\end{table}

With lateral inhibition the three coarsest bands are censored at depth 10, and the SSI of $18.2$ is close to the ceiling of $19$; it shows that the coarse maps persist through the whole stack while the finest change at every layer, but it does not resolve how much longer the coarse maps would persist in a deeper network. The $100$-layer network, measured over the same horizon, keeps its coarse bands about as persistent ($C(9)=0.87$ against $0.92$) but its fine bands change more slowly ($\tau_\text{fine}=1.03$ against $0.50$), so its SSI is lower, $8.6$; a value of ${\approx}29$ given in an earlier version used a $40$-layer horizon and is not comparable. Several regulations reach the ceiling (input blurs with $\lambda\ge0.85$, SSI $18.4$--$19.0$, Appendix~\ref{app:frame}), and differences between them are not meaningful at this horizon. An earlier implementation centred the masks half a bin off the zero frequency, which made them slightly asymmetric; correcting this changes the SSI by at most $0.04$ ($18.14$ to $18.16$ for the network of Figure~\ref{fig:depth_diag}), and every value in the paper uses the corrected definition.

\subsection{Saliency: interpretability is a property of the rule, not the wiring}
\label{app:saliency}

\paragraph{Protocol.} The saliency $s=|\partial\,\mathrm{logit}_y/\partial x|$ of the true class $y$ is summed over channels. The object mask is the set of pixels above $0.1$ (grayscale) or above an Otsu threshold of the luminance (colour), dilated by one pixel. Images whose mask covers under $3\%$ or over $97\%$ of the frame, or whose saliency vanishes ($\max s<10^{-8}$), would be excluded and counted; no trained network had any. Besides the log-ratio $\log_{10}(\overline{s}_\text{on}/\overline{s}_\text{off})$ we report the area-normalised object mass (the share of saliency on the mask divided by the mask's share of the image; $1$ for uniform saliency) and the fraction of images favouring the object. We use the first $1000$ test images and five networks per dataset (s.d.\ of the mean log-ratio over seeds ${\le}0.02$ with lateral inhibition). The CNN's mean log-ratio is $+0.64$ to $+1.13$ on the four grayscale datasets and $+0.58$ to $+0.98$ on their colour versions, with $83$--$100\%$ of images favouring the object (fewest on Fashion-MNIST); the plain CNN reaches $-0.13$ to $+0.47$ and $-0.06$ to $+0.70$, and on EMNIST favours the background in $94\%$ of images. The MLP reaches $+0.44$ to $+0.71$, against $+0.03$ to $+0.30$ plain.

\begin{figure}[h]\centering
\includegraphics[width=\textwidth]{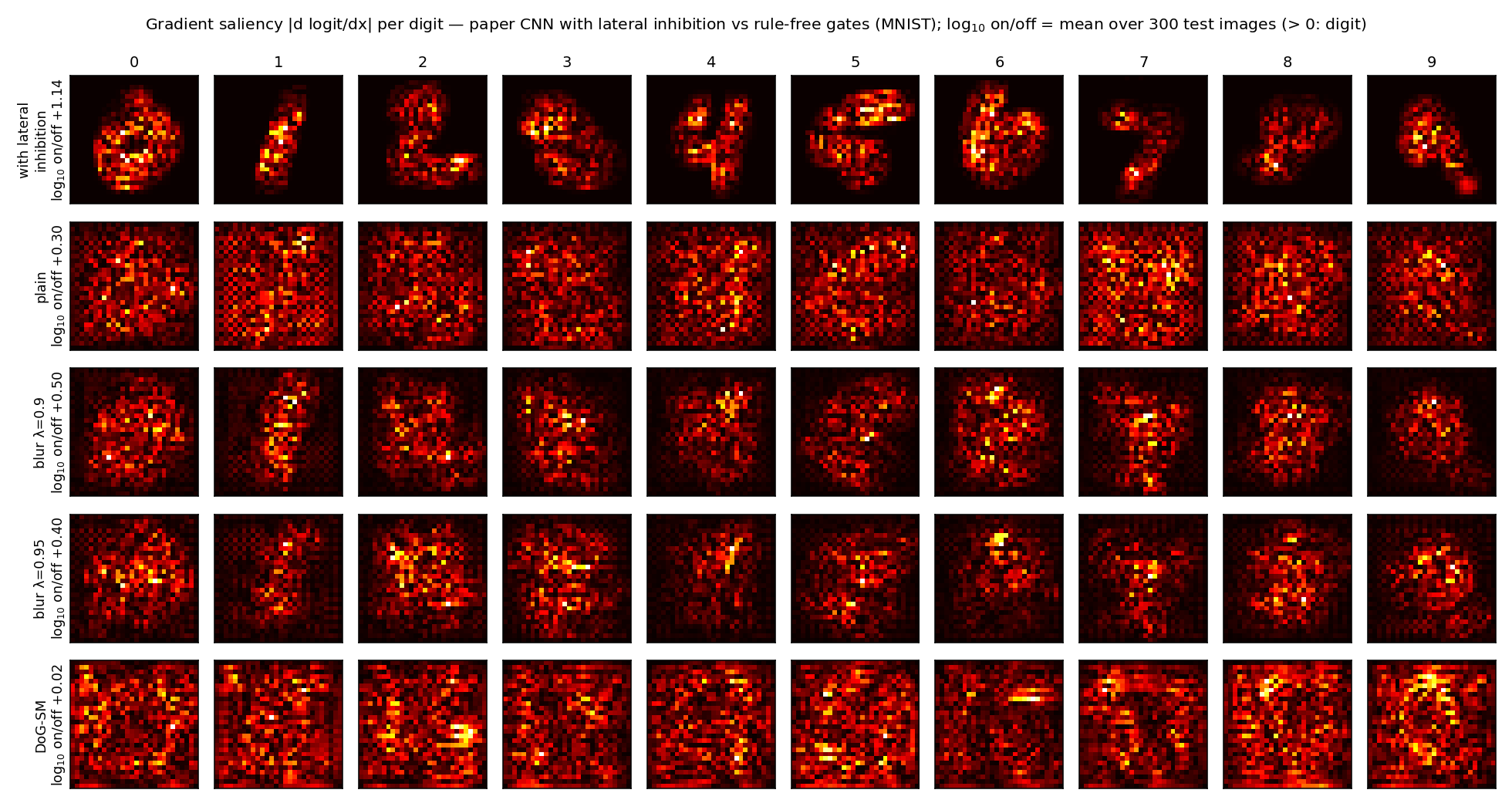}
\caption{\textbf{Gradient saliency per digit (paper CNN, MNIST).} Lateral inhibition (top row) concentrates
saliency on the digit (log-ratio $\log_{10}(\overline{s}_\text{on}/\overline{s}_\text{off})=+1.13$, saliency per pixel $13$ times higher on the digit; mean over $1000$ test images and five networks); the plain network ($+0.47$), both input blurs ($+0.48$, $+0.37$) and DoG-SM ($+0.04$) remain diffuse, although DoG-SM
recovers most of the noise gain of lateral inhibition. Interpretability comes from the rule, not from the shared
input-to-output wiring nor from lateral filtering.}
\label{fig:blursal}
\end{figure}

On the paper CNN (MNIST), we replaced both lateral-inhibition layers with rule-free alternatives and trained each with the paper recipe (five seeds each; the CA and plain networks are the paper's five CNNs). The linear blur regulation $y=(1-\lambda)z+\lambda\,\pi(m)$ has the \emph{same} input-to-output wiring as lateral inhibition and reads the same occupancy $m$ (the per-channel normalised image at the first layer), but has no birth--survival rule. The output blur $y=(1-\lambda)z+\lambda\,\pi(m_z)$ (Appendix~\ref{app:frame}) is the same blend computed from the layer output instead of the input, and the published difference-of-Gaussians surround modulation (DoG-SM)~\citep{Hasani2019} likewise filters the layer output $z$ only. Table~\ref{tab:blursal} gives held-out clean and noise accuracy and the signed log-ratio and region separability of gradient saliency ($|\partial\,\mathrm{logit}_y/\partial x|$, Section~\ref{sec:saliency}) over the first $1000$ test images, and Figure~\ref{fig:blursal} shows the saliency maps (its row labels: log-ratio over $300$ correctly classified test images of one network).

\begin{table}[h]\centering\small
\begin{tabular}{lcccc}
\hline
regulation & clean acc (\%) & noise acc (\%) & saliency log-ratio & region sep. \\
\hline
CA (with lateral inhibition) & 98.6 & 91.7 & $\mathbf{+1.13}$ & \textbf{14.5} \\
plain & 98.9 & 76.4 & $+0.47$ & 3.2 \\
input blur, $\lambda=0.9$ & 98.8 & 87.2 & $+0.48$ & 3.1 \\
input blur, $\lambda=0.95$ & 97.9 & 90.9 & $+0.37$ & 2.4 \\
output blur, $\lambda=0.9^{\dagger}$ & 99.3 & 81.4 & $+0.24$ & 1.7 \\
output blur, $\lambda=0.95^{\dagger}$ & 99.2 & 83.4 & $+0.31$ & 2.1 \\
DoG-SM & 99.1 & 86.7 & $+0.04$ & 1.1 \\
\hline
\end{tabular}
\caption{Paper CNN on MNIST with lateral inhibition replaced by rule-free regulations; mean over five seeds (s.d.\ ${\le}1.0$ point for noise accuracy, ${\le}0.15$ for the log-ratio and ${\le}1.0$ for region separability, $0.1$ for lateral inhibition). Noise accuracy: Gaussian, shot and impulse noise at the five standard severities. Saliency over the first $1000$ test images (true class): the signed log-ratio $\log_{10}(\overline{s}_\text{on}/\overline{s}_\text{off})$ ($>0$: more saliency per pixel on the digit) and the symmetric region separability $\max(\text{on}/\text{off},\text{off}/\text{on})$; every network favours the digit on average. $^{\dagger}$One network (seed $0$); the input blur of the same seed scores $85.9\%$ and $89.6\%$ noise accuracy at $\lambda=0.9$ and $0.95$.}
\label{tab:blursal}
\end{table}

The two blurs separate the role of the layer input from that of the rule. Reading the input carries most of the noise resilience: the stronger input blur recovers nearly all of the gain of lateral inhibition ($90.9\%$ against $91.7\%$, at $0.7$ points lower clean accuracy), from $76.4\%$ for the plain network, whereas the output blur, which smooths as much but never sees $x$, recovers well under half of it at the same blend weights ($81.4\%$ and $83.4\%$; $85.9\%$ and $89.6\%$ for the input blur of the same seed), and nothing at $\lambda=0.5$ ($76.7\%$, as the plain network). DoG-SM, a stronger filter of the output, recovers most of the gain ($86.7\%$), so input regulation is not the only route to noise resilience, but it is the one that a blur needs. The object-localised saliency, by contrast, needs the rule: neither blur reproduces it, whether it reads the input (log-ratio $+0.37$ to $+0.48$) or the output ($+0.24$ to $+0.31$), nor does DoG-SM ($+0.04$), against $+1.13$ with lateral inhibition. On the MLP the picture is the same for saliency (log-ratio $+0.72$ with lateral inhibition, $+0.07$ to $+0.36$ for the input blur, $+0.26$ to $+0.28$ for the output blur, $+0.30$ plain; one network each), while noise accuracy separates nothing there: at an equal training budget the plain MLP is the most robust ($97.1\%$), the output blur stays at its level ($96.5\%$) and the input blur and lateral inhibition trade robustness for their structure ($92.7$--$96.1\%$ and $85.2\%$).

\paragraph{Checks of the saliency result.} Table~\ref{tab:salchecks} collects the checks reported in Section~\ref{sec:saliency}: a causal test without gradients, the gradient with every gate route detached (the exact trunk-route term of the backward pass, Appendix~\ref{app:li-backward}), untrained networks, and saliency split by region (digit = pixels $>0.1$; far background = at least three pixels from the digit). In each noise condition the same number of pixels is replaced by uniform noise.

\begin{table}[h]\centering\small
\begin{tabular}{lcccc}
\hline
 & CNN, with & CNN, plain & MLP, with & MLP, plain \\
\hline
clean accuracy (\%) & 98.8 & 99.0 & 96.5 & 98.4 \\
object pixels noised (\%) & 60.5 & 74.4 & 85.3 & 94.4 \\
as many background pixels noised (\%) & 96.3 & 67.0 & 79.8 & 96.9 \\
\hline
saliency log-ratio $\log_{10}(\overline{s}_\text{on}/\overline{s}_\text{off})$, full gradient & $+1.13$ & $+0.47$ & $+0.71$ & $+0.30$ \\
\quad gate routes detached (trunk only) & $+1.28$ & -- & vanishes & -- \\
\quad untrained network & $+0.89$ & $+1.09$ & $+0.36$ & $+0.00$ \\
images favouring the object (\%) & 100 & 100 & 100 & 100 \\
area-normalised object mass & 3.36 & 2.00 & 2.55 & 1.58 \\
region separability $\max(\text{on}/\text{off},\text{off}/\text{on})$ & 14.5 & 3.2 & 5.3 & 2.0 \\
saliency share on far background (\%) & 1 & 13 & 5 & 18 \\
\hline
\end{tabular}
\caption{``With'': with lateral inhibition; ``plain'': without. Checks of the saliency result (MNIST, five seeds, all $1000$ test images; s.d.\ over seeds $\le2.1$ points for accuracies, $\le0.15$ for log-ratios and $\le1.1$ for region separability). Saliency of the true class; log-ratio $>0$ means more saliency per pixel on the object. ``Vanishes'': the MLP's input gradient with every gate route detached is at rounding level ($4\times10^{-13}$ per pixel against $4\times10^{-2}$), so no ratio is defined; the untrained MLP with lateral inhibition has $5$ such images, excluded. Region separability is the symmetric score, which does not tell object from background. The CNN with lateral inhibition relies on the object, ignores the far background, and its localisation survives with the gate routes detached; the MLP's localised gradient comes from the gate route, and it is fragile to noise in the edge ring.}
\label{tab:salchecks}
\end{table}

\begin{figure}[h]
\centering
\includegraphics[width=0.8\textwidth]{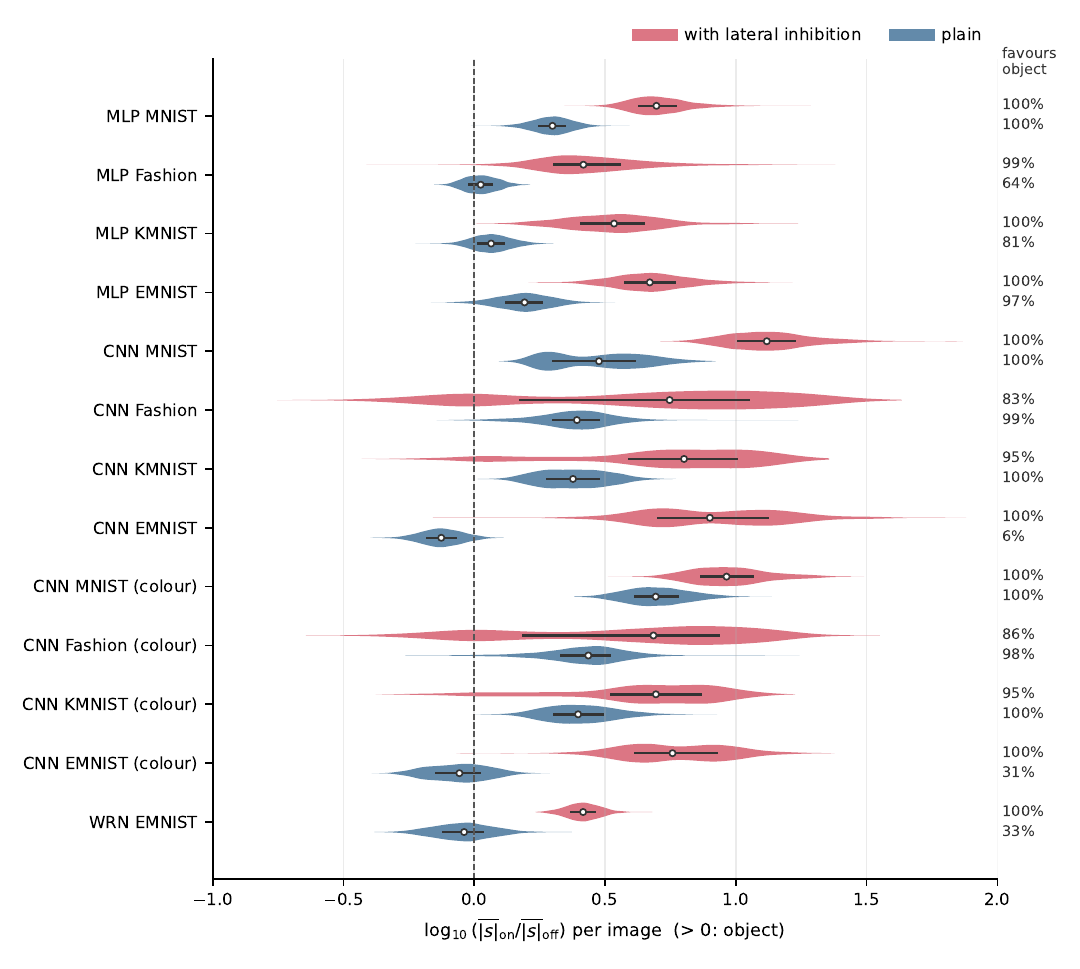}
\caption{\textbf{Saliency on and off the object, per image.} Distribution over images of the signed log-ratio $\log_{10}(\overline{s}_\text{on}/\overline{s}_\text{off})$ of the true-class gradient saliency, for every trained network of Section~\ref{sec:saliency} (five seeds pooled, $1000$ test images each; WRN: one network), with lateral inhibition and plain; bar: interquartile range, dot: median; right: the share of images favouring the object (log-ratio $>0$). No image was excluded. With lateral inhibition every network favours the object, on average and in $83$--$100\%$ of images; several plain networks favour the background (EMNIST CNN, grayscale and colour, and WRN).}
\label{fig:saliency_dist}
\end{figure}

\subsection{Ranking of the alternatives}
\label{app:ranking}

For a test image of digit $s$ and an alternative $d\ne s$, $\log[p(d\,|\,x)/p(s\,|\,x)]=z_d-z_s$, the logit gap. We took every source digit $s=0,\dots,9$ and all its MNIST test images, for the MLP and the CNN with and without lateral inhibition (five seeds each), and fitted a temperature $T$ per network by minimising the negative log-likelihood on its validation split~\citep{Guo2017}; dividing the logits by $T$ rescales every gap and leaves every ranking unchanged. Table~\ref{tab:ranking} reports calibration (expected calibration error, $15$ bins), the median margin (top logit minus the second) and entropy over correctly classified test images, before and after calibration, and, averaged over the ten source digits: the $\log_{10}$ ratio of the most likely alternative and the spread of the nine alternatives (mean gap over the images of $s$, divided by $\ln10$); the agreement of the ranking between seeds (mean pairwise Spearman correlation of the five seeds' mean-gap vectors); its stability over images (median Spearman correlation between an image's gaps and the network's mean gaps, and the share of images whose most likely alternative is the network's); and its agreement with two references for resemblance. The first is model-free: for each test image of $s$, the share of its $50$ nearest training images (Euclidean distance between raw images) that belong to $d$, averaged over the images of $s$. The second is the held-out errors of all twenty networks pooled, counted from $s$ to $d$. We have no human confusability data for MNIST.

\begin{table}[h]\centering\small
\setlength{\tabcolsep}{3.5pt}
\begin{tabular}{lcccccccccccc}
\hline
 & & ECE (\%) & margin & entropy & top alt. & spread & \multicolumn{3}{c}{ranking stability} & \multicolumn{2}{c}{agreement with} \\
network & $T$ & raw / cal. & raw / cal. & raw / cal. & raw / cal. & raw / cal. & seeds & images & top & visual & errors \\
\hline
MLP, with & 1.38 & 1.69 / 0.59 & 8.5 / 6.1 & .004 / .038 & $-3.9$ / $-2.9$ & 3.3 / 2.4 & 0.90 & 0.79 & 0.42 & 0.65 & 0.74 \\
MLP, plain & 1.31 & 1.16 / 0.42 & 8.9 / 6.8 & .008 / .055 & $-3.8$ / $-2.9$ & 0.5 / 0.4 & 0.73 & 0.82 & 0.53 & 0.54 & 0.58 \\
CNN, with & 1.15 & 0.26 / 0.44 & 9.5 / 8.2 & .001 / .004 & $-4.9$ / $-4.3$ & 4.7 / 4.1 & 0.97 & 0.78 & 0.40 & 0.58 & 0.68 \\
CNN, plain & 2.61 & 0.76 / 0.23 & 24.6 / 9.4 & $2\!\times\!10^{-7}$ / .001 & $-12.2$ / $-4.7$ & 9.1 / 3.5 & 0.96 & 0.74 & 0.41 & 0.61 & 0.64 \\
\hline
\end{tabular}
\caption{Ranking of the alternatives to the correct digit (MNIST, five seeds, every source digit). $T$: fitted temperature; raw / cal.: before / after dividing the logits by $T$. Margin in logits, entropy in nats (medians over correctly classified test images). Top alt.\ and spread: $\log_{10}$ ratio of the most likely alternative to the correct digit, and between the most and least likely alternatives. Ranking stability and agreement: Spearman correlations (``top'': share of images whose most likely alternative is the network's), defined in the text. All columns except $T$, ECE, margin and entropy are means over the ten source digits.}
\label{tab:ranking}
\end{table}

The plain CNN is overconfident ($T=2.6$); once that is removed, its alternatives lie in the same range as with lateral inhibition, and its ranking is as consistent across seeds and as close to both references. The plain MLP is calibrated about as well as the MLP with lateral inhibition, so its flat alternatives are not a matter of scale: they span less than half an order of magnitude, agree less between seeds and agree less with both references. Within single images every network ranks its alternatives only loosely like its average (Spearman $0.74$--$0.82$), and its average top alternative is the top alternative of under half to just over half of the images.

\section{Backward pass of the lateral-inhibition layer}
\label{app:li-backward}

This appendix derives the backward pass of a layer that carries lateral inhibition: the gradient of the training loss with respect to the layer's learnable parameters, and the error that the layer passes back to the layers below it. We first write the chain rule at the level of dependencies only, which already shows the two properties that matter---the gates are isolated from the optimiser, and the input gradient splits into a trunk and a gate route (Eq.~\eqref{eq:two-routes})---and then evaluate every factor. A last part shows why the gates are computed from the layer input and not from its output.

\paragraph{Notation.} We consider one layer and drop the layer index. Its input is $x$ and its output $y$, both flattened to vectors; $L$ is the training loss and
\begin{equation}
e_y\equiv\frac{\partial L}{\partial y}
\end{equation}
is the error arriving at the layer output from the layers above. Products $\odot$ are elementwise, $\diag(v)$ is the diagonal matrix with $v$ on its diagonal, $\one$ is the all-ones vector, and all derivatives of elementwise functions are applied elementwise. The forward pass is
\begin{align}
t&=Wx+b, & t'&=\omega\odot\frac{t-\mu\one}{\sigma}+\nu, & z&=\tanh t', & \tilde z&=\phi(z),\label{eq:bw-trunk}\\
\tilde x&=\phi(x), & m&=\Pi\,\tilde x, & B&=\psi\bigl(\mathrm{Birth}(m)\bigr), & S&=\psi\bigl(\mathrm{Survival}(m)\bigr),\label{eq:bw-gate}\\
P&=(1-\tilde z)\,B+\tilde z\,S, & N&=(1-\tilde z)^2+\tilde z^2, & \tilde y&=\min(1,P/N), & y&=\pi(\tilde y)=\eta\,\tilde y-(\eta-1).\label{eq:bw-blend}
\end{align}
Here $W$ and $b$ are the layer's weights and bias; for a convolution, $W$ is the matrix of the convolution acting on the flattened input, so its entries are tied across spatial positions. $t$ is the pre-activation. $t'$ is the normalised pre-activation: $\mu$ is the mean of $t$ over the normalisation group and $\sigma=\sqrt{v+\epsilon}$, with $v$ the variance over the group and $\epsilon=10^{-5}$ (the PyTorch default), the normalisation group being (all units of the layer for the MLP's LayerNorm, one channel group of one image for the WRN's GroupNorm), and $\omega$, $\nu$ are the learnable scale and shift. Without normalisation (the CNN) $t'=t$. $z\in[-1,1]$ is the activation and $\tilde z=\phi(z)\in[0,1]$ its CA state, the state that lateral inhibition regulates; $\phi(u)=\tfrac12(u+1)$ as in Appendix~\ref{sec:formulation}. On the gate side of the layer, $\tilde x$ is the input's CA state, $\Pi$ the fixed linear neighbourhood operator on the grid of $x$ (each position takes the mean of its eight Moore neighbours, averaged over the input channels where the layer reduces them), and $\psi$ the fixed linear resampling of the gate maps from the grid of $x$ to the grid of $z$ (bilinear; the identity when the grids agree, as in the MLP). The gates are computed before the resampling, as in the implementation; since they are nonlinear, $\psi(\mathrm{Birth}(m))\ne\mathrm{Birth}(\psi(m))$ in general. $B$ and $S$ are the Birth and Survival gates of Appendix~\ref{sec:formulation}, built from bumps $\beta_{\gamma,\delta}(m;c)$ with logistic steps $\theta_\gamma$. The bound $\min(1,\cdot)$ is the clip of Appendix~\ref{app:range}; we write $\kappa=\mathbb 1[P/N<1]$ for its derivative ($\kappa=1$ everywhere for \texttt{B5/S3}, whose blend stays below $0.66$). $P$ is the blend's numerator and $N$ its self-energy normaliser ($N\equiv1$ for the plain blend used in the CNN). The zero-centring map $\pi$ has output scale $\eta$: $\eta=2$ when the output is returned to the activation range ($y=2\tilde y-1$, the MLP) and $1$ when it is passed on as an occupancy ($y=\tilde y$, the CNN and WRN). The learnable parameters are $W$, $b$, $\omega$ and $\nu$; $\psi$, $\Pi$, the gate functions and the bumps (with their slope $\gamma$ and half-width $\delta$) are fixed.

\paragraph{Structure of the backward pass.} Equations~\eqref{eq:bw-trunk}--\eqref{eq:bw-blend} make $y$ a function of two quantities, $y=Y(z,m)$: the activation $z$, computed on the trunk from $x$ and the parameters $\theta=\{W,b,\omega,\nu\}$, and the occupancy $m$, computed from $x$ alone. The maps $Y$ and $m(x)$ contain no learnable parameter, so $\partial m/\partial\theta=0$, and the chain rule gives, before any factor is evaluated,
\begin{align}
\frac{\partial L}{\partial\theta}&=\Bigl(\frac{\partial y}{\partial z}\,\frac{\partial z}{\partial\theta}\Bigr)^{\!\top}e_y ,\label{eq:bw-struct-theta}\\
\frac{\partial L}{\partial x}&=\Bigl(\underbrace{\frac{\partial y}{\partial z}\,\frac{\partial z}{\partial x}}_{\text{trunk route}}\;+\;\underbrace{\frac{\partial y}{\partial m}\,\frac{\partial m}{\partial x}}_{\text{gate route}}\Bigr)^{\!\top}e_y .\label{eq:bw-struct-x}
\end{align}
Two properties follow from the dependencies alone. First, the parameter gradient~\eqref{eq:bw-struct-theta} has no gate term: the rule reaches it only through $\partial y/\partial z$, evaluated at the current gate values, and since the rule's parameters are fixed, no gradient ever updates them. This does not make training independent of the rule: the gate values enter every weight gradient of the layer through $\partial y/\partial z$, and, as shown below, the gate slopes of every layer enter the weight gradients of the layers below it through the error they pass down. Second, the input gradient~\eqref{eq:bw-struct-x} is the sum of the usual trunk route, through the parameters, and a gate route that contains none; the gate route is how the rule shapes the gradients of the earlier layers. The rest of this appendix evaluates the factors: $\partial y/\partial z$ and $\partial z/\partial\theta$ for the parameter gradient, then $\partial z/\partial x$, $\partial y/\partial m$ and $\partial m/\partial x$ for the error passed below.

\paragraph{Gradient with respect to $W$ and $b$.} The weights and bias enter only through $t$, so with the error at the pre-activation
\begin{equation}
e_t\equiv\frac{\partial L}{\partial t}
\end{equation}
the parameter gradients are the usual outer product and sum,
\begin{equation}
\frac{\partial L}{\partial W}=e_t\,x^{\top},\qquad \frac{\partial L}{\partial b}=e_t .
\label{eq:bw-W}
\end{equation}
For a convolution, the gradient of a kernel weight is the sum of the entries of $e_t\,x^\top$ that share it, i.e.\ the correlation of $e_t$ with the input over spatial positions, and $\partial L/\partial b$ sums $e_t$ over positions; the WRN's convolutions have no bias, their shift being the normalisation's $\nu$. The pre-activation $t$ reaches the loss only through the trunk, $t\to t'\to z\to\tilde z\to\tilde y\to y\to L$; the gate values $B$ and $S$ depend on $x$ but not on $t$. The chain rule therefore gives
\begin{equation}
\frac{\partial L}{\partial t}=\frac{\partial L}{\partial y}\,\frac{\partial y}{\partial z}\,\frac{\partial z}{\partial t},
\qquad
\frac{\partial z}{\partial t}=\frac{\partial z}{\partial t'}\,\frac{\partial t'}{\partial t},
\qquad
\frac{\partial y}{\partial z}=\frac{\partial y}{\partial\tilde y}\,\frac{\partial\tilde y}{\partial\tilde z}\,\frac{\partial\tilde z}{\partial z},
\label{eq:bw-chain}
\end{equation}
where, written as a column vector, the gradient is assembled from the transposed Jacobians, $e_t=(\partial t'/\partial t)^{\!\top}(\partial z/\partial t')^{\!\top}(\partial y/\partial z)^{\!\top}e_y$. Each factor is a Jacobian matrix. The output, blend and activation maps act element by element, so their Jacobians are diagonal:
\begin{equation}
\frac{\partial y}{\partial\tilde y}=\eta\,I,\qquad \frac{\partial\tilde y}{\partial\tilde z}=\diag(\Phi),\qquad \frac{\partial\tilde z}{\partial z}=\tfrac12\,I,\qquad \frac{\partial z}{\partial t'}=\diag(f'),
\end{equation}
so that $\partial y/\partial z=\diag(\tfrac{\eta}{2}\Phi)$, while the normalisation couples the units of a group, $\partial t'/\partial t=\diag(\omega)\,J$. Multiplying the diagonal factors element by element, with $f'=1-z^2$ and $\Phi=\partial\tilde y/\partial\tilde z$ (which includes the clip factor $\kappa$), $e_t$ factorises into four terms,
\begin{equation}
\boxed{\;e_t=J^{\top}\Bigl[\omega\odot f'\odot\tfrac{\eta}{2}\,\Phi\odot e_y\Bigr]\;}
\label{eq:bw-e}
\end{equation}
read from the right: the incoming error $e_y$; the blend factor $\tfrac{\eta}{2}\Phi$, which is $\partial y/\partial z$; the activation derivative $f'$; and the normalisation, its scale $\omega$ and Jacobian $J$. We derive each in turn.

\emph{The blend factor $\tfrac{\eta}{2}\Phi$.} By the chain rule through Eq.~\eqref{eq:bw-blend}, $\partial y/\partial z=(\partial y/\partial\tilde y)(\partial\tilde y/\partial\tilde z)(\partial\tilde z/\partial z)=\eta\cdot\Phi\cdot\tfrac12$. $\Phi$ is the \emph{state sensitivity} of lateral inhibition: how strongly its output responds to a change of the regulated state $\tilde z$ at fixed input. Only $\tilde z$ depends on the parameters, and $B$, $S$ do not depend on $\tilde z$, so the quotient rule on $\tilde y=P/N$ with $\partial P/\partial\tilde z=S-B$ and $\partial N/\partial\tilde z=-2(1-\tilde z)+2\tilde z=2(2\tilde z-1)$ gives
\begin{equation}
\Phi=\frac{\partial\tilde y}{\partial\tilde z}=\kappa\,\frac{(S-B)\,N-P\cdot2(2\tilde z-1)}{N^2}=\kappa\,\frac{(S-B)-2(2\tilde z-1)\,\tilde y}{N},
\label{eq:bw-phi}
\end{equation}
For the plain blend ($N\equiv1$) the second term vanishes and $\Phi=S-B$: lateral inhibition scales the error reaching the weights by the local survival--birth contrast. Where the rule is near-degenerate ($S\approx B$) the weights receive little gradient; where the gates disagree they receive it in full, with a sign set by which gate dominates. The normaliser adds the state-dependent term $-2(2\tilde z-1)\tilde y/N$: it vanishes at $\tilde z=\tfrac12$, where $N=\tfrac12$ and $\Phi=2(S-B)$, and grows towards the rails with the sign opposite to the side of the state (negative for $\tilde z>\tfrac12$, positive for $\tilde z<\tfrac12$). The gate values $B$ and $S$ enter $\Phi$ but their slopes $B'$, $S'$ do not: the slopes of a layer's own rule do not appear in that layer's direct weight gradient, though they enter the weight gradients of the layers below (Eq.~\eqref{eq:dLdx}). Where the clip acts ($\kappa=0$), the layer passes no error through its trunk. The per-layer factor $|\tfrac{\eta}{2}\Phi|$ compounds multiplicatively over the stacked lateral-inhibition layers, so it must stay near unity for a deep stack to train; the normaliser raises it towards $2|S-B|$ (at $\eta=2$), which is why it is used where many layers are stacked (the MLP and the WRN) and omitted in the CNN.

\emph{The activation factor $f'$.} $\partial z/\partial t'=1-\tanh^2t'=1-z^2$, close to $1$ near $z=0$ and vanishing at the rails $z=\pm1$.

\emph{The normalisation factor $J^\top(\omega\odot\cdot)$.} With the standardised pre-activation $\bar t=(t-\mu\one)/\sigma$, $\partial t'/\partial t=\diag(\omega)\,J$ with the standardisation Jacobian
\begin{equation}
J=\frac{\partial\bar t}{\partial t}=\frac1\sigma\Bigl(I-\tfrac1n\one\one^{\top}-\tfrac1n\bar t\,\bar t^{\top}\Bigr)
\end{equation}
over a normalisation group of $n$ units (block diagonal across groups). Here $\bar t=(t-\mu\one)/\sigma$ with $\sigma=\sqrt{v+\epsilon}$, so the formula is exact with the implementation's $\epsilon$. It removes from the error its component along $\one$, the direction of a shift of $t$, to which the normalisation is invariant. Along $\bar t$, the direction of a rescaling, $J\bar t=\tfrac{\epsilon}{v+\epsilon}\,\bar t/\sigma$: the component is removed only up to the fraction $\epsilon/(v+\epsilon)$, because with $\epsilon>0$ the normalisation is invariant to rescaling only approximately (to within $10^{-5}/v$). Without normalisation $J=I$ and $\omega=\one$. The normalisation's own parameters receive $\partial L/\partial\omega=\bar t\odot e_{t'}$ and $\partial L/\partial\nu=e_{t'}$, with $e_{t'}=f'\odot\tfrac\eta2\Phi\odot e_y$ the error at $t'$ (summed over positions where the parameters are shared).

\emph{The incoming error $e_y$.} $e_y$ is what the layers above send back. For the last layer with lateral inhibition it comes from the classifier; for every other layer it is given by the recursion below, so Eq.~\eqref{eq:bw-e} applied layer by layer yields every weight gradient of the network.

\paragraph{The error passed to the layer below.} The layer below receives $\partial L/\partial x=(\mathrm dy/\mathrm dx)^{\top}e_y$, with the two routes of Eq.~\eqref{eq:bw-struct-x}: the \emph{trunk} $x\to t\to t'\to z\to\tilde z\to\tilde y$, and the \emph{gate} $x\to\tilde x\to m\to\{\mathrm{Birth},\mathrm{Survival}\}\to\{B,S\}\to\tilde y$. The trunk route is the chain above with $\partial t/\partial x=W$. Along the gate route, $\partial\tilde x/\partial x=\tfrac12$, $\partial m/\partial\tilde x=\Pi$, the resampling has the Jacobian $\Psi$ ($\Psi=I$ when the grids agree), and, since $N$ does not depend on the gates, $\partial\tilde y/\partial B=\kappa(1-\tilde z)/N$ and $\partial\tilde y/\partial S=\kappa\tilde z/N$ on the grid of $z$. Without resampling ($\Psi=I$) the two combine into the gate's \emph{input sensitivity}
\begin{equation}
g\equiv\frac{\partial\tilde y}{\partial m}=\kappa\,\frac{(1-\tilde z)\,B'(m)+\tilde z\,S'(m)}{N},
\label{eq:bw-g}
\end{equation}
with $B'$, $S'$ the derivatives of the gate functions. Differentiating the soft-OR $\mathrm{Birth}(m)=1-\prod_{c\in\mathcal B}(1-\beta_{\gamma,\delta}(m;c))$ gives
\begin{equation}
B'(m)=\sum_{c\in\mathcal B}\beta'_{\gamma,\delta}(m;c)\prod_{c'\in\mathcal B,\,c'\ne c}\bigl(1-\beta_{\gamma,\delta}(m;c')\bigr),\qquad
\beta'_{\gamma,\delta}(m;c)=\theta_\gamma'(m-\ell_c)\,\theta_\gamma(u_c-m)-\theta_\gamma(m-\ell_c)\,\theta_\gamma'(u_c-m),
\end{equation}
with $\ell_c=(c-\delta)/8$, $u_c=(c+\delta)/8$ the lower and upper bump edges and, for the logistic step, $\theta_\gamma'=\gamma\,\theta_\gamma(1-\theta_\gamma)$; $S'(m)$ is the same over the set $\mathcal S$. Each bump's slope is positive below its centre and negative above it, largest in magnitude at the band edges $\ell_c$ and $u_c$ and decaying away from them, but, for the logistic step, never exactly zero. At the operating point ($\gamma=15$, $\delta=0.5$) the bumps have no plateau, and the slope of the survival gate of \texttt{B5/S3} is $0.135$ even at occupancy $0$ (maximum $2.79$) and exceeds a tenth of its maximum over $64\%$ of the occupancy range. The gate route therefore carries most of its gradient where the neighbourhood occupancy is near the edge of a birth or survival band, but not only there. Collecting both routes with $\Lambda=\diag(\Phi)$, $F=\diag(f')$, $\Omega=\diag(\omega)$, $A_B=\diag(\kappa(1-\tilde z)/N)$, $A_S=\diag(\kappa\tilde z/N)$ on the grid of $z$ and $D_{B'}=\diag(B'(m))$, $D_{S'}=\diag(S'(m))$ on the grid of $x$,
\begin{equation}
\boxed{\;\frac{\mathrm dy}{\mathrm dx}=\frac{\eta}{2}\Bigl(\underbrace{\Lambda\,F\,\Omega\,J\,W}_{\text{trunk route}}\;+\;\underbrace{\bigl(A_B\,\Psi\,D_{B'}+A_S\,\Psi\,D_{S'}\bigr)\,\Pi}_{\text{gate route}}\Bigr)\;}
\label{eq:jac-li}
\end{equation}
Without resampling ($\Psi=I$), $A_B D_{B'}+A_S D_{S'}=\diag(g)\equiv G$, and the gate route is $G\,\Pi$.
Transposing gives the error passed to the layer below as the sum of the two routes of Eq.~\eqref{eq:two-routes},
\begin{equation}
\frac{\partial L}{\partial x}=\frac{\eta}{2}\Bigl(\underbrace{W^{\top}J^{\top}\,\Omega\,F\,\bigl[\Phi\odot e_y\bigr]}_{\text{trunk route}}\;+\;\underbrace{\Pi^{\top}\Bigl[B'\odot\Psi^{\top}\bigl[\tfrac{\kappa(1-\tilde z)}{N}\odot e_y\bigr]+S'\odot\Psi^{\top}\bigl[\tfrac{\kappa\tilde z}{N}\odot e_y\bigr]\Bigr]}_{\text{gate route}}\Bigr).
\label{eq:dLdx}
\end{equation}
For a single-channel Moore neighbourhood each row of $\Pi$ has eight entries $\tfrac18$, so with the $\tfrac12$ of the remap each of a unit's eight neighbours receives $\tfrac1{16}$ of its gate-route error; a channel-mean reduction over $C_{\mathrm{in}}$ channels divides this by $C_{\mathrm{in}}$. The gate route contains no $F$, $\Omega$, $J$ or $W$: it is an additive pathway that bypasses the activation, the normalisation and the weights, and it contains no learnable parameter. The slopes $B'$, $S'$ that were absent from the layer's own weight gradient appear here, so the rule shapes the gradients of the \emph{earlier} layers, through $e_y$ of the layer below, without ever producing a gradient for itself. In the WRN the block output adds the skip connection, so the error at a block input also receives the skip term (identity, or a $1\times1$ convolution), which bypasses lateral inhibition.

At the layer that reads the image (CNN, first stage) the input $x$ is the image itself, so the gate route ends at the pixels and feeds no parameter; it contributes only to input gradients such as the saliency maps of Figure~\ref{fig:saliency}. The per-channel image normalisation (Eq.~\eqref{eq:perchannel-gate}) adds, per channel $c$, the factor $\mathrm{sign}(m_c-\mathrm{med}_c)/\kappa_c$ and the weight $w_c$ to $g$; its statistics $\mathrm{med}_c$, $\kappa_c$, $w_c$ are held constant under differentiation. The input gradient of the CNN is therefore a fixed-statistic surrogate, not the derivative of the forward map, which depends on the image also through these statistics (Appendix~\ref{app:perchgrad}). The choice changes no weight gradient, since the statistics are functions of the image alone.

\paragraph{A shallow variant.} Suppose the gates instead read the layer output: its occupancy is computed from $\tilde z$ rather than $\tilde x$ ($m=\Pi\tilde z$). Both routes then reach $x$ only through the single bottleneck $z$, and
\begin{equation}
\frac{\mathrm dy}{\mathrm dx}\Big|_{\text{shallow}}=\frac{\eta}{2}\,\bigl(\Lambda+G\,\Pi\bigr)\,F\,\Omega\,J\,W .
\label{eq:jac-shallow}
\end{equation}
The gate term $G\Pi$ has moved \emph{inside} the trunk envelope $F\Omega JW$: the activation throttle $F=\diag(1-z^2)$ now multiplies it too, so as the $\tanh$ stack saturates ($f'\to0$) the gate contribution vanishes together with the trunk, leaving no gradient route through a saturated layer. In the lateral-inhibition layer \eqref{eq:jac-li} the gate route sits \emph{outside} $F$ and survives saturation, giving an independent highway for the backpropagated error exactly where a bounded-activation stack is most prone to vanishing gradients. This is the gradient-level reason to compute the gates from the input $x$ and not the output $z$. The per-site operator norm of \eqref{eq:jac-li} at initialisation is what our epoch-0 probe reports; keeping it near unity is the condition for stable depth.

\section{The gate route in training}
\label{app:steering}

The weights of a layer with lateral inhibition cannot change its own gates, which read the layer below, but they set the input of every gate above them. Through the gate route of Appendix~\ref{app:li-backward} they receive gradient that changes where the occupancy of the layers above falls relative to the birth and survival bands. To measure how much of the learning signal takes this route, we computed each weight gradient twice on $256$ test images: normally, and with every gate route cut, the occupancy computed from a detached copy of the layer input while the forward pass is unchanged. In the trained MLP, the gradient that remains with the gate routes cut is below $0.2\%$ of the full gradient in layers $1$--$7$, about $1\%$ in layer~$8$ and $14\%$ in layer~$9$, the last layer below a gate. About $80\%$ of its cells sit below the survival band, where both gates are close to zero and the trunk factor vanishes, so error that crosses a layer through the trunk is suppressed at every layer it crosses, while the gate route passes it on through the slopes of the bands. In the trained WRN, despite its skip connections, the remainder is $0.02$--$4\%$ in every layer but the last, against $30$--$86\%$ at initialisation: training moves the network into a regime in which nearly all learning goes through the gates. The CNN is the exception. Its first lateral-inhibition layer reads the image and cannot be steered, and the gate route of its second carries about $5\%$ of the gradient of the first convolution.

To see what the gate route does, we retrained networks from the paper's initialisation and data order with every gate route cut (Table~\ref{tab:steering}). Training is deterministic: the MLP retrained normally reproduces the paper network exactly.

\begin{table}[h]\centering\small
\begin{tabular}{llcccccc}
\hline
network & gate routes & clean & noise & sal.\ log-ratio & cls.\ LFPR & SSI & lens@1 \\
\hline
MLP-10, MNIST & kept & 96.0 & -- & $+0.71$ & 0.69 & 18.2 & 62.2 \\
 & cut & 94.1 & -- & $+0.77$ & 0.23 & 9.8 & 54.9 \\
CNN, MNIST & kept & 98.5 & 92.2 & $+1.12$ & -- & -- & -- \\
 & cut & 98.6 & 91.8 & $+1.15$ & -- & -- & -- \\
WRN, CIFAR-10 & kept & 90.7 & 39.1 & -- & -- & -- & -- \\
 & cut & 80.7 & 36.4 & -- & -- & -- & -- \\
\hline
\end{tabular}
\caption{Networks retrained with every gate route cut, against the same networks trained normally from the same initialisation and data order; one network per row. Clean and noise accuracy in \%. Noise: Gaussian, shot and impulse noise at the five standard severities (CNN: $2000$ test images; WRN: the nine-level ladder over severities $1$--$5$ on $1500$ test images). Sal.\ log-ratio: $\log_{10}(\overline{s}_\text{on}/\overline{s}_\text{off})$ of the saliency, as in Section~\ref{sec:saliency}; cls.\ LFPR: classifier-weight LFPR; SSI: scale-separation index; lens@1: accuracy of the final classifier read at the first hidden layer (\%), as in Section~\ref{sec:scalesep}. The WRN is trained for $100$ epochs instead of the paper's $400$.}
\label{tab:steering}
\end{table}

The behaviour under noise and the localisation of saliency do not need the gate route: they come from the rule acting forward on whatever occupancy arises. What the gate route adds depends on the network and on the optimiser. The MLP is trained with Adam, which rescales each parameter's update by its own gradient history, so the small remaining gradient still trains it ($94.1\%$); what it loses is the coherence of its classifier weights and half of its scale separation. The WRN is trained with SGD, whose steps are proportional to the gradient, so cutting the gate routes shrinks every step by one to three orders of magnitude, and the network cannot fit its training set (final training loss $0.49$, against $0.05$). In the CNN, whose steerable gate carries little of the gradient, nothing measurable changes. The rule is therefore fixed in form but not out of the optimiser's reach: in a deep stack, training learns largely by placing each layer's activity where the fixed rule above it will act.

\section{Architectures}
\label{app:arch}

\begin{figure}[h]
\centering
\includegraphics[width=\textwidth]{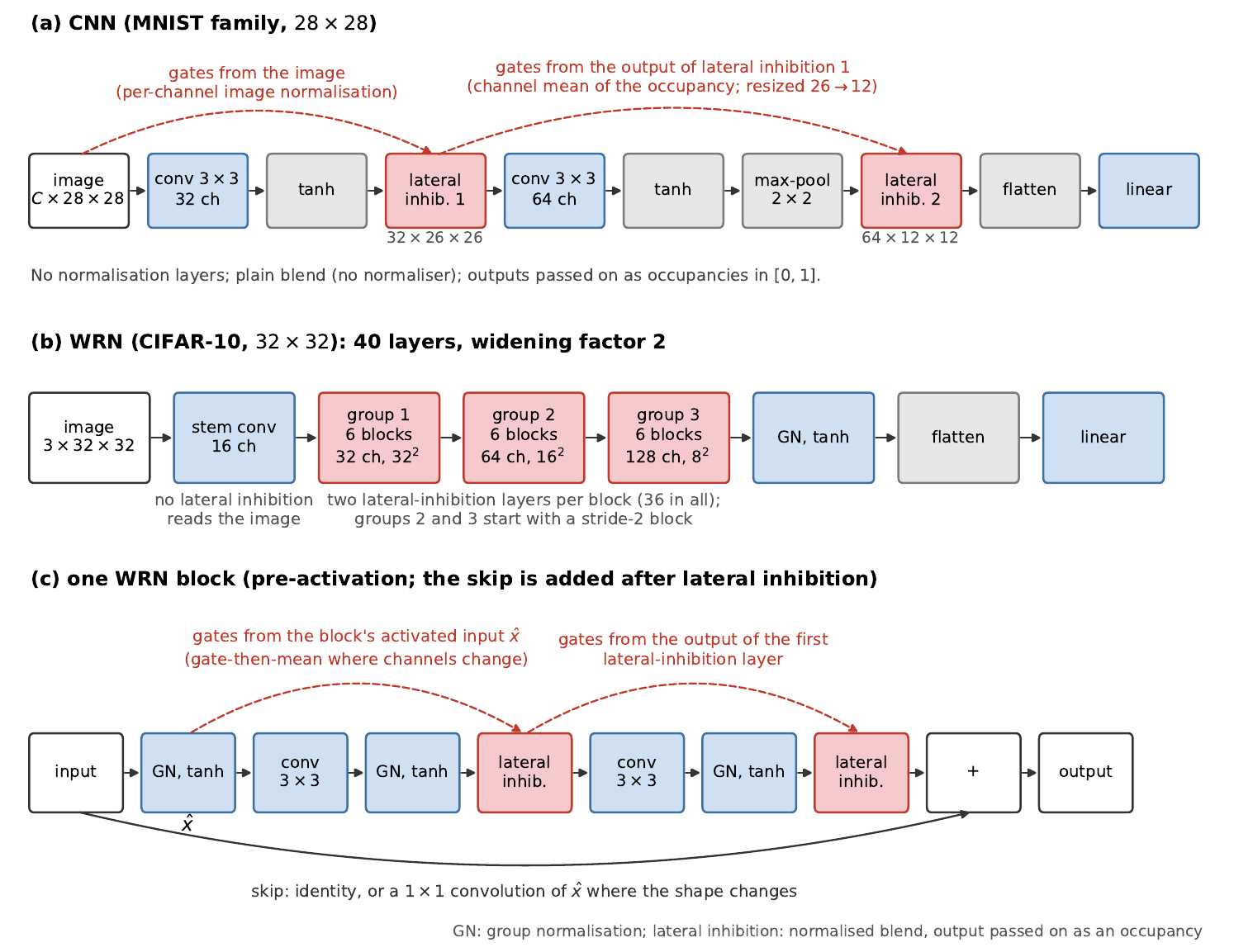}
\caption{\textbf{Architectures and wiring.} (a)~The CNN. (b)~The WRN, and (c)~one of its 18 residual blocks. Blue: learned operations; grey: fixed operations; red: lateral inhibition, with a dashed red arrow from the tensor its birth and survival gates are read from. Below the CNN's lateral-inhibition layers: the shape of their output.}
\label{fig:arch}
\end{figure}

Figure~\ref{fig:arch} shows the CNN and the WRN and where lateral inhibition is wired in each (the MLP repeats one layer, $\mathrm{Linear}\to\mathrm{LayerNorm}\to\tanh\to$ lateral inhibition, ten times on the $28\times28$ grid). In the CNN the first lateral-inhibition layer reads its gates from the image itself; in the WRN the stem convolution comes first, so every lateral-inhibition layer reads learned features. Each WRN block carries two lateral-inhibition layers, the first reading the block's activated input and the second the output of the first, and the skip connection is added after them.

\section{Matched-accuracy training of the 2-layer MLP}
\label{app:matched}

The 2-layer MLP of Figure~\ref{fig:noise_resilience} ($1.24$M parameters) uses the MLP settings of Section~\ref{sec:hyperparams} with a constant learning rate, no weight decay or clipping, and mild geometric augmentation (rotation ${\le}15^\circ$, translation ${\le}3$ pixels, scaling $0.95$--$1.05$, shear ${\le}10^\circ$). Each pair is first trained for a fixed budget ($400$ epochs on MNIST and FashionMNIST, $1000$ on KMNIST and EMNIST); a target is set $0.5$ points below the lowest best validation accuracy of the network with lateral inhibition over five seeds, and each network is retrained from the same initialisation, ending with a cosine decay, until its validation accuracy reaches the target ($+0.25$ points with lateral inhibition, $-0.25$ plain), so the two are compared at matched clean accuracy.

\section{Noise corruption protocol}
\label{app:noise}

The robustness evaluation uses the three noise corruptions of the common-corruptions benchmark~\cite{Hendrycks2019}. A test image is first scaled to $x\in[0,1]^{3\times H\times W}$ (an RGB image; see below for grayscale). For a severity parameter $c$, every element $x_i$ (one pixel of one channel) is corrupted independently:
\begin{align}
\text{Gaussian:}\quad & \tilde x_i=\mathrm{clip}\bigl(x_i+\varepsilon_i\bigr), && \varepsilon_i\sim\mathcal N(0,c^2),\\
\text{shot (Poisson):}\quad & \tilde x_i=\mathrm{clip}\bigl(k_i/c\bigr), && k_i\sim\mathrm{Poisson}(c\,x_i),\\
\text{impulse:}\quad & \tilde x_i=\begin{cases}0 & \text{with probability } c/2,\\ 1 & \text{with probability } c/2,\\ x_i & \text{otherwise},\end{cases}
\end{align}
with $\mathrm{clip}$ to $[0,1]$, after which $\tilde x$ is quantised to 8 bits ($\tilde x\leftarrow\lfloor255\,\tilde x\rfloor/255$) and normalised with the training statistics, $(\tilde x-\mu)/\sigma$. The severity ladders are
\begin{itemize}
\item Gaussian: standard deviation $c\in\{0.08,\,0.12,\,0.18,\,0.26,\,0.38,\,0.50,\,0.70,\,0.90,\,1.20\}$;
\item shot: scale $c\in\{60,\,25,\,12,\,5,\,3,\,2,\,1.2,\,0.7,\,0.4\}$, where a smaller scale means fewer counts and stronger noise;
\item impulse: replaced fraction $c\in\{0.03,\,0.06,\,0.09,\,0.17,\,0.27,\,0.40,\,0.55,\,0.70,\,0.90\}$.
\end{itemize}
The first five values of each ladder are the standard ImageNet-C severities of the benchmark (the CIFAR-10-C release uses milder values, e.g.\ $c=0.04$--$0.10$ for Gaussian noise); levels six to nine extend them toward stronger corruption.

\paragraph{Uniform-noise blend.} The breakdowns below add a fourth noise, not part of the benchmark: the image is blended with uniform noise, $\tilde x=(1-t)\,x+t\,u$ with $u_i\sim\mathcal U[0,1]$ drawn independently for every pixel and channel, at the fractions $t=0,0.1,\dots,1$. The blend needs no clipping, the same draw $u$ is used at every $t$ and for every network, and at $t=1$ the image is pure noise, so every network falls to chance. Unlike the three ladders, it also lowers the contrast of the image by the factor $1-t$.

\begin{figure}[h]
\centering
\includegraphics[width=\textwidth]{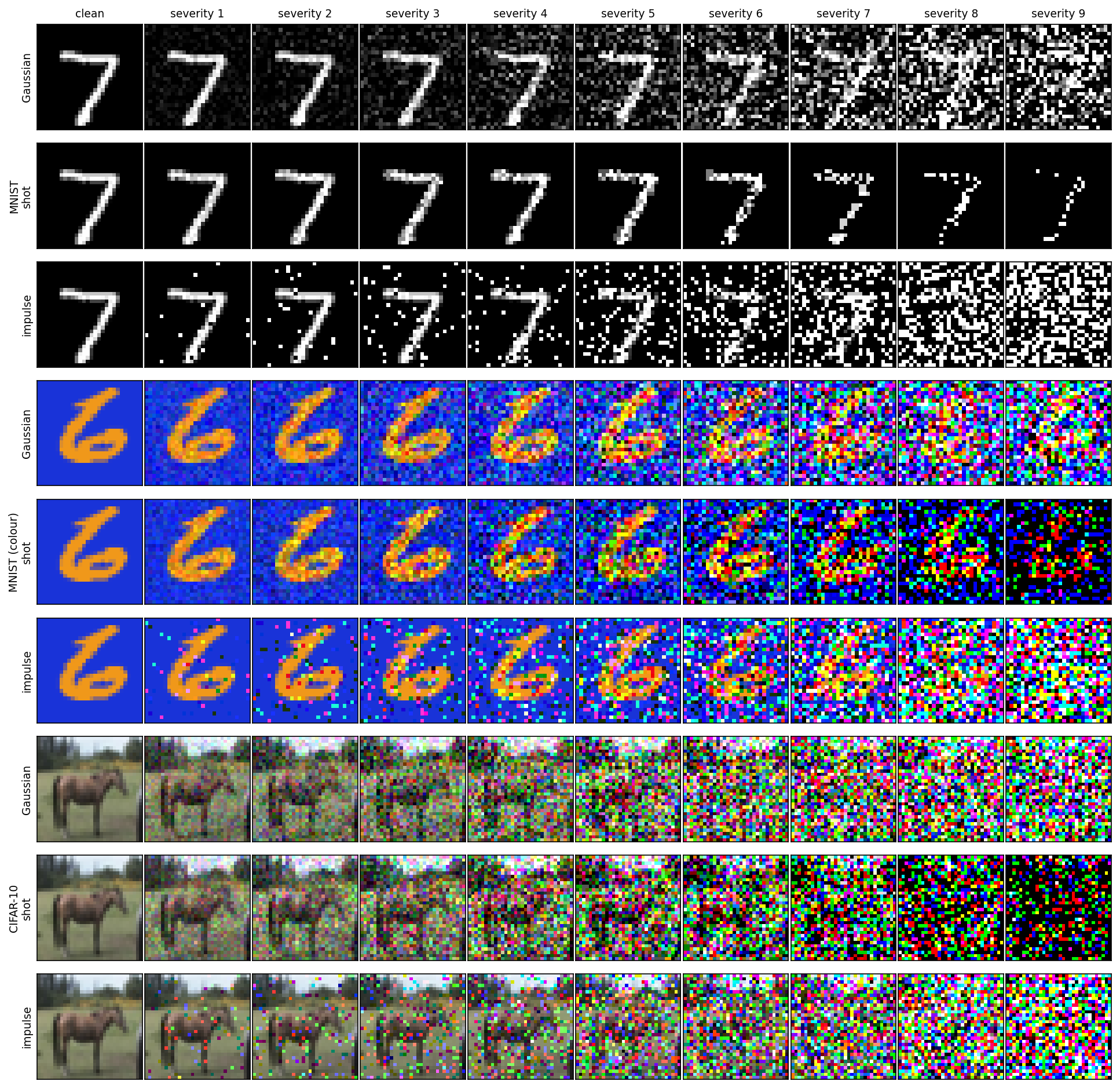}
\caption{\textbf{Example corrupted test images.} One MNIST digit, one colour-MNIST digit and one CIFAR-10 image under Gaussian, shot and impulse noise, from clean to severity~9, as the networks see them (the grayscale digit noised on its single channel). Gaussian and impulse noise saturate toward pure noise; shot noise leaves black regions black and does not.}
\label{fig:noise_examples}
\end{figure}

\paragraph{Saturating and non-saturating noise.} Gaussian and impulse noise are independent of the signal: every pixel receives the same amount of noise, and at the top of the ladder ($c=1.2$, or $90\%$ of the pixels replaced) the image is close to pure noise and every network approaches chance. Shot noise is signal-dependent: its variance, $x_i/c$, grows with the intensity, and a black pixel stays black ($\mathrm{Poisson}(0)=0$). It therefore never saturates: at the top of its ladder it breaks up the bright parts of the image into sparse counts, but leaves a black background clean and the object recognisable (Figure~\ref{fig:noise_examples}). The averaged noise accuracy thus combines two corruptions that approach chance with one that does not.

\paragraph{Grayscale images.} The corruption code operates on three-channel images. A grayscale image is copied into three identical channels, corrupted, and one channel is kept, $\tilde x^{\mathrm{gray}}=\tilde x_1$, so the grayscale image receives exactly the corruption defined above (Gaussian noise of standard deviation $c$; impulse noise that sets pixels to black or white). This matches MNIST-C~\cite{MuGilmer2019}: measured as the mean absolute change per pixel, MNIST-C shot noise equals our severity~$5$ ($0.0425$ against $0.0426$) and MNIST-C impulse noise our severity~$4$ ($0.0849$ against $0.0851$). All grayscale results in the paper use this procedure.

\paragraph{Evaluation.} For a given dataset we corrupt the first $1500$ images of the test set, with a random seed fixed for each noise type and severity (and, within it, for each image), so all networks in a comparison are evaluated on identical corrupted images. At each severity we average accuracy over the three corruptions. The resilience curves (Figure~\ref{fig:noise_resilience}) show this average at severities $0$ (clean) to $9$; the gains $\Delta_{1\text{--}5}$ and $\Delta_{1\text{--}9}$ are the mean differences over the standard severities and over the full ladder, and \emph{noise-mCA} is the mean over severities $1$--$9$. Single noise accuracies reported in tables use the five standard severities.

\paragraph{Uncertainty.}
\label{app:uncertainty}
The s.d.\ over seeds in the figures describes training variability only, on fixed test images and fixed noise draws; the three noise types and nine severities are repeated measurements on the same images, not independent experiments. For every comparison of Figure~\ref{fig:noise_resilience} we therefore kept the per-image results and the pairing (the $k$-th network with lateral inhibition is paired with the $k$-th plain network, and both see the same corrupted images) and computed three intervals for the gain (Table~\ref{tab:uncertainty}): a $t$-interval over the five seed pairs (training variability), a class-stratified bootstrap over the $1500$ test images with the networks fixed (test-image sampling), and a bootstrap that resamples both the seed pairs and, within them, the images ($2000$ resamples each). To measure the noise draws, we repeated the evaluation of four comparisons (CNN on MNIST and on colour FashionMNIST, 2-layer MLP on MNIST, WRN on CIFAR-10) with three further, independent draws of every corruption: the gain changes by an s.d.\ of $0.09$--$0.26$ points ($\Delta_{1\text{--}5}$) and $0.08$--$0.12$ ($\Delta_{1\text{--}9}$) between draws, far below the other two sources. Every gain or loss stated in the text has the same sign in all five seed pairs and an interval that excludes zero, except the 2-layer MLP on EMNIST ($\Delta_{1\text{--}5}={+}0.1$, interval ${-}0.8$ to ${+}1.0$; over the full ladder it is slightly negative, ${-}1.2$). The intervals concern these networks, test images and noise ladders; they do not cover the choice of datasets or of the rule.

\begin{table}[h]\centering\footnotesize
\setlength{\tabcolsep}{3pt}
\resizebox{\textwidth}{!}{%
\begin{tabular}{llccccccc}
\hline
 & & \multicolumn{5}{c}{$\Delta_{1\text{--}5}$ (points)} & \multicolumn{2}{c}{$\Delta_{1\text{--}9}$} \\
network & dataset & mean & per seed pair & seeds & images & both & mean & both \\
\hline
MLP-2 & MNIST & $+5.2$ & $+5.1\ +4.4\ +4.6\ +4.8\ +6.9$ & $[+3.9, +6.4]$ & $[+4.7, +5.6]$ & $[+4.4, +6.1]$ & $+2.2$ & $[+1.5, +2.9]$ \\
MLP-2 & Fashion & $+5.0$ & $+4.9\ +4.7\ +6.5\ +3.5\ +5.3$ & $[+3.6, +6.3]$ & $[+4.4, +5.6]$ & $[+3.9, +6.1]$ & $+3.1$ & $[+2.2, +4.0]$ \\
MLP-2 & KMNIST & $+4.6$ & $+5.0\ +5.9\ +3.9\ +4.3\ +3.9$ & $[+3.5, +5.7]$ & $[+4.0, +5.2]$ & $[+3.6, +5.5]$ & $+2.2$ & $[+1.4, +2.9]$ \\
MLP-2 & EMNIST & $+0.1$ & $+0.1\ +0.1\ -1.0\ +1.1\ +0.3$ & $[-0.8, +1.0]$ & $[-0.4, +0.7]$ & $[-0.8, +1.0]$ & $-1.2$ & $[-1.9, -0.5]$ \\
CNN & MNIST & $+14.7$ & $+14.2\ +15.4\ +15.1\ +15.3\ +13.5$ & $[+13.7, +15.7]$ & $[+14.1, +15.2]$ & $[+13.9, +15.5]$ & $+13.8$ & $[+12.9, +14.7]$ \\
CNN & Fashion & $+17.4$ & $+19.3\ +18.7\ +15.2\ +17.8\ +15.8$ & $[+15.2, +19.6]$ & $[+16.6, +18.1]$ & $[+15.7, +18.9]$ & $+11.0$ & $[+9.6, +12.3]$ \\
CNN & KMNIST & $+15.6$ & $+14.7\ +16.9\ +16.8\ +15.6\ +13.9$ & $[+14.0, +17.2]$ & $[+14.7, +16.5]$ & $[+14.3, +16.9]$ & $+10.3$ & $[+9.4, +11.2]$ \\
CNN & EMNIST & $+32.8$ & $+33.2\ +31.5\ +31.4\ +34.6\ +33.3$ & $[+31.1, +34.5]$ & $[+31.9, +33.6]$ & $[+31.4, +34.2]$ & $+20.9$ & $[+19.8, +21.9]$ \\
CNN & MNIST, colour & $+17.3$ & $+16.2\ +19.4\ +20.7\ +15.7\ +14.8$ & $[+14.2, +20.5]$ & $[+16.7, +18.0]$ & $[+15.3, +19.7]$ & $+11.9$ & $[+10.3, +13.9]$ \\
CNN & Fashion, colour & $+22.4$ & $+22.3\ +21.9\ +17.7\ +24.3\ +25.9$ & $[+18.6, +26.3]$ & $[+21.5, +23.3]$ & $[+19.6, +24.9]$ & $+12.3$ & $[+10.5, +13.8]$ \\
CNN & KMNIST, colour & $+19.3$ & $+19.8\ +21.2\ +19.9\ +18.0\ +17.7$ & $[+17.5, +21.1]$ & $[+18.4, +20.3]$ & $[+17.8, +20.9]$ & $+9.7$ & $[+8.5, +11.0]$ \\
CNN & EMNIST, colour & $+33.1$ & $+33.3\ +34.5\ +31.8\ +34.3\ +31.8$ & $[+31.5, +34.7]$ & $[+32.2, +34.1]$ & $[+31.7, +34.5]$ & $+18.8$ & $[+17.9, +19.6]$ \\
WRN & CIFAR-10 & $+6.3$ & $+8.9\ +3.1\ +8.4\ +5.2\ +6.1$ & $[+3.4, +9.3]$ & $[+5.9, +6.8]$ & $[+4.4, +8.3]$ & $+3.8$ & $[+2.5, +5.2]$ \\
GCN & MNIST & $+17.2$ & $+16.5\ +13.0\ +16.2\ +20.0\ +20.1$ & $[+13.5, +20.9]$ & $[+16.3, +18.0]$ & $[+14.7, +19.6]$ & $+9.5$ & $[+7.9, +11.2]$ \\
\hline
\end{tabular}}
\caption{Noise gain with lateral inhibition (with minus plain, accuracy points, mean over Gaussian, shot and impulse noise and over severities $1$--$5$ or $1$--$9$), with $95\%$ intervals: over the five seed pairs ($t$-interval), over the $1500$ test images (class-stratified bootstrap, networks fixed), and over both (bootstrap of seed pairs and images). Per seed pair: the gain of each of the five pairs.}
\label{tab:uncertainty}
\end{table}

\paragraph{Exploratory choices.} The recipe was not fixed in advance. The rule (Section~\ref{sec:rulesweep}), the classifier initialisation, the input scaling, the flatten head in place of global average pooling and the augmentation were chosen during development, on the MNIST family and CIFAR-10, and we did not hold out a separate evaluation set that these choices never saw. The large gains of the CNN are far outside the intervals above, but differences between design variants of a few points, such as the ablations of the appendices, are single comparisons made in the course of that exploration and should be read as such.

\begin{figure}[p]
\centering
\includegraphics[width=\textwidth]{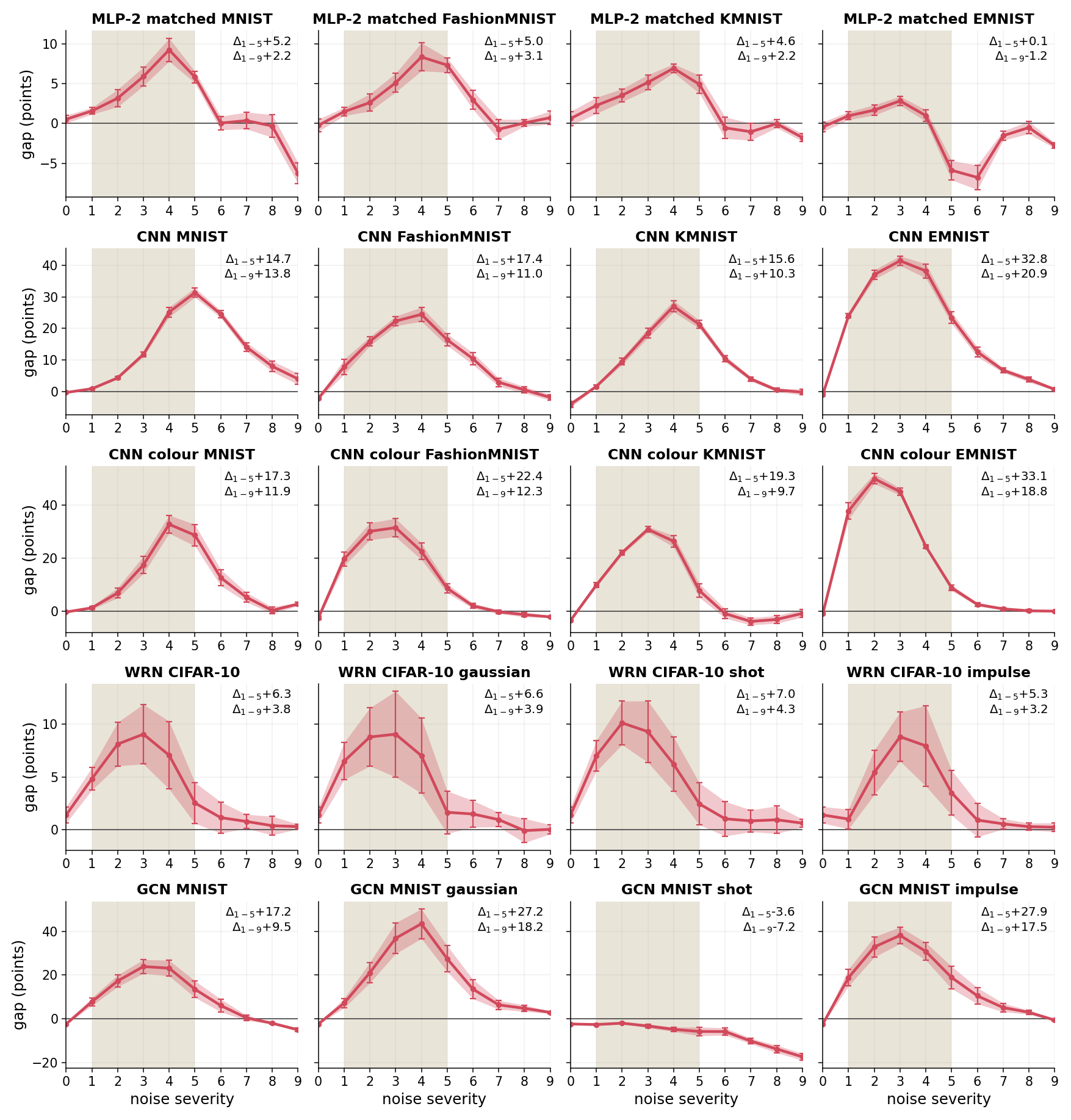}
\caption{\textbf{Accuracy gap under noise.} Accuracy with lateral inhibition minus plain, in points, for the networks and datasets of Figure~\ref{fig:noise_resilience}: mean over Gaussian, shot and impulse noise (the last three panels of the bottom two rows: the WRN and the GCN under each noise type), per seed pair, then mean over the five pairs (line) $\pm1$ s.d.\ (band and error bars). Severity $0$ is clean; the shaded band marks the standard severities $1$--$5$. $\Delta_{1\text{--}5}$ and $\Delta_{1\text{--}9}$: the mean gap over those severities, as in Figure~\ref{fig:noise_resilience}.}
\label{fig:noise_gap}
\end{figure}

\paragraph{The gap at each severity.} Figure~\ref{fig:noise_gap} plots, for every network of Figure~\ref{fig:noise_resilience}, the accuracy with lateral inhibition minus that of the plain network at each severity, per seed pair; severity $0$ is the clean-accuracy difference.

\begin{figure}[p]
\centering
\includegraphics[width=\textwidth]{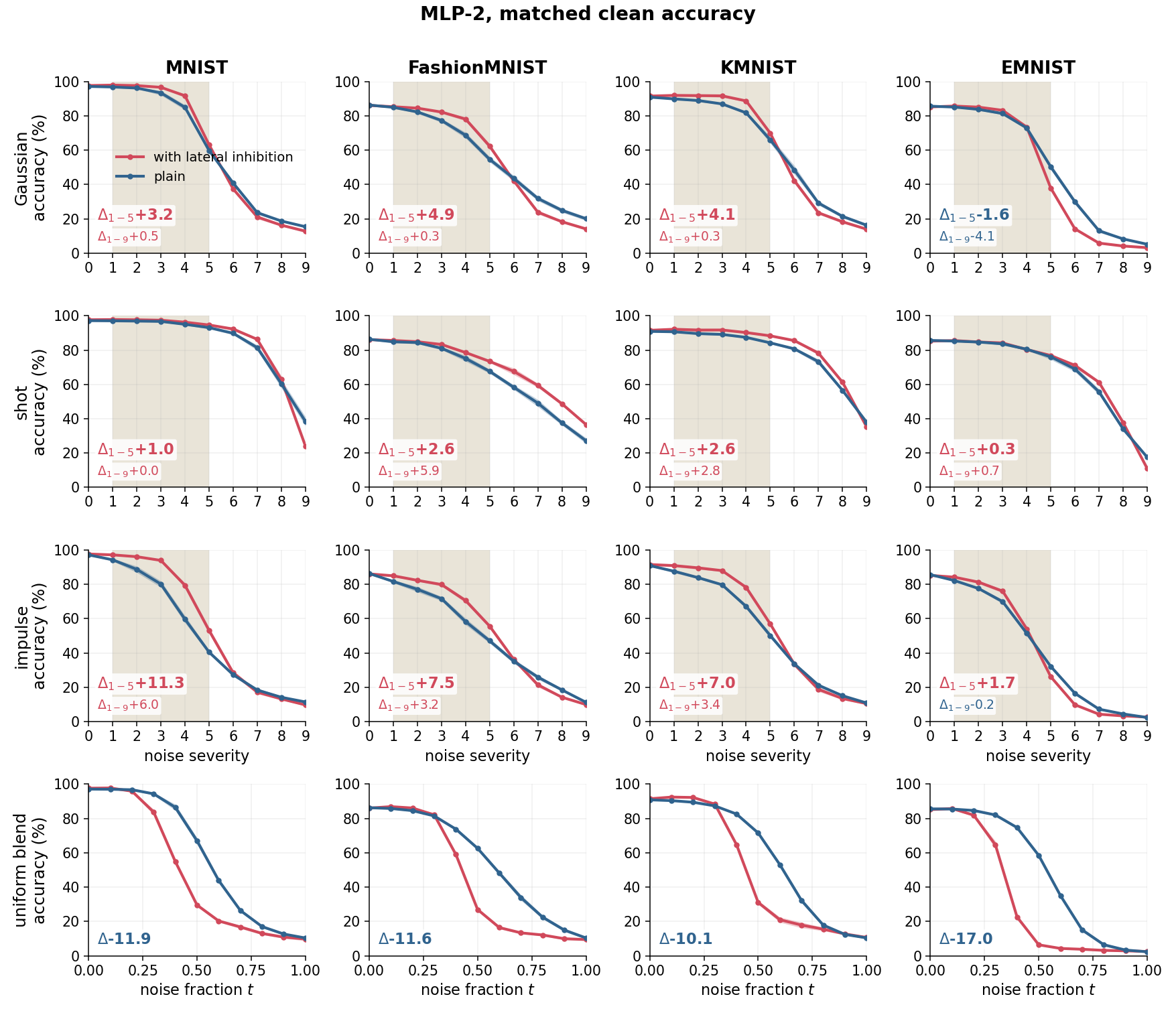}
\caption{\textbf{Noise breakdown, 2-layer MLP trained to matched clean accuracy} (Figure~\ref{fig:noise_resilience}, first row). Columns: dataset. Rows: Gaussian, shot and impulse noise on the 9-severity ladder (whose mean is Figure~\ref{fig:noise_resilience}), and the uniform-noise blend over the noise fraction $t$. With lateral inhibition (red) and plain (blue), mean $\pm1$ s.d.\ over five networks. $\Delta_{1\text{--}5}$ and $\Delta_{1\text{--}9}$: mean gain with lateral inhibition over severities $1$--$5$ and $1$--$9$; $\Delta$ (uniform blend): mean gain over $t=0.1$--$1$.}
\label{fig:nb_fcn2}
\end{figure}

\begin{figure}[p]
\centering
\includegraphics[width=\textwidth]{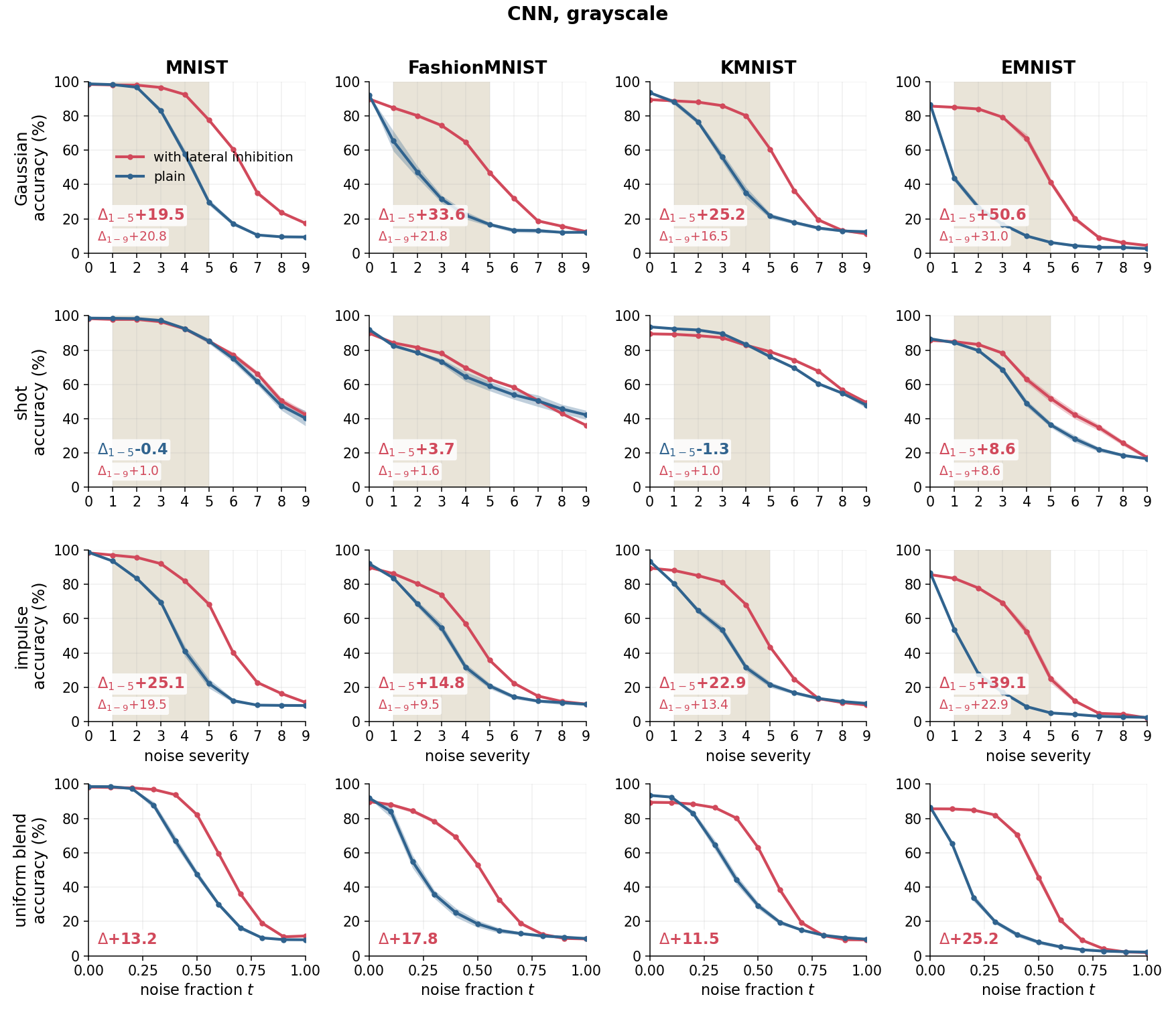}
\caption{\textbf{Noise breakdown, CNN on the grayscale MNIST family} (Figure~\ref{fig:noise_resilience}, second row). Layout as in Figure~\ref{fig:nb_fcn2}.}
\label{fig:nb_cnngray}
\end{figure}

\begin{figure}[p]
\centering
\includegraphics[width=\textwidth]{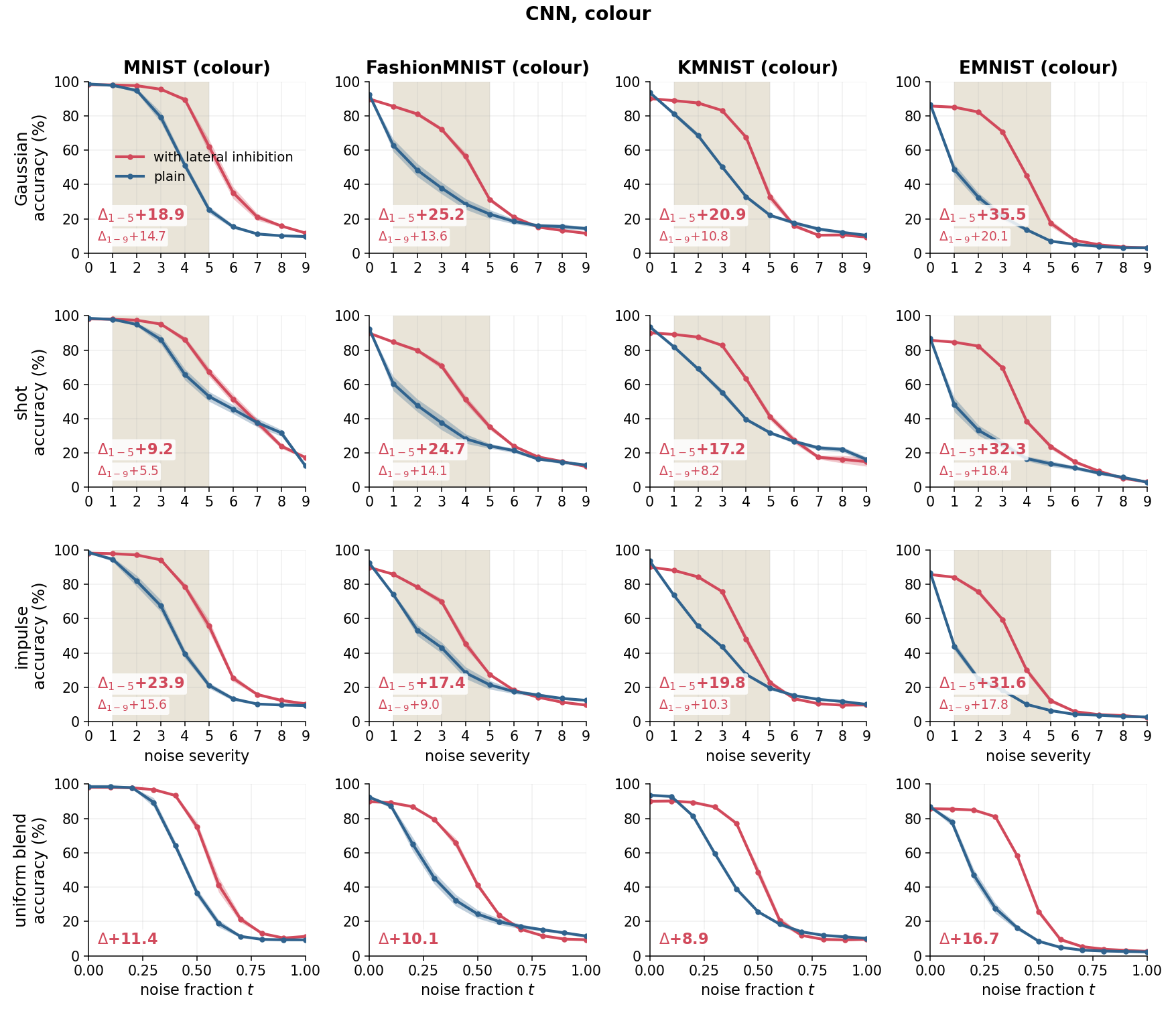}
\caption{\textbf{Noise breakdown, CNN on the colour MNIST family} (Figure~\ref{fig:noise_resilience}, third row). Layout as in Figure~\ref{fig:nb_fcn2}.}
\label{fig:nb_cnncolour}
\end{figure}

\paragraph{Breakdown by noise type.} Figures~\ref{fig:nb_fcn2}--\ref{fig:nb_cnncolour} show each noise type separately for the MLP at matched clean accuracy and the CNN (grayscale and colour); the WRN's and the GCN's breakdowns are the bottom two rows of Figure~\ref{fig:noise_resilience} (uniform blend: $\Delta={+}3.0$ and ${+}12.1$). The 2-layer MLP's gain comes mostly from impulse noise (${+}1.7$ to ${+}11.3$), with Gaussian ${-}1.6$ to ${+}4.9$ and shot ${+}0.3$ to ${+}2.6$, and it is below its plain counterpart under the uniform blend (${-}10$ to ${-}17$). The CNN (Figures~\ref{fig:nb_cnngray} and~\ref{fig:nb_cnncolour}) gains under every noise type: most under Gaussian (${+}19$ to ${+}51$) and impulse noise (${+}15$ to ${+}39$), and under the uniform blend (${+}9$ to ${+}25$). Under shot noise the grayscale CNN gains least (${-}1.3$ to ${+}8.6$), because shot noise leaves the black background clean, while the colour CNN, whose background is not black, gains ${+}9$ to ${+}32$.

\paragraph{Standard benchmarks.} For completeness we evaluated the paper networks on the noise corruptions of the official releases of CIFAR-10-C~\cite{Hendrycks2019} and MNIST-C~\cite{MuGilmer2019}, on the full test sets. Table~\ref{tab:officialc} gives the WRN on CIFAR-10-C. Its five severities cover only the mild end of our ladder (its strongest Gaussian noise, $c=0.10$, lies between our severities $1$ and $2$), which is why Figure~\ref{fig:noise_resilience} shows the WRN on our ladder. MNIST-C has one fixed severity per corruption and no Gaussian noise, and its two noise corruptions coincide with our severities~$5$ (shot) and $4$ (impulse). On it (mean over five seeds, s.d.\ $\le2.7$ points), the CNN with lateral inhibition reaches $85.0\%$ on shot noise and $83.5\%$ on impulse noise, against $85.0\%$ and $41.7\%$ without (clean $98.6\%$ and $98.9\%$).

\begin{table}[h]\centering\small
\begin{tabular}{llcccc}
\hline
network & benchmark & corruption & with (\%) & plain (\%) & $\Delta$ \\
\hline
WRN & CIFAR-10 & clean & 91.8 & 91.0 & $+0.8$ \\
 & CIFAR-10-C & Gaussian & 74.8 & 68.8 & $+6.0$ \\
 & & shot & 79.5 & 74.5 & $+5.0$ \\
 & & impulse & 79.2 & 76.3 & $+2.8$ \\
 & & speckle & 80.3 & 75.5 & $+4.8$ \\
\hline
\end{tabular}
\caption{``With'': with lateral inhibition; ``plain'': without. The WRN on the official CIFAR-10-C noise corruptions, inference only on the paper networks: mean over severities $1$--$5$, all $10{,}000$ test images per severity; mean over five networks (s.d.\ ${\le}1.3$ points). The gain is positive for every noise type in each of the five pairs.}
\label{tab:officialc}
\end{table}

\section{Continuous CA dynamics as a sanity check}
\label{app:gol}

\begin{figure}[t]
\centering
\includegraphics[height=0.5\textwidth]{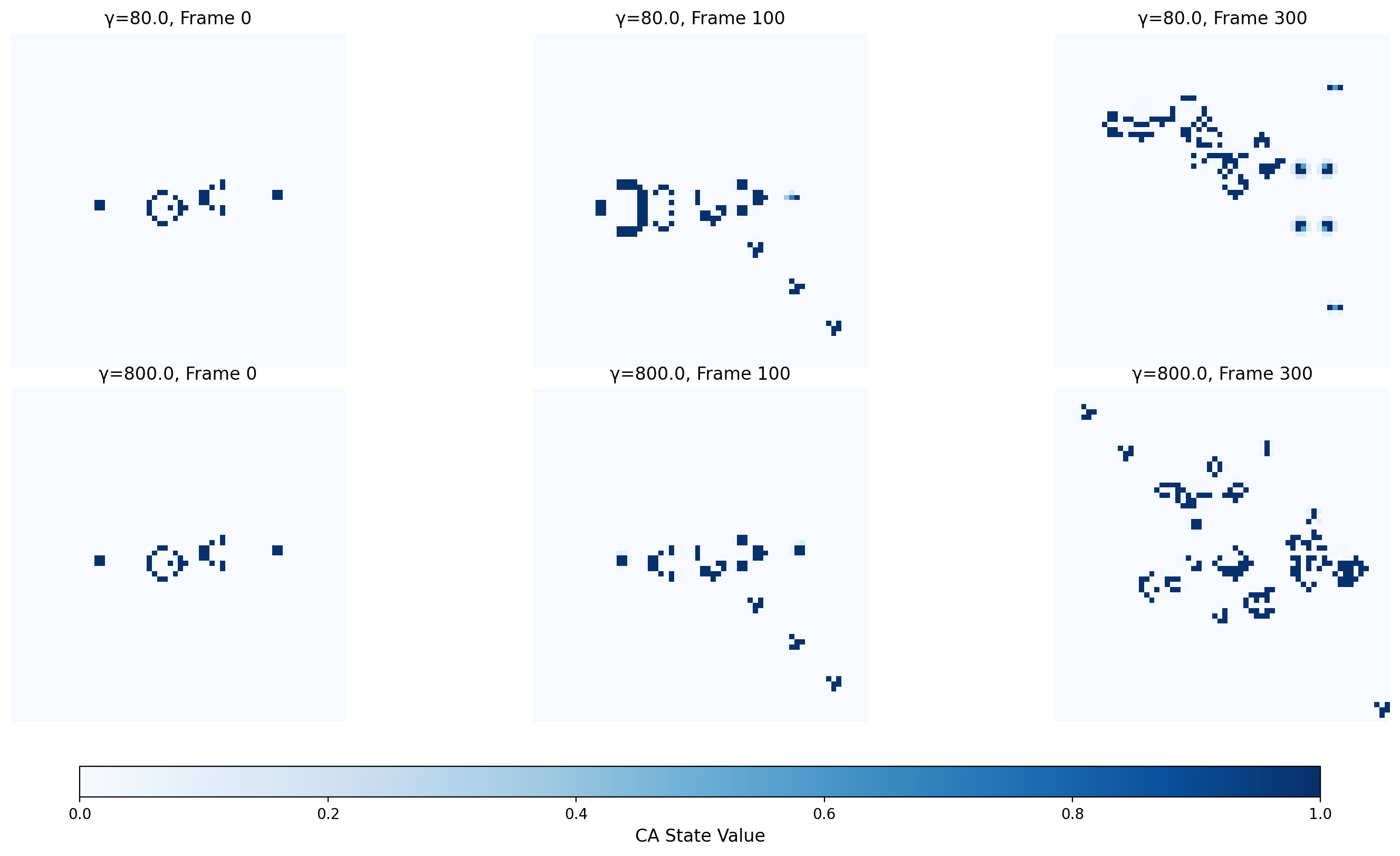}
\caption{\textbf{Glider gun in the continuous CA variant.} The Gosper glider gun simulated under the continuous B3/S23 rule for two slope values $\gamma$. Larger $\gamma$ approaches the discrete rule and preserves the emitted gliders over more steps; both values shown are far above the slope $\gamma=15$ used inside the networks, which does not sustain the gun as a standalone automaton (see text).}
\label{fig:glider_gun_continuous}
\end{figure}

The continuous rule used here is the simplest form that admits backpropagation; it is far less expressive than SmoothLife or Lenia~\cite{Rafler2011,Chan2019}. Its fidelity to the discrete Game of Life is governed by the slope $\gamma$ of the birth and survival gates: as $\gamma\to\infty$ the rule approaches the discrete limit, and hallmark structures such as a ``Gosper glider gun'' persist over hundreds of steps (Figure~\ref{fig:glider_gun_continuous}); as $\gamma\to 0$ the gates become mushy and the dynamics wash out entirely. This sanity check therefore shows only that the continuous relaxation \emph{can} recover Game-of-Life behaviour in the appropriate regime---it does not describe the operating point used inside the network. In particular, the slope we use in all network experiments, $\gamma=15$, is well below what is needed to sustain a glider gun as a standalone automaton. This is deliberate: inside a network the setting is fundamentally different from an iterated CA---each layer applies the rule \emph{once}, with a distinct affine transformation between layers rather than a shared recurrence, so faithful Game-of-Life dynamics are neither expected nor required. We consequently do not use CA simulation to select the rule; instead we choose the birth/survival sets empirically and inspect the resulting activation maps directly (Section~\ref{sec:coherence}), asking whether they exhibit coherent, class-IV-like structure rather than whether they reproduce any particular automaton.

\section{Percolation of the co-activation graph during training}
\label{app:percolation}

We trained the MNIST MLP and CNN of the paper, with and without lateral inhibition, with their paper recipes and three seeds each, and saved the network after every epoch ($300$ epochs for the MLP, $200$ for the CNN). At every epoch we took a fixed random subsample of $1000$ hidden units (all ten layers of the MLP; both convolutional stages of the CNN, one node per channel and position) and their activations on a fixed set of $1000$ test images. Units whose activation varies by less than $10^{-3}$ (s.d.\ over the images) no longer respond to the input; their residual fluctuations are at rounding level and correlate spuriously with one another, so we drop them and call the others active. Edges are weighted by the Pearson correlation of two active units. Following Erd\H{o}s--R\'enyi percolation~\citep{ErdosRenyi1960}, as used for the connectivity of living neural networks~\citep{Breskin2006} and, recently, for the pruning of artificial ones~\citep{Pesce2026}, we add the edges strongest first and record the fraction of active units in the largest connected component (the giant component) against the mean degree $\langle k\rangle=2E/N$, $N$ the number of active units. We use positive correlations only: the plain networks' activations are signed and symmetric while those with lateral inhibition are one-sided, so that anti-correlations, which act as long-range shortcuts, would otherwise favour the plain networks. We report the transition $\langle k\rangle_c$, at which the giant component reaches half of the active units (a random graph: $\langle k\rangle_c=2\ln2\approx1.4$); the largest jump of the giant component caused by a single edge (a staircase of discrete jumps indicates groups of units merging in stages); the overlap (Jaccard index) of the edge set at $\langle k\rangle=3$ with that of the previous epoch; and $r_c$, the correlation of the edge at which the giant component reaches half of the units, which places each graph relative to its own threshold without choosing a degree. As a measurement floor, we recomputed these quantities at the last epoch on four further disjoint sets of test images. A rule-free control, the input blur $y=(1-\lambda)z+\lambda(2m-1)$ of Appendix~\ref{app:frame} at $\lambda=0.5$ and $0.9$, was trained with the same recipe and analysed identically (one network each).

\begin{figure}[!htbp]
\centering
\includegraphics[width=\textwidth]{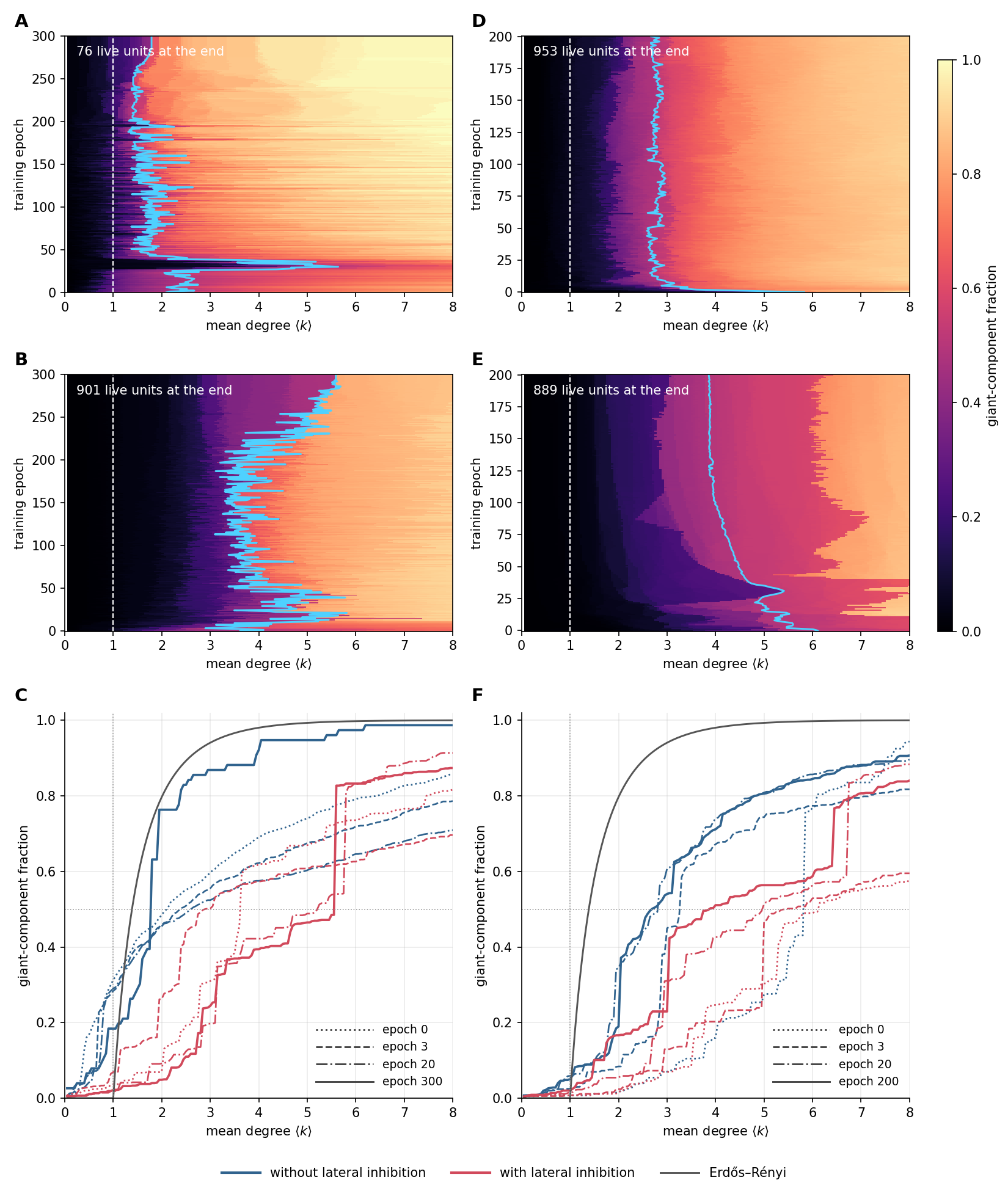}
\caption{\textbf{Percolation of the co-activation graph during training} (MNIST; left, MLP; right, CNN; one of the three seeds of Table~\ref{tab:percolation} each). (A,~D)~Plain network and (B,~E)~with lateral inhibition: the fraction of active hidden units in the giant component (colour) as edges are added strongest first, against the mean degree $\langle k\rangle$ and the training epoch; the cyan line marks the transition $\langle k\rangle_c$, where the giant component reaches half of the units, and the dashed line the random-graph threshold $\langle k\rangle=1$; the number of active units at the end of training is noted. (C,~F)~The percolation curves at epochs $0$, $3$, $20$ and the last, without (blue) and with (red) lateral inhibition, and the Erd\H{o}s--R\'enyi curve of a random graph with the same mean degree (grey). Edges: positive correlations between the active units among $1000$ sampled hidden units, over $1000$ test images.}
\label{fig:percolation}
\end{figure}

\paragraph{The plain MLP loses its active units.} From about epoch $25$, in every network, the plain MLP's hidden units stop responding: of the $1000$ sampled, $870$--$880$ are active at epoch $30$, $590$--$640$ at epoch $35$, $220$--$250$ at epoch $50$ and $67$--$78$ at the end (Figure~\ref{fig:netsci}B), while its validation accuracy keeps rising to $98.3$--$98.5\%$. Counted over all units, as in an earlier version of this analysis, the inactive units appear as isolated nodes and the graph seems to break up; on the active units its giant component never falls below $0.64$ at $\langle k\rangle=8$. The dark band of Figure~\ref{fig:percolation}A (epochs $\approx28$--$38$) is the transient while the units stop. With lateral inhibition the opposite happens: $713$--$729$ units are active at initialisation and $901$ at the end. In the CNN almost all units stay active in both networks.

\paragraph{Readouts.} Table~\ref{tab:percolation} gives the plateau of training. With lateral inhibition the giant component forms $2$--$3$ times later than in the plain MLP and $1.3$--$1.5$ times later in the CNN, in every network, and in the MLP it grows in larger steps. In the MLP its graph also changes little over training: $93\%$ of its edges are kept from one epoch to the next, more than between disjoint image sets at a fixed epoch ($61\%$), whereas the plain MLP's graph changes from epoch to epoch ($68\%$ kept) by more than its image-set floor ($84\%$). In the CNN both graphs keep $95$--$99\%$ of their edges between epochs. In the MLP both blurs reproduce the late threshold and the stability, and the blur at $\lambda=0.9$ also the steps; in the CNN the blur at $\lambda=0.9$ reproduces all three.

\paragraph{The graph lies in the image plane.} Every MLP hidden unit sits at a pixel (its index modulo $784$), so the graph can be compared with the image. With lateral inhibition, at $\langle k\rangle=3$ (one network, epochs $100$ and $300$), $76$--$84\%$ of the edges join units within two pixels of each other (random pairs: $3\%$), $16$--$17\%$ join the same pixel in different layers, and only $11$--$12\%$ lie within one layer. A force-directed layout (Fruchterman--Reingold, \texttt{networkx}) of the graph at $\langle k\rangle=4$, which sees only the edges, recovers the angle of each unit's pixel about the image centre (circular correlation $0.59$--$0.78$ in three networks, $0.76$ with the blur at $\lambda=0.9$; chance $0.05$), and the giant component's diameter is $35$--$67$ steps against $\approx11$ for a random graph with the same nodes and edges; its clustering coefficient is $0.43$ against $0.002$--$0.005$. The graph is thus a spatial, lattice-like graph in which each pixel's units, across layers, act together; near the threshold its giant component is one contiguous region of the image, which region varying between networks and epochs. Two reference graphs on the same nodes have the same signature: a two-dimensional random geometric graph~\citep{Dall2002} percolates at $\langle k\rangle_c=4.36$ with largest jump $0.29$ (with lateral inhibition, same network: $4.22$ and $0.29$), and a graph whose edges are added by increasing pixel distance at $3.51$. The late threshold and the staircase are therefore what a graph embedded in the image plane does, not a sign of a sharper transition: the steepest rise of the giant component is $1.7$ times steeper with lateral inhibition ($\max\mathrm{d}S/\mathrm{d}\langle k\rangle=6.0$--$7.2$ against $3.5$--$4.1$), but its rise from $20\%$ to $80\%$ of the units spans a wider range of $\langle k\rangle$ ($2.3$--$3.3$ against $1.8$--$2.3$). With $1000$ units and no finite-size scaling we do not interpret it as a phase transition. The degree distributions are not scale-free: on the graphs of correlations above $0.5$, a power law is rejected against both a lognormal and an exponential distribution in every network (normalised likelihood-ratio tests~\citep{Clauset2009}, $p<0.01$; $p=0.01$--$0.05$ for the plain MLP's $67$--$78$ units), and with lateral inhibition the degrees are narrow (mean $28$, maximum about $80$). The small-world index $\sigma=(C/C_r)/(L/L_r)$ exceeds $1$ in every network, but only because the clustering $C$ is far above that of a random graph ($C/C_r=90$--$160$ with lateral inhibition at $\langle k\rangle=4$, $5$--$6$ in the plain MLP); with lateral inhibition the mean path $L$ is $2.6$--$2.9$ times that of a random graph, as in a lattice, against $1.2$--$1.3$ in the plain MLP, so $\sigma$ does not identify a small world here. The plain MLP's graph has no geometry (angle correlation at chance, $0.13$--$0.17$; $1.5$--$2\%$ of its strongly correlated pairs within two pixels) and short paths, as in a random graph, but its clustering is about six times that of a random graph ($0.30$--$0.32$) and the correlations of its $67$--$78$ active units have effective rank $\approx11$: a few global, overlapping features rather than a random graph.

\paragraph{Distance from the threshold.} $r_c$ is $0.79$--$0.81$ with lateral inhibition against $0.50$--$0.54$ in the plain MLP ($0.94$ with the blur at $\lambda=0.9$), and $0.60$--$0.63$ against $0.52$--$0.53$ in the CNN: with lateral inhibition the giant component is already in place among the strongly correlated pairs, which are mostly spatial neighbours, so this ordering restates the spatial smoothness of the maps (Section~\ref{sec:coherence}). In the plain MLP $r_c$ falls over training, but mostly because units stop: on the fixed set of units active at the end it falls only from $0.56$--$0.59$ at epoch $25$ to $0.50$--$0.54$.

\paragraph{Isolated nodes in Appendices~\ref{app:gridrule} and~\ref{app:collapse}.} The graph of those appendices (the $100$ highest-variance units of each layer, $|$correlation$|$ edges at $\langle k\rangle=6$) has the same problem in the plain network: on the paper's MNIST network $47\%$ of its nodes are inactive units, whose spurious mutual correlations take up the edge budget and leave active units isolated. Rebuilt on the active units only, at the same mean degree, the isolate fraction of the plain network is $0.17$ on MNIST, $0.29$ on KMNIST, $0.36$ on FashionMNIST and $0.14$ on EMNIST, against $0.22$, $0.25$, $0.20$ and $0.24$ with lateral inhibition, which has no inactive units. The plain network's higher isolate fraction on MNIST and KMNIST ($0.69$--$0.72$) therefore reflects units that stopped responding rather than a fragmented representation; the effective ranks of Appendix~\ref{app:collapse} are unaffected. The local-average and random-neighbour arms of Appendix~\ref{app:gridrule} were not rechecked.

\paragraph{Limitations.} One dataset (MNIST); the blur control is one network each; the geometry is measured in the MLP only (the CNN's units also have positions); a fixed graph size of $1000$ units. Code: \texttt{paper\_code/percolation\_*.py} and \texttt{paper\_code/isolates\_dead\_units.py} in the clean repository; videos of the graph over training and over $\langle k\rangle$ are produced by \texttt{paper\_figures/percolation\_epoch\_video.py} and \texttt{percolation\_video.py}. Videos of the graph forming as edges are added, strongest first, are at \href{https://doi.org/10.6084/m9.figshare.34906809}{doi.org/10.6084/m9.figshare.34906809} (MLP) and \href{https://doi.org/10.6084/m9.figshare.34906869}{doi.org/10.6084/m9.figshare.34906869} (CNN), and of the graph at $\langle k\rangle=4$ over training at \href{https://doi.org/10.6084/m9.figshare.34906875}{doi.org/10.6084/m9.figshare.34906875} (MLP) and \href{https://doi.org/10.6084/m9.figshare.34906884}{doi.org/10.6084/m9.figshare.34906884} (CNN).

\begin{table}[h]\centering\small
\resizebox{\textwidth}{!}{%
\begin{tabular}{llcccccccc}
\hline
network & & active units & $\langle k\rangle_c$ & giant at $\langle k\rangle=8$ & largest jump & edge overlap & floor overlap & $r_c$ \\
\hline
MLP & with lateral inhibition & $901$ & $4.96$ ($4.22$--$5.98$) & $0.90$ ($0.89$--$0.90$) & $0.32$ ($0.29$--$0.36$) & $0.93$ ($0.93$--$0.94$) & $0.61$ & $0.80$ ($0.79$--$0.81$) \\
    & plain & $67$--$78$ & $1.87$ ($1.65$--$2.07$) & $0.97$ ($0.97$--$0.98$) & $0.18$ ($0.16$--$0.19$) & $0.68$ ($0.67$--$0.69$) & $0.84$ ($0.82$--$0.86$) & $0.53$ ($0.50$--$0.54$) \\
    & blur $\lambda=0.9$ & $967$ & $4.56$ & $0.86$ & $0.33$ & $0.99$ & --- & $0.94$ \\
    & blur $\lambda=0.5$ & $948$ & $3.65$ & $0.94$ & $0.17$ & $0.88$ & --- & --- \\
CNN & with lateral inhibition & $889$--$898$ & $3.87$ ($3.54$--$4.07$) & $0.80$ ($0.70$--$0.85$) & $0.21$ ($0.19$--$0.23$) & $0.99$ ($0.99$--$0.99$) & $0.59$ ($0.58$--$0.59$) & $0.61$ ($0.60$--$0.63$) \\
    & plain & $950$--$954$ & $2.81$ ($2.78$--$2.85$) & $0.90$ ($0.89$--$0.91$) & $0.17$ ($0.13$--$0.21$) & $0.95$ ($0.95$--$0.96$) & $0.57$ ($0.57$--$0.58$) & $0.52$ ($0.52$--$0.53$) \\
    & blur $\lambda=0.9$ & $959$ & $4.95$ & $0.87$ & $0.28$ & $0.99$ & --- & --- \\
    & blur $\lambda=0.5$ & $955$ & $3.05$ & $0.85$ & $0.18$ & $0.96$ & --- & --- \\
\hline
\end{tabular}}
\caption{Percolation of the co-activation graph of the active units, over the plateau of training (MLP epochs $\ge75$, CNN epochs $\ge50$); positive-correlation edges. With and without lateral inhibition: mean over three networks (range); blur: one network. Active units: of the $1000$ sampled, at the end of training. $\langle k\rangle_c$: mean degree at which the giant component reaches half of the active units. Largest jump: of the giant-component fraction by a single edge. Edge overlap: Jaccard index of the edge sets at $\langle k\rangle=3$ of consecutive epochs. Floor overlap: between the edge sets of disjoint sets of test images at the last epoch. $r_c$: correlation of the edge at which the giant component reaches half of the units, at the end of training.}
\label{tab:percolation}
\end{table}

\section{Where lateral inhibition operates}
\label{app:occupancy}

\begin{figure}[t]
\centering
\includegraphics[width=\textwidth]{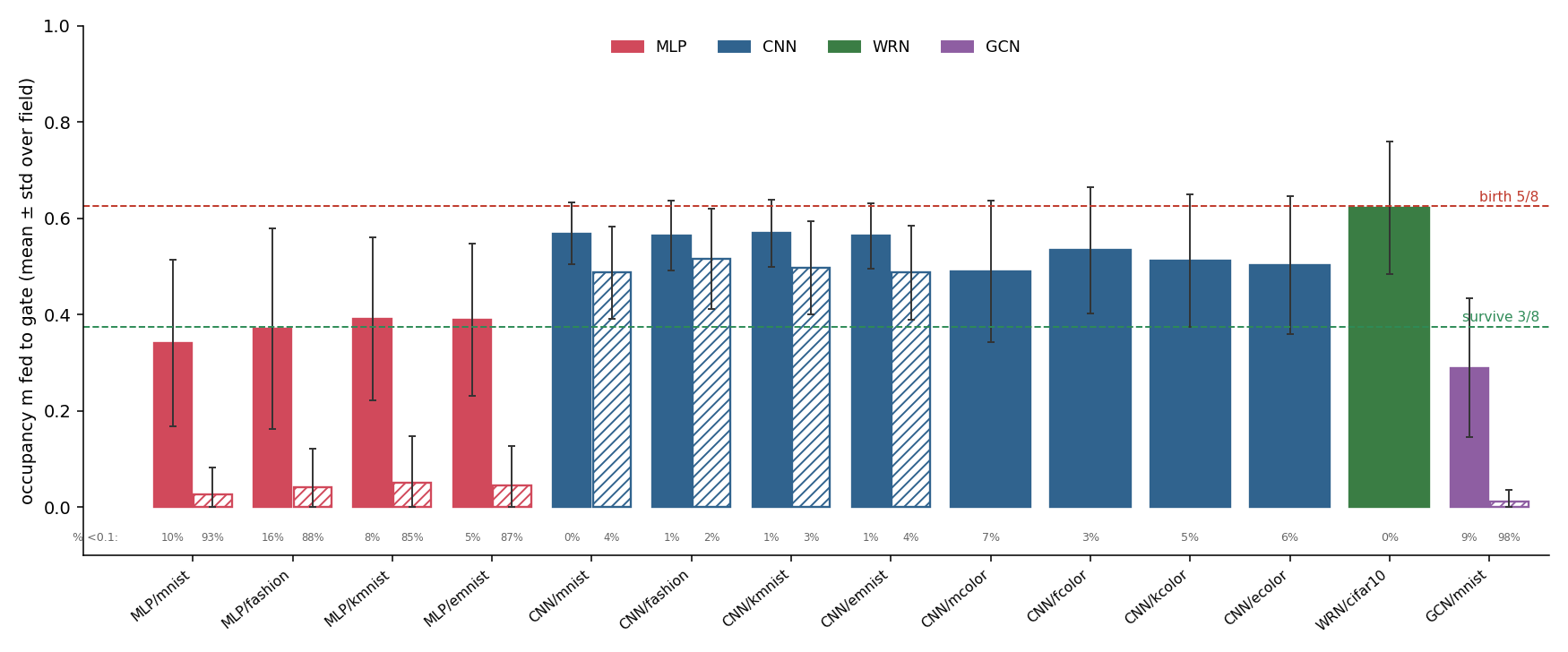}
\caption{\textbf{Operating point of lateral inhibition across architectures.} Mean $\pm$ standard deviation over the field (every lateral-inhibition layer, position and $250$ test images; the lower bar stops at $0$) of the neighbourhood occupancy $m$ fed to the birth and survival gates, for the MLP, CNN and WRN on each dataset and the GCN on MNIST (seed $0$). On the grayscale datasets, whose background is black, the field is split in two: the \emph{object} (solid bar), the positions whose $3\times3$ neighbourhood in the image contains a pixel above $0.1$, i.e.\ the strokes dilated by one pixel; and the \emph{background} (hatched bar), all other positions. The colour datasets and CIFAR-10 have no such background and show the whole field. The percentage below each bar is the fraction of the field under $0.1$ (sparsity). Dashed lines mark the B5/S3 birth ($5/8$) and survival ($3/8$) set-points. The MLP and the GCN read an empty background and an object field near or below survival, the CNN sits between the set-points on both (its first layer reads the per-channel normalised occupancy), and the WRN is dense near birth.}
\label{fig:occupancy}
\end{figure}

A mechanistic view of lateral inhibition itself asks \emph{where}, relative to the B5/S3 set-points (birth at $5/8$ and survival at $3/8$ occupancy), each architecture drives the neighbourhood occupancy $m$ that the rule reads. Figure~\ref{fig:occupancy} plots the mean and spread of $m$ over the field (every lateral-inhibition layer, position and test image) together with the fraction of the field below $0.1$ (its sparsity); on the black-background MNIST family it splits the field into the object and the background. The four backbones occupy distinct regimes. The MLP reads a nearly empty background and an object field close to the survival set-point, so its lateral inhibition acts predominantly through the Survival pathway, keeping the strokes and suppressing the rest, consistent with the sharp, digit-shaped maps of Section~\ref{sec:coherence}. The CNN sits between the two set-points on the object and the background alike, with little sparsity: pooled over its two lateral-inhibition layers, the occupancy is dominated by the stage-2 layer, which reads dense learned features, while the layer that reads the image presents the sparse, black-and-white-like field produced by the per-channel normalisation of Appendix~\ref{sec:channels}. The WRN is denser still, with mean occupancy near the birth set-point (${\approx}5/8$) and a sawtooth profile across blocks. The GCN resembles the MLP: an empty background and an object field below the survival set-point. Lateral inhibition therefore engages a different part of the Birth/Survival rule in each architecture---which helps explain why its representational and robustness effects are most pronounced where it reads a sparse field: the MLP's hidden states, the image in the CNN's first layer, and both layers of the GCN.

\section{Noise resilience along training}
\label{app:snapshots}

\begin{figure}[h]
\centering
\includegraphics[width=\textwidth]{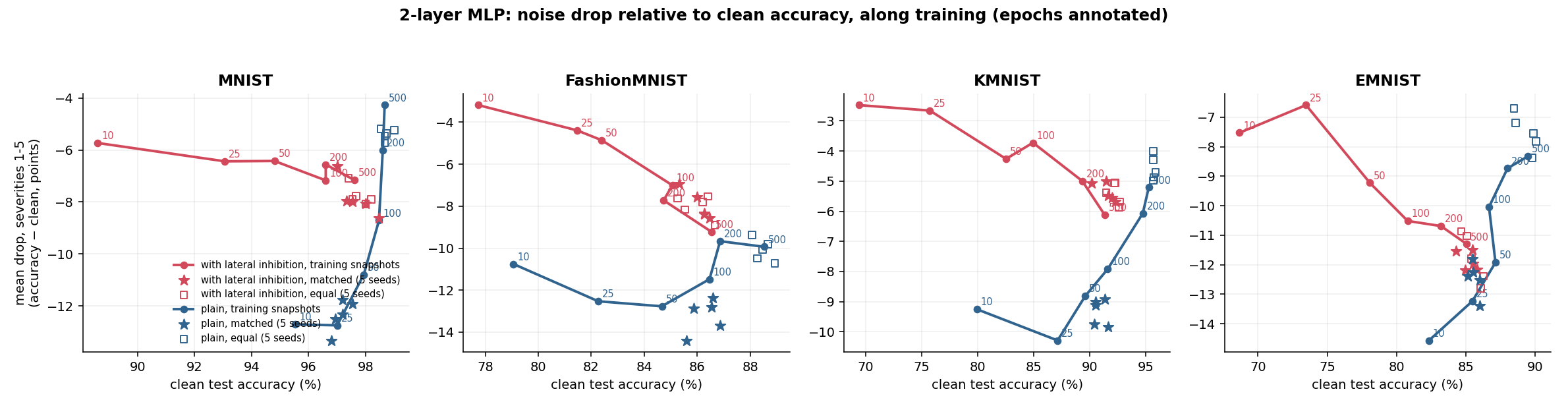}
\caption{\textbf{Noise drop along training} (2-layer MLP, seed 0). Mean drop over severities $1$--$5$ (accuracy minus clean accuracy) against clean accuracy, for snapshots at the annotated epochs, with (red) and without (blue) lateral inhibition. Stars: the networks trained to matched clean accuracy (five seeds); squares: the networks trained for an equal budget (five seeds).}
\label{fig:snapshots}
\end{figure}

The comparison between the networks with and without lateral inhibition depends on how long each is trained, because the two change in opposite directions during training. We trained the 2-layer MLP of the matched-accuracy experiment (seed 0, both arms, all four datasets) for $500$ epochs and saved it at epochs $10$, $25$, $50$, $100$, $200$ and $500$. For each snapshot we measured the drop under noise relative to the network's own clean accuracy, the mean of accuracy minus clean accuracy over severities $1$--$5$ (Figure~\ref{fig:snapshots}).

The plain network becomes more robust with training. Its drop shrinks steadily as its clean accuracy rises, from about $-13$ to $-4.5$ points on MNIST and from $-14$ to $-8$ on EMNIST; FashionMNIST is the exception, staying at about $-10$ throughout. The network with lateral inhibition becomes less robust with training. It starts very robust ($-2$ to $-7$ points at epoch $10$), and its drop grows as it trains (KMNIST $-2.1\to-6.2$, FashionMNIST $-3.2\to-9.4$): its accuracy under noise stays nearly constant while its clean accuracy rises. The curves cross. On MNIST, KMNIST and EMNIST the network with lateral inhibition has the smaller drop early in training, and the plain network has it by epoch $200$--$500$. The effect is not a floor effect: accuracy under noise stays far above chance at severities $1$--$5$, and the same trend holds for the drop expressed as a fraction of the accuracy above chance.

A comparison at matched clean accuracy therefore compares the plain network early in its training, before it has gained this robustness, with the network with lateral inhibition late in its training; a comparison at an equal training budget compares both late. The two are complementary; Figure~\ref{fig:noise_resilience} reports the comparison at matched clean accuracy.

\section{Reproducibility from nearly identical initialisations}
\label{app:riddling}

We follow the protocol of \citet{LyGong2026}. For each network (the MNIST CNN, with and without lateral inhibition, recipe of Section~\ref{sec:hyperparams}) one reference initialisation is drawn. Each member of an $\varepsilon$-close pair is the reference plus independent uniform noise $U(-\varepsilon/2,\varepsilon/2)$ on every parameter, for $\varepsilon=10^{-1},10^{-2},\dots,10^{-8}$; the smallest value is near the resolution of single precision for weights initialised in $[0.01,0.05]$. Both members are trained with the same seed and batch order and with deterministic GPU algorithms, so $\varepsilon$ is the only difference between them; a control pair started from the reference itself ($\varepsilon=0$) ends bitwise identical, with zero churn, for both networks. The churn of a pair is the fraction of the $10{,}000$ MNIST test images on which the two trained networks predict different classes. The functional uncertainty exponent $\gamma_f$ is the slope of the logarithm of the mean churn against $\log\varepsilon$, fitted by weighted least squares; $\gamma_f$ near zero means that a more precise initialisation does not make the trained function more reproducible.

With label noise, half of the training images, the same ones for every network, carry a class drawn uniformly from the nine wrong ones; validation and test labels are clean. Under this noise the plain CNN eventually memorises the wrong labels: in a pilot with three pairs at $\varepsilon=10^{-1},10^{-4},10^{-7}$, its test accuracy fell from $91.7\%$ at epoch $30$ to $82.6\%$ at epoch $50$ and $65.5\%$ at epoch $100$, and its churn at small $\varepsilon$ rose from $2.7\%$ to $9.3\%$ and $25.7\%$, while the network with lateral inhibition stayed near $95\%$ accuracy and below $0.2\%$ churn. To compare the networks before memorisation, and at the training horizon of \citet{LyGong2026}, the label-noise experiment is evaluated after $30$ epochs, with five pairs per $\varepsilon$. At every $\varepsilon\le10^{-4}$ all $25$ pairs of the plain network ($2.3$--$3.1\%$ churn) lie above all $25$ pairs of the network with lateral inhibition ($0.03$--$0.14\%$; one-sided rank test $p<10^{-14}$). The clean-label experiment is evaluated after $100$ epochs, with seven or eight pairs per $\varepsilon$; there the test accuracies are $98.4\%$ with lateral inhibition and $99.0\%$ plain.

\begin{figure}[h]
\centering
\includegraphics[width=\textwidth]{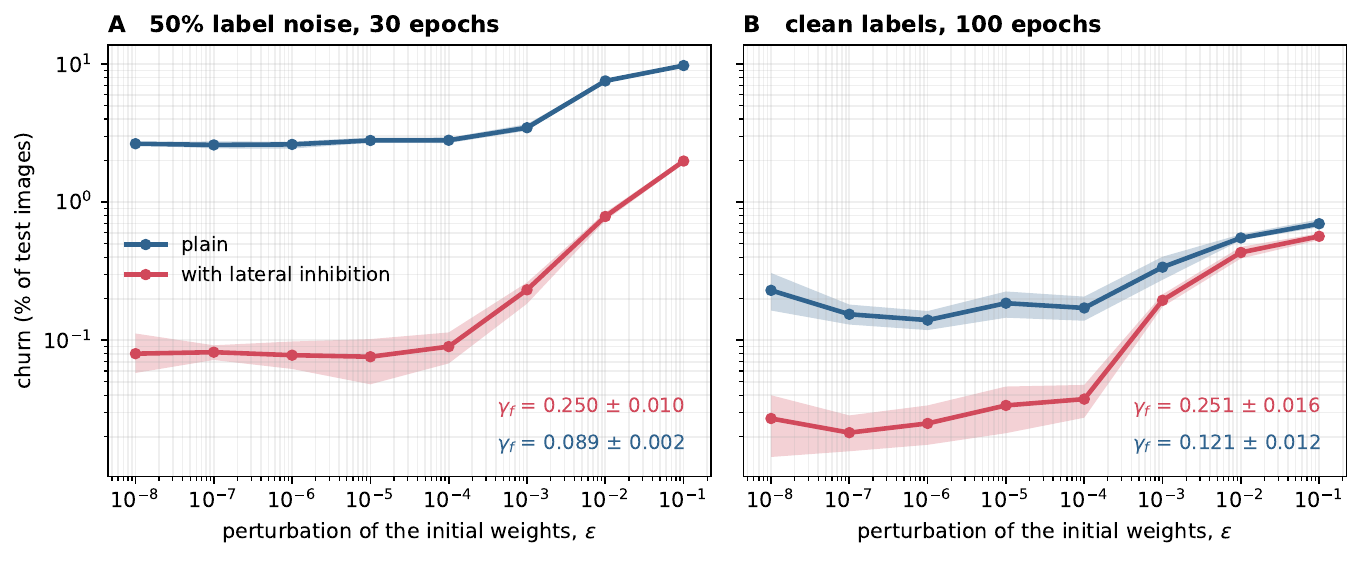}
\caption{\textbf{Churn between networks trained from nearly identical initialisations} (MNIST CNN). Churn, the percentage of test images on which the two networks of an $\varepsilon$-close pair disagree, against the size $\varepsilon$ of the perturbation of the initial weights, with (red) and without (blue) lateral inhibition; mean over pairs, band: $95\%$ bootstrap interval; $\gamma_f$: functional uncertainty exponent. (A)~$50\%$ label noise, $30$ epochs, five pairs per $\varepsilon$. (B)~Clean labels, $100$ epochs, seven or eight pairs per $\varepsilon$.}
\label{fig:riddling}
\end{figure}

\clearpage

\end{document}